%% file: main.tex
\documentclass[a4paper, 11pt]{article}
\ifdefined\XeTeXversion
\else
  \pdfoutput=1
\fi
\usepackage{etoolbox,verbatim}
\PassOptionsToPackage{authoryear,round}{natbib}

\newtoggle{arxiv}
\toggletrue{arxiv} 

\newtoggle{customthms}
\toggletrue{customthms} 

\newtoggle{colt}
\togglefalse{colt} 

\input{preamble/_preamble_includes}

\newcommand{\myparagraph}[1]{\vspace{2pt}\noindent{\bf #1}}

\title{Ordered Action Tokens for Visuomotor Policy Learning}

\author{
  \small Chaoqi Liu$^{1,2}$\quad
  Yue Zhao$^2$\quad
  Haonan Chen$^{1,2}$\quad
  Xiaoshen Han$^{1}$\quad
  Jiawei Gao$^{1}$\\
  \small
  Ehsan Adeli$^2$\quad
  Yilun Du$^1$\\
  \rule{.38\textwidth}{.7pt}\\
  \footnotesize
  $^1$Harvard University\quad
  $^2$Stanford University
}
\paperwebsite{ordered-action-tokenization.github.io}
\papercodes[Chaoqi-LIU/oat]{https://github.com/Chaoqi-LIU/oat}
  {Chaoqi-LIU/praxis-vla}{https://github.com/Chaoqi-LIU/praxis-vla}

\date{\vspace{-.5em}}

\begin{document}

\definecolor{titleblockbg}{HTML}{F5F5F5}
\newlength{\titleblockrulesep}
\setlength{\titleblockrulesep}{0.0em}

\begin{tcolorbox}[
colback=titleblockbg,
colframe=gray!50,
boxrule=0pt,
arc=2mm
]
\maketitle
\vspace{-1.5em}
\vspace{\titleblockrulesep}
\tcbline
\vspace{\titleblockrulesep}
\noindent\begin{minipage}{\linewidth}
  \centering
  \includegraphics[width=\linewidth]{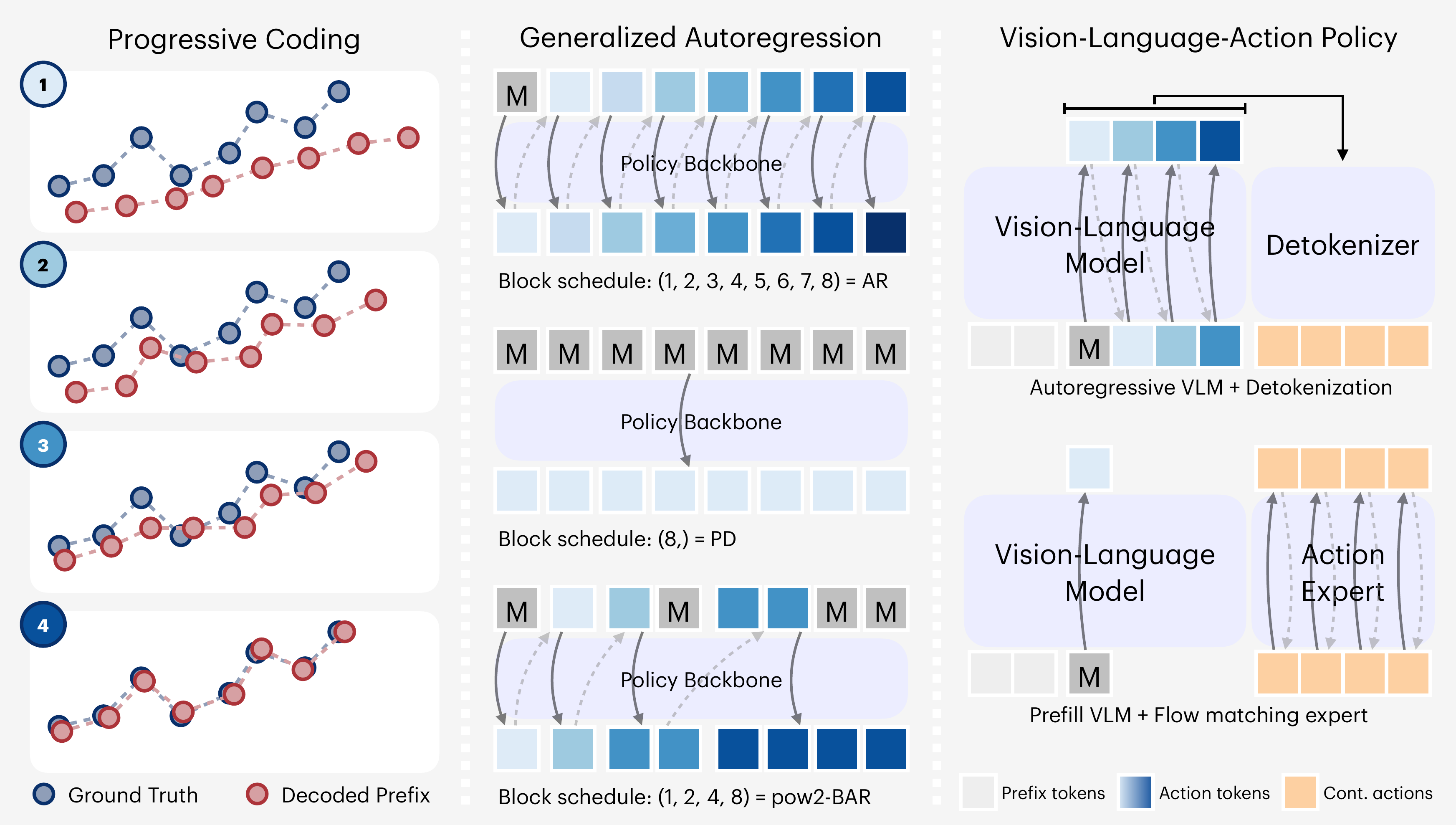}
  \captionof{figure}{\textbf{\oatfamily tokens for visuomotor policy learning.}
  \textbf{Left:} \oatfamily encodes each action chunk as an ordered token
  sequence whose prefixes decode to plausible action chunks, with later tokens
  refining control.
  \textbf{Middle:} In autoregressive policies, \oatfamily tokens can be generated
  under token-wise autoregression (\textbf{top}), parallel decoding
  (\textbf{middle}), or power-of-two grouping (\textbf{bottom}).
  \textbf{Right:} We validate \oatfamily in two prevailing uses of tokens:
  autoregressive policies that generate tokens decoded into control
  (\textbf{top}), and token co-training policies, where token losses train the
  vision-language model while a flow-matching expert generates actions from its
  prefill context (\textbf{bottom}).}
  \label{fig:oat_teaser}
\end{minipage}\par
\vspace{\titleblockrulesep}
\makeatletter
\ifdefempty{\metadatalist}{}{\tcbline\vspace{\titleblockrulesep}\metadatalist\par}
\makeatother
\end{tcolorbox}
\begin{abstract}
\input{body/abstract}
\end{abstract}

\input{body/introduction}
\input{body/prelim}
\input{body/action_tokenization}
\input{body/oat}
\input{body/oat_policy}
\input{body/experiments}
\input{body/related_work}
\input{body/conclusion}

% \clearpage
\acknowledgments{The computations in this paper were carried out in part on the FASRC cluster supported by the FAS Division of Science Research Computing Group at Harvard University, in part on cloud computing resources provided through the Lambda Research Grant Program, and in part on the Delta system at the National Center for Supercomputing Applications, the Anvil supercomputer at Purdue University, and the Bridges-2 system at the Pittsburgh Supercomputing Center through allocation CIS260779 from the Advanced Cyberinfrastructure Coordination Ecosystem: Services \& Support (ACCESS) program, which is supported by U.S. National Science Foundation grants \#2138259, \#2138286, \#2138307, \#2137603, and \#2138296.}

\clearpage
\begingroup
\setcitestyle{numbers,square,sort&compress}
\bibliographystyle{plainnat}
\bibliography{refs}
\endgroup
\clearpage
\appendix
\crefalias{section}{appendix}
\input{body/appendix}
\end{document}

%% file: preamble/_preamble_includes.tex
\input{preamble/color_defs}
\input{preamble/header}
\input{preamble/theorem_styles}
\input{preamble/characters}
\input{preamble/general_macros}
\input{preamble/specific_macros}
\input{preamble/algnames}

\usepackage{preamble/color-edits}
\input{preamble/comment_macros}

\iftoggle{arxiv}
{
\input{preamble/arxiv_title}
}
{}

%% file: preamble/color_defs.tex
\usepackage[dvipsnames,table]{xcolor}
\usepackage{tcolorbox}
\tcbuselibrary{skins,breakable}

\tcbset{
  aibox/.style={
      width=\linewidth,
      top=6pt,
      bottom=0pt,
      left=1.5pt,
      right=1.5pt,
      colback=blue!60!gray!5,
      colframe=black,
      colbacktitle=black,
      enhanced,
      center,
      attach boxed title to top left={yshift=-0.1in,xshift=0.15in},
      boxed title style={boxrule=0pt,colframe=white,},
    }
}
\newtcolorbox{AIbox}[2][]{aibox,title=#2,#1}

\definecolor{primalcolor}{HTML}{A60000}
\definecolor{contrarycolor}{HTML}{00A6A6}
\definecolor{darkcontrarycolor}{HTML}{004C4C}
\definecolor{lightblue}{HTML}{2970CC}
\definecolor{lightpurple}{HTML}{673147}
\definecolor{ForestGreen}{HTML}{FF5733}
\definecolor{myred}{HTML}{AA4A44}
\definecolor{hyppurple}{HTML}{800080}
\definecolor{notionprimary}{HTML}{5645D4}
\definecolor{notionprimarydeep}{HTML}{3A2A99}
\definecolor{notionlinkblue}{HTML}{0075DE}
\definecolor{notionlinkbluepressed}{HTML}{005BAB}
\definecolor{notionink}{HTML}{1A1A1A}
\definecolor{notioncharcoal}{HTML}{37352F}
\definecolor{notionslate}{HTML}{5D5B54}
\definecolor{notionsteel}{HTML}{787671}
\definecolor{notionhairline}{HTML}{E5E3DF}
\definecolor{highlightaccent}{HTML}{8A1538}

\newcommand{\linkcolor}{notionlinkblue}
\newcommand{\urlcolor}{notionslate}
\definecolor{icnclr}{HTML}{000000}
\definecolor{citecyan}{HTML}{5E55A2}
\newcommand{\citecolor}{notionprimarydeep}

\newcommand{\thmcolordark}{red!30!black}

\tcbset{
  highlightbox/.style={
      enhanced,
      width=\linewidth,
      arc=6pt,
      outer arc=6pt,
      boxrule=0.45pt,
      colback=notionlinkblue!3!white,
      colframe=notionhairline,
      coltitle=notionprimarydeep,
      fonttitle=\sffamily\bfseries\footnotesize,
      fontupper=\small,
      colbacktitle=white,
      attach boxed title to top left={xshift=11pt,yshift=-0.55\baselineskip},
      boxed title style={
        enhanced,
        arc=4pt,
        outer arc=4pt,
        boxrule=0pt,
        colback=white,
        colframe=white,
        left=5pt,
        right=5pt,
        top=1pt,
        bottom=1pt,
        boxsep=0pt,
      },
      left=10pt,
      right=10pt,
      top=13pt,
      bottom=9pt,
      boxsep=0pt,
      before skip=0.76\baselineskip,
      after skip=0.76\baselineskip,
      borderline west={2.2pt}{0pt}{highlightaccent},
    },
}
\newtcolorbox{HighlightBox}[2][]{highlightbox,title={#2},#1}

%% file: preamble/header.tex
\usepackage{fontawesome}
\usepackage[T1]{fontenc}
\usepackage{capt-of}
\usepackage[font=small,labelfont=bf]{caption}
\usepackage{pgf}

\usepackage{titletoc}

\providecommand{\acknowledgments}[1]{\section*{Acknowledgments}#1}

\usepackage{natbib}
\usepackage{scalefnt}
\usepackage{graphicx}
\usepackage{amssymb,upgreek,bm}
\usepackage{mathtools}
\usepackage[english]{babel}  
\usepackage{multirow,tcolorbox,tikz,tikz-cd,turnstile,xspace,xparse}
\usepackage{array,relsize,breakcites,appendix}
\usepackage{longtable,tablefootnote,booktabs,float,makecell}
\usepackage{siunitx}
\usepackage[normalem]{ulem}
\usepackage{tabu}
\usepackage[shortlabels]{enumitem} 
\usepackage{footnote}
\usepackage{boxedminipage}
\usepackage{multirow,nicefrac}
\usepackage{verbatim} % gives the comment environment

\floatstyle{ruled}
\newfloat{algorithm}{t}{loa}
\floatname{algorithm}{Algorithm}

\usepackage[
    colorlinks=true,
    linkcolor=\linkcolor,
    urlcolor=\urlcolor,
    citecolor=\citecolor,
    breaklinks,
    pagebackref
]{hyperref}

\usepackage{prettyref}

\usepackage[capitalise,nameinlink]{cleveref}
\crefname{figure}{Fig.}{Figs.}
\Crefname{figure}{Fig.}{Figs.}
\crefname{table}{Table}{Tables}
\Crefname{table}{Table}{Tables}
\crefname{algorithm}{Algo.}{Algos.}
\Crefname{algorithm}{Algo.}{Algos.}
\crefname{section}{Sec.}{Secs.}
\Crefname{section}{Sec.}{Secs.}
\crefname{appendix}{Appendix}{Appendices}
\Crefname{appendix}{Appendix}{Appendices}

\iftoggle{colt}{
    \DeclareRobustCommand{\qed}{
        \usepackage{thmtools}
          \ifmmode \mathqed
          \else
            \leavevmode\unskip\penalty9999 \hbox{}\nobreak\hfill
            \quad\hbox{\qedsymbol}%
          \fi
    }
}
{
    \usepackage{amsthm}
     
}
\usepackage{thm-restate}

\iftoggle{arxiv}
{
    \usepackage{mathrsfs}
    \usepackage[
        a4paper,
        top=1in,
        bottom=1in,
        left=1in,
        right=1in]{geometry}
    \usepackage[bitstream-charter]{mathdesign}
    \DeclareMathAlphabet{\mathcalcm}{OMS}{cmsy}{m}{n}
    \SetMathAlphabet{\mathcalcm}{bold}{OMS}{cmsy}{b}{n}
    \let\mathcal\mathcalcm
     
    \usepackage[scaled=0.92]{PTSans}

    \usepackage{subfig}

    \usepackage{fullpage}
   
}
{
    \usepackage{mathrsfs}
}

\DeclareMathAlphabet{\mathbfsf}{\encodingdefault}{\sfdefault}{bx}{n}

\numberwithin{equation}{section}

\iftoggle{arxiv}
{
    }
{
    
}
\iftoggle{arxiv}
{
    
}
{
    
}

%% file: preamble/theorem_styles.tex
\usepackage{thmtools}

\Crefname{equation}{Eq.}{Eqs.}
\Crefname{assumption}{Assumption}{Assumptions}
\Crefname{condition}{Condition}{Conditions}
\Crefname{claim}{Claim}{Claims}
\Crefname{property}{Property}{Properties}
\Crefname{construction}{Construction}{Constructions}

\declaretheoremstyle[
    headformat=\normalfont\textcolor{\thmcolordark}{\bfseries\NAME\,\NUMBER}\NOTE,%
    notefont={\normalfont\textcolor{\thmcolordark}{\bfseries}}, 
    notebraces={}{},
    bodyfont=\normalfont\itshape,
    spaceabove = 6pt,
    spacebelow = 6pt,
    ]{coloredthmversion}

\declaretheoremstyle[
    headformat=\normalfont\textcolor{\thmcolordark}{\bfseries\NAME\,\NUMBER}\NOTE,%
    bodyfont=\normalfont\itshape,
    spaceabove = 6pt,
    spacebelow = 6pt,
    ]{coloredthm}

\declaretheoremstyle[
    headformat=\normalfont\textcolor{\thmcolordark}{\bfseries\NAME\,\NUMBER}\NOTE,%
    bodyfont=\normalfont,
    spaceabove = 6pt,
    spacebelow = 6pt,
    ]{coloreddef}

\iftoggle{customthms}
{
    \theoremstyle{coloredthmversion}
}
{}

\iftoggle{customthms}{
  \theoremstyle{coloredthm}
  \newtheorem{theorem}{Theorem}
  \newtheorem{lemma}{Lemma}[section]
  \newtheorem{corollary}{Corollary}[section]
  \newtheorem{proposition}[lemma]{Proposition}
}
{}

\newtheorem*{thminformal*}{Informal Theorem}

\iftoggle{customthms}{
    \theoremstyle{coloreddef}
    \newtheorem{definition}{Definition}[section]

    \newtheorem{property}{Property}[section]
}
{
  
}

\newtheorem{assumption}{Assumption}[section]
\newtheorem{condition}{Condition}[section]

\makeatletter
\newcommand{\neutralize}[1]{\expandafter\let\csname c@#1\endcsname\count@}
\makeatother

\iftoggle{customthms}{
    \newtheoremstyle{named}{}{}{\itshape}{}{\bfseries}{}{.5em}{\Cref{#3} {\normalfont (informal)} }{}
    \theoremstyle{named}
    
    \theoremstyle{plain}
}
{}

\newtheorem*{theorem*}{Theorem}
\newtheorem*{lemma*}{Lemma}
\newtheorem*{corollary*}{Corollary}
\newtheorem*{proposition*}{Proposition}
\newtheorem*{claim*}{Claim}
\newtheorem*{fact*}{Fact}
\newtheorem*{observation*}{Observation}
\newtheorem*{definition*}{Definition}
\newtheorem*{remark*}{Remark}
\newtheorem*{example*}{Example}

%% file: preamble/characters.tex
\def\x{\mathbf{x}}
\def\y{\mathbf{y}}

\def\ddefloop#1{\ifx\ddefloop#1\else\ddef{#1}\expandafter\ddefloop\fi}
\def\ddef#1{\expandafter\def\csname bb#1\endcsname{\ensuremath{\mathbb{#1}}}}
\ddefloop ABCDEFGHIJKLMNOPQRSTUVWXYZ\ddefloop

\def\ddefloop#1{\ifx\ddefloop#1\else\ddef{#1}\expandafter\ddefloop\fi}
\def\ddef#1{\expandafter\def\csname frak#1\endcsname{\ensuremath{\mathfrak{#1}}}}
\ddefloop ABCDEFGHIJKLMNOPQRSTUVWXYZ\ddefloop

\def\ddefloop#1{\ifx\ddefloop#1\else\ddef{#1}\expandafter\ddefloop\fi}
\def\ddef#1{\expandafter\def\csname fr#1\endcsname{\ensuremath{\mathfrak{#1}}}}
\ddefloop ABCDEFGHIJKLMNOPQRSTUVWXYZ\ddefloop

\def\ddefloop#1{\ifx\ddefloop#1\else\ddef{#1}\expandafter\ddefloop\fi}
\def\ddef#1{\expandafter\def\csname eul#1\endcsname{\ensuremath{\EuScript{#1}}}}
\ddefloop ABCDEFGHIJKLMNOPQRSTUVWXYZ\ddefloop

\def\ddefloop#1{\ifx\ddefloop#1\else\ddef{#1}\expandafter\ddefloop\fi}
\def\ddef#1{\expandafter\def\csname scr#1\endcsname{\ensuremath{\mathscr{#1}}}}
\ddefloop ABCDEFGHIJKLMNOPQRSTUVWXYZ\ddefloop

\def\ddefloop#1{\ifx\ddefloop#1\else\ddef{#1}\expandafter\ddefloop\fi}
\def\ddef#1{\expandafter\def\csname b#1\endcsname{\ensuremath{\mathbf{#1}}}}
\ddefloop ABCDEFGHIJKLMNOPQRSTUVWXYZ\ddefloop

\def\ddefloop#1{\ifx\ddefloop#1\else\ddef{#1}\expandafter\ddefloop\fi}
\def\ddef#1{\expandafter\def\csname bhat#1\endcsname{\ensuremath{\hat{\mathbf{#1}}}}}
\ddefloop ABCDEFGHIJKLMNOPQRSTUVWXYZ\ddefloop

\def\ddefloop#1{\ifx\ddefloop#1\else\ddef{#1}\expandafter\ddefloop\fi}
\def\ddef#1{\expandafter\def\csname btil#1\endcsname{\ensuremath{\tilde{\mathbf{#1}}}}}
\ddefloop ABCDEFGHIJKLMNOPQRSTUVWXYZ\ddefloop

\def\ddefloop#1{\ifx\ddefloop#1\else\ddef{#1}\expandafter\ddefloop\fi}
\def\ddef#1{\expandafter\def\csname bst#1\endcsname{\ensuremath{\mathbf{#1}^\star}}}
\ddefloop ABCDEFGHIJKLMNOPQRSTUVWXYZ\ddefloop

\def\ddefloop#1{\ifx\ddefloop#1\else\ddef{#1}\expandafter\ddefloop\fi}
\def\ddef#1{\expandafter\def\csname bst#1\endcsname{\ensuremath{\mathbf{#1}^\star}}}
\ddefloop abcdeghijklmnopqrstuvwxyz\ddefloop

\def\ddefloop#1{\ifx\ddefloop#1\else\ddef{#1}\expandafter\ddefloop\fi}
\def\ddef#1{\expandafter\def\csname bhat#1\endcsname{\ensuremath{\hat{\mathbf{#1}}}}}
\ddefloop abcdefghijklmnopqrstuvwxyz\ddefloop

\def\ddefloop#1{\ifx\ddefloop#1\else\ddef{#1}\expandafter\ddefloop\fi}
\def\ddef#1{\expandafter\def\csname b#1\endcsname{\ensuremath{\mathbf{#1}}}}
\ddefloop abcdeghijklnopqrstuvwxyz\ddefloop

\def\ddefloop#1{\ifx\ddefloop#1\else\ddef{#1}\expandafter\ddefloop\fi}
\def\ddef#1{\expandafter\def\csname barb#1\endcsname{\ensuremath{\bar{\mathbf{#1}}}}}
\ddefloop abcdefghijklmnopqrstuvwxyz\ddefloop

\def\ddef#1{\expandafter\def\csname c#1\endcsname{\ensuremath{\mathcal{#1}}}}
\ddefloop ABCDEFGHIJKLMNOPQRSTUVWXYZ\ddefloop
\def\ddef#1{\expandafter\def\csname h#1\endcsname{\ensuremath{\widehat{#1}}}}
\ddefloop ABCDEFGHIJKLMNOPQRSTUVWXYZ\ddefloop
\def\ddef#1{\expandafter\def\csname hc#1\endcsname{\ensuremath{\widehat{\mathcal{#1}}}}}
\ddefloop ABCDEFGHIJKLMNOPQRSTUVWXYZ\ddefloop
\def\ddef#1{\expandafter\def\csname t#1\endcsname{\ensuremath{\widetilde{#1}}}}
\ddefloop ABCDEFGHIJKLMNOPQRSTUVWXYZ\ddefloop
\def\ddef#1{\expandafter\def\csname tc#1\endcsname{\ensuremath{\widetilde{\mathcal{#1}}}}}
\ddefloop ABCDEFGHIJKLMNOPQRSTUVWXYZ\ddefloop

%% file: preamble/general_macros.tex
\usepackage{tcolorbox}
\tcbuselibrary{skins,breakable}

\tcbset{
  aibox/.style={
      width=\linewidth,
      top=6pt,
      bottom=0pt,
      left=1.5pt,
      right=1.5pt,
colback=blue!60!gray!5,
colframe=black,
      colbacktitle=black,
      enhanced,
      center,
      attach boxed title to top left={yshift=-0.1in,xshift=0.15in},
      boxed title style={boxrule=0pt,colframe=white,},
    }
}

\let\Pr\relax
\DeclareMathOperator{\Pr}{\mathbb{P}}

\newcommand{\ballkr}[1][r]{\cB_{k}(r)}

\DeclareMathSymbol{\shortminus}{\mathbin}{AMSa}{"39}

%% file: preamble/specific_macros.tex
\makeatletter
\DeclareRobustCommand{\floatcite}[1]{%
  [\begingroup
  \def\floatcite@sep{}%
  \@for\floatcite@key:=#1\do{%
    \floatcite@sep\citenum{\floatcite@key}%
    \def\floatcite@sep{, }%
  }%
  \endgroup]%
}
\makeatother

%% file: preamble/algnames.tex
\definecolor{oatmaroon}{HTML}{8A1538}
\definecolor{oatplum}{HTML}{7A4EAB}

\newcommand{\methodfont}[1]{\textcolor{oatmaroon}{\texttt{#1}}}
\NewDocumentCommand{\oatfamily}{o}{\ensuremath{\mbox{\methodfont{OAT}}\IfValueT{#1}{_{#1}}}\xspace}
\NewDocumentCommand{\oat}{o}{\ensuremath{\mbox{\methodfont{OAT}}^{\mathrm{sing}}\IfValueT{#1}{_{#1}}}\xspace}
\NewDocumentCommand{\oatpowtwo}{o}{\ensuremath{\mbox{\methodfont{OAT}}^{\mathrm{pow2}}\IfValueT{#1}{_{#1}}}\xspace}

\newcommand{\oatlong}{Ordered Action Tokenization\xspace}

\newcommand{\bin}{\texttt{Bin}\xspace}
\newcommand{\fast}{\texttt{FAST}\xspace}
\newcommand{\quest}{\texttt{QueST}\xspace}
\newcommand{\acodec}{\texttt{ACodec}\xspace}
\newcommand{\BAR}{\texttt{BAR}\xspace}
\newcommand{\ctrltransformer}{\texttt{Transformer}\xspace}

\newcommand{\paligemma}{\texttt{PaliGemma2}\xspace}
\newcommand{\qwenvl}{\texttt{Qwen3VL}\xspace}
\newcommand{\benchmarkfont}[1]{{\fontfamily{zi4}\selectfont #1}}
\newcommand{\libero}{\benchmarkfont{LIBERO}\xspace}
\newcommand{\robomimic}{\benchmarkfont{RoboMimic}\xspace}
\newcommand{\metaworld}{\benchmarkfont{MetaWorld}\xspace}
\newcommand{\robocasa}{\benchmarkfont{RoboCasa}\xspace}
\newcommand{\robocasaThreeSixFive}{\benchmarkfont{RoboCasa365}\xspace}
\newcommand{\simpler}{\benchmarkfont{SimplerEnv}\xspace}

\newcommand{\propertyI}{\textcolor{oatplum}{P.1}\xspace}
\newcommand{\propertyII}{\textcolor{oatplum}{P.2}\xspace}
\newcommand{\propertyIII}{\textcolor{oatplum}{P.3}\xspace}
\newcommand{\tokyes}{\textcolor{green!50!black}{\(\checkmark\)}}
\newcommand{\tokno}{\textcolor{oatmaroon}{\(\times\)}}

%% file: preamble/comment_macros.tex
\addauthor{cl}{gray}
\addauthor{lz}{blue}

\newcommand{\ignore}[1]{}

%% file: body/abstract.tex
% !TEX root = ../main.tex

Action tokenization maps continuous robot action chunks to discrete tokens and
has become an important interface for modern visuomotor policies. Existing
approaches either rely on analytical discretization methods that produce
prohibitively long token sequences or learned latent tokenizers that lack
structure, limiting their compatibility with downstream policies. In this work,
we identify three desiderata for action tokenization -- high compression, total
decodability, and an ordered token space -- and introduce \oatlong
(\oatfamily), a learned action tokenizer that satisfies all three. \oatfamily
discretizes action chunks into an ordered sequence of tokens using a transformer
with registers, finite scalar quantization, and ordering-inducing training
mechanisms. 
By training each token prefix to decode into a valid action chunk, \oatfamily
places coarse control information in early tokens and uses later tokens to refine
residual detail, yielding an anytime tradeoff between inference cost and action
fidelity.
We validate \oatfamily in two prevailing uses of action tokens: autoregressive
policies that generate tokens for control, and token co-training policies that
use token losses to shape the vision-language model context consumed by a
flow-based action expert.
Across three policy backbones and more than 60 tasks spanning five simulation
benchmarks and real-world settings, \oatfamily consistently delivers strong
policy performance while offering significantly greater flexibility at
inference time.

%% file: body/introduction.tex
% !TEX root = ../main.tex

\section{Introduction}

Action tokens are the interface between continuous robot control
and sequence modeling, yet the design of this interface remains under-examined.
Experience from language and vision shows that tokenization shapes learning dynamics,
model capacity, scalability, and downstream
performance~\citep{sennrich2016neural,oord2017vq,mentzer2024fsq,bachmann2025flextok}.
Likewise, in robot control, action tokenization is not merely offline compression:
tokens determine the output length, the
validity of arbitrary policy samples, and how control-relevant information is organized for prediction or supervision.
Therefore, tokens must be designed with the policy interface in mind, so that they are easy for
policies to predict or learn from.
This raises a basic question: what properties should an action tokenizer satisfy
to serve as a policy interface?

In this paper, we study \emph{what constitutes good action tokens} and argue that an effective action tokenizer must simultaneously satisfy three key desiderata: high compression, total decodability, and ordered token structure.
\textbf{(1) High compression} keeps token sequences compact.
\textbf{(2) Total decodability} ensures that arbitrary policy outputs map to executable actions when tokens
are decoded.
\textbf{(3) Ordered structure} places control-relevant information early, implicitly forming a coarse-to-fine representation.

Prior action tokenization methods satisfy subsets of these desiderata, but not all simultaneously.
Per-dimension binning (\bin) is simple and reliably decodable, but it serializes every action dimension at every step, producing long token sequences as action dimension and prediction horizon grow~\citep{rt12022arxiv,rt22023arxiv,kim24openvla}.
Frequency-domain tokenizers such as \fast introduce a useful low-to-high-frequency order, but byte-pair encoding compression makes detokenization only partially defined: an unconstrained policy sample is not
guaranteed to expand into the fixed-shape coefficient array required for control~\citep{gage1994bpe,sennrich2016neural,pertsch2025fast}.
Learned latent tokenizers such as \quest and \acodec compress action chunks through discrete bottlenecks~\citep{mete2024quest,lee2024behaviorgenwithlatentactions,dong2026actioncodecmakesgoodaction},
but the reconstruction quality they optimize for does not necessarily improve closed-loop policy rollout success at inference.

To bridge this gap, we propose \oatlong (\oatfamily), a learned tokenizer that
discretizes continuous action chunks into compact, totally decodable, and
ordered token sequences. \oatfamily employs transformer-based register tokens to
aggregate temporal information, finite scalar quantization to construct a
discrete bottleneck, and nested dropout to train prefixes at multiple budgets to
decode into plausible action chunks. This prefix training implicitly induces a coarse-to-fine
structure: early tokens capture high-impact control information, while later
tokens refine residual detail.

We validate the effectiveness of \oatfamily in two prevailing uses of action tokens.
In autoregressive policies, \oatfamily tokens are generated and detokenized into actions; their
ordering provides an inductive bias aligned with next-token prediction and
supports variable inference budgets through prefix-based decoding. In 
token co-training policies, action token losses supervise the vision-language
model (VLM), whose context conditions a flow-matching action expert at inference.
Because \oatfamily trains its one-token prefix to reconstruct the complete
action chunk, predicting the first target directly from the VLM prefill context
imposes a plan-like, action-chunk-level objective on the representation consumed
by the expert. \cref{sec:prelim} gives the background for both policy roles.

For scalable autoregressive inference with \oatfamily, we further introduce a
scheduling framework for block autoregressive decoding that unifies token-wise
autoregression, one-shot parallel
decoding, fixed-size block prediction~\citep{liu2025fasterefficientautoregressivevision,dong2026actioncodecmakesgoodaction},
and intermediate schemes. We focus on two variants in this paper: \oat for
token-wise autoregression and \oatpowtwo for power-of-two block decoding; the
latter reduces policy-call complexity from linear to logarithmic in the token
horizon.

\begin{figure}[t]
    \centering
    \begin{tabular}{ccccc}
      \includegraphics[width=0.17\linewidth]{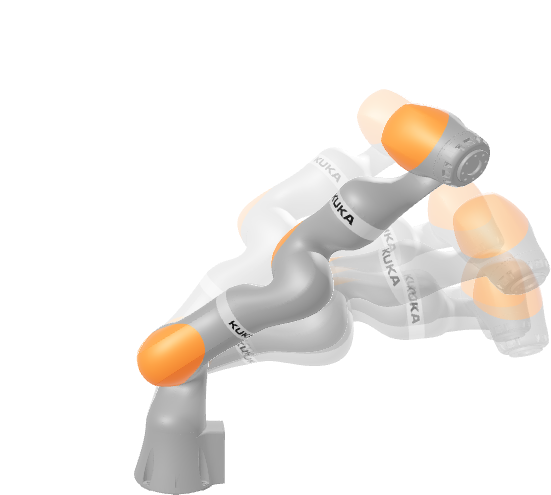} &
      \includegraphics[width=0.17\linewidth]{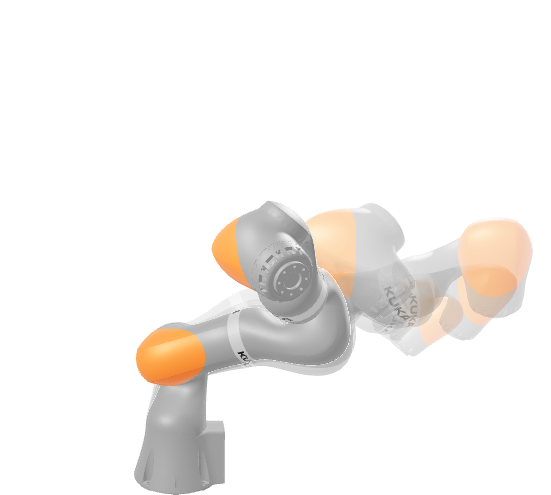} &
      \includegraphics[width=0.17\linewidth]{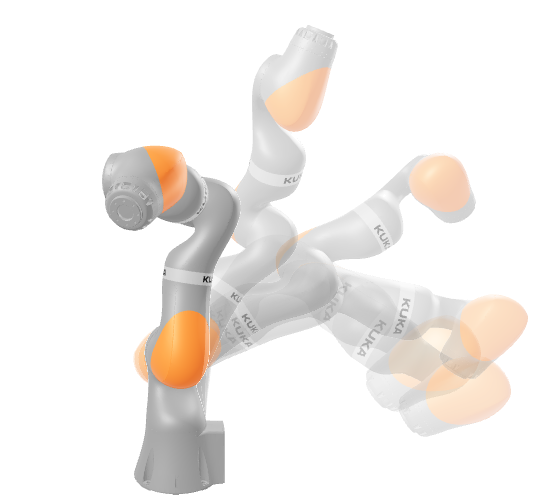} &
      \includegraphics[width=0.17\linewidth]{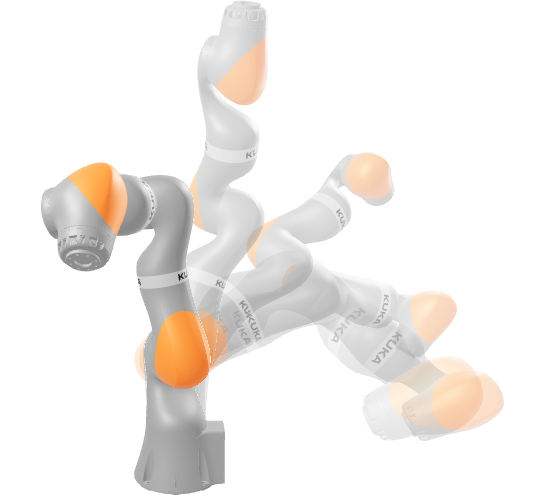} &
      \includegraphics[width=0.17\linewidth]{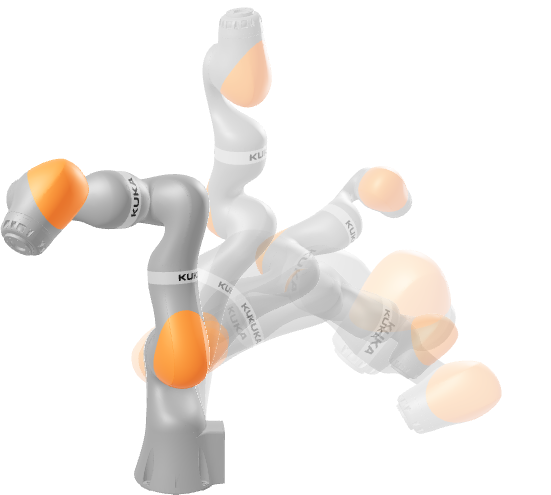} \\
      \footnotesize \oatfamily[1] & \footnotesize \oatfamily[2] &
      \footnotesize \oatfamily[4] & \footnotesize \oatfamily[8] &
      \footnotesize Ground truth
    \end{tabular}
    \caption{\textbf{Prefix reconstruction on an iiwa arm.} Columns show
    action chunk reconstructions decoded from the first 1, 2, 4, and 8
    \oatfamily{} tokens, followed by the ground-truth action chunk. Increasing the prefix
    budget progressively refines the reconstructed trajectory while every
    mask-padded prefix detokenizes to an executable action chunk. See the
    \href{https://ordered-action-tokenization.github.io/\#prefix-lab}{interactive
    prefix lab} on the project website.}
    \label{fig:oat_prefix_reconstruction}
\end{figure}

\myparagraph{Contributions.}
In summary, this paper makes three contributions, as illustrated in \cref{fig:oat_teaser}:
\begin{enumerate}[leftmargin=1.6em,itemsep=0.08em,topsep=0.25em]
\item We analyze representative action tokenizers as policy interfaces and
formalize three desiderata for visuomotor policy learning: high compression,
total decodability, and ordered structure.

\item We propose \oatfamily, a learned action tokenizer that satisfies these
desiderata with compact, totally decodable, ordered tokens whose prefixes decode
to executable action chunks, and further introduce a framework for scalable
block autoregressive decoding.

\item We conduct extensive experiments and ablations showing that \oatfamily is
effective across lightweight policies and vision-language-action (VLA)-scale
systems, covering both autoregressive (AR) policies that generate action tokens and
token co-training (TC) policies that use action tokens as supervision.
\end{enumerate}

%% file: body/prelim.tex
% !TEX root = ../main.tex

\section{Preliminaries}
\label{sec:prelim}

We first define notation and background used throughout the paper.

\myparagraph{Action chunks and tokens.}
Robot policies commonly execute control through short chunks of continuous
actions. We write one action chunk as
\[
    A = a_{1:H_a}\in\mathbb{R}^{H_a\times D_a},
\]
where $H_a$ is the action horizon and $D_a$ is the action dimension.
Tokenized autoregressive policies represent this continuous chunk as a discrete sequence of policy
symbols~\citep{rt12022arxiv,rt22023arxiv,kim24openvla,bonatti2022pactperceptionactioncausaltransformer,fu2024incontextimitationlearningnexttoken}.
For a tokenizer with token horizon $H_l$, the action tokenizer is
\[
    \mathcal{T}:a_{1:H_a}\mapsto T_{1:H_l},
    \qquad
    T_i\in\mathcal{V}.
\]
Here $\mathcal{V}$ is the action token vocabulary. A corresponding detokenizer,
denoted by $\mathcal{T}^{-1}$, maps a token sequence back to a
continuous action chunk,
\[
    \mathcal{T}^{-1}:T_{1:H_l}\mapsto\hat{a}_{1:H_a}.
\]

\myparagraph{Token-wise autoregressive policies.}
Let $o_{1:H_o}$ denote the observation history available to the policy, with
observation horizon $H_o$. The standard token-wise autoregressive policy models the action
token sequence left to right:
\[
    p_\pi(T_{1:H_l}\mid o_{1:H_o})
    =
    \prod_{i=1}^{H_l}p_\pi(T_i\mid T_{<i},o_{1:H_o}).
\]
After sampling $T_{1:H_l}$, the policy detokenizes it with $\mathcal{T}^{-1}$
and executes the resulting chunk, typically using receding-horizon
control~\citep{zhao2023actpolicy,chi2024diffusionpolicy,zhang2025actionchunkingexploratorydata}.
\cref{sec:bar} introduces block autoregressive decoding, which keeps this
tokenizer--detokenizer interface but groups token positions during generation
instead of generating one position at a time.

\myparagraph{Token co-training policies.}
We use \emph{token co-training} to denote a VLA training setup that separates
token-based VLM supervision from continuous action
prediction~\citep{driess2025knowledge,intelligence2025pi05vla,fang2026molmoact2actionreasoningmodels,dong2026actioncodecmakesgoodaction}.
During training, an action token loss supervises the VLM, while a flow-matching
loss trains a separate action expert conditioned on detached VLM
context~\citep{driess2025knowledge,fang2026molmoact2actionreasoningmodels,dong2026actioncodecmakesgoodaction}. Its
objective has the schematic form
\[
    \mathcal{L}_{\mathrm{TC}}
    =
    \mathcal{L}_{\mathrm{tok}}
    +
    \lambda\mathcal{L}_{\mathrm{flow}}.
\]
Here $\lambda$ weights the flow-matching objective relative to the token
prediction objective.
At inference, the VLM is prefilled once, its action token logits are discarded,
and the expert generates continuous action chunks from the cached context.
Action tokens therefore serve as supervision targets rather than being decoded
into actions for execution. \Cref{sec:tc} and
\cref{appendix:tc} provide details about this paradigm.

%% file: body/action_tokenization.tex
% !TEX root = ../main.tex

\section{Action Tokenization as a Policy Interface}
\label{sec:action_token_prelim}

Following the notation in \cref{sec:prelim}, this section analyzes what makes
an action tokenizer a useful policy interface. The tokenizer determines the
discrete targets presented to the policy, how many targets it must model,
whether arbitrary generated sequences can be detokenized into valid actions,
and how token structure interacts with the policy objective.

\subsection{Tokenizer Desiderata}

Rate and distortion provide a general lens for lossy
compression~\citep{shannon1948maththeoryofcommunication,blau2019rethinkinglossycomprratedistortionpercep}
and are widely used to analyze learned discrete
representations~\citep{tschannen2018recentadvvaerepr,oord2017vq,mentzer2024fsq,bachmann2025flextok,zhao2025npq}.
For action tokenization, compression remains a policy requirement because long
token sequences increase the number of prediction targets and, for autoregressive policies,
generation depth. Policy learning also adds requirements that compression alone
does not capture: sampled tokens must decode reliably, and token order should
expose control structure that helps policies generate tokens or learn from token
supervision.
% Block-wise generation adds a separate compatibility question for
% scalable AR inference rather than a fourth core desideratum.

\myparagraph{\propertyI High compression.}
The token horizon $H_l$ should be small relative to the raw action size
$H_aD_a$. Long token sequences increase training difficulty and, for autoregressive
policies, inference latency because the policy must model more token targets and
generate more tokens. A suitable tokenizer keeps the rate low while preserving
motion relevant to control.

\myparagraph{\propertyII Total decodability.}
The detokenizer should be a total function over the policy's discrete output
space. When tokens are decoded into control, a policy can emit any token
sequence supported by its output distribution. The detokenizer must therefore
map arbitrary policy samples, not only training codes produced by the encoder,
to valid continuous actions.

\myparagraph{\propertyIII Ordered token structure.}
Reconstruction error alone does not guarantee useful policy targets. An ordered
token structure places control-relevant information early and leaves residual
detail to later tokens. For autoregressive policies, this creates learnable generation
targets that align with the inductive bias of next-token prediction. For token co-training policies, an ordered representation can make the
first supervised target describe global action-chunk structure rather than a
single coordinate or a latent without an explicit global role. Many such orders
are possible; for example, a
coarse-to-fine order places high-impact motion early and leaves later tokens to
refine residual detail.

\begin{table}[t]
  \centering
  \footnotesize
  \setlength{\tabcolsep}{2.8pt}
  \renewcommand{\arraystretch}{1.12}
  \providecommand{\tokna}{}
  \renewcommand{\tokyes}{\textcolor{green!50!black}{\large\(\checkmark\)}}
  \renewcommand{\tokno}{\textcolor{oatmaroon}{\large\(\times\)}}
  \renewcommand{\tokna}{\textcolor{gray!70!black}{\large\(\text{--}\)}}
  \begin{tabular*}{\linewidth}{@{\hspace{0.55em}\extracolsep{\fill}}>{\raggedright\arraybackslash}m{0.118\linewidth}|>{\centering\arraybackslash}m{0.083\linewidth}>{\centering\arraybackslash}m{0.103\linewidth}>{\centering\arraybackslash}m{0.125\linewidth}>{\centering\arraybackslash}m{0.095\linewidth}|>{\raggedright\arraybackslash}m{0.389\linewidth}@{\hspace{0.55em}}}
    \toprule
    \multicolumn{1}{c|}{\multirow{2}{*}{Scheme}} &
    \multicolumn{1}{c}{\multirow{2}{*}{Compact}} &
    \multicolumn{1}{c}{Total} &
    \multicolumn{1}{c}{Ordered} &
    \multicolumn{1}{c|}{Block} &
    \multicolumn{1}{c}{\multirow{2}{*}{Policy implication}} \\
    \multicolumn{1}{c|}{} &
    \multicolumn{1}{c}{} &
    \multicolumn{1}{c}{decodable} &
    \multicolumn{1}{c}{structure} &
    \multicolumn{1}{c|}{compat.} &
    \multicolumn{1}{c}{} \\
    \midrule
    \bin~\floatcite{rt12022arxiv,rt22023arxiv}
      & \tokno & \tokyes & \tokno & \tokyes
      & {\scriptsize Long coordinate sequences raise inference cost and give
        weak conditioning structure.} \\
    \fast~\floatcite{pertsch2025fast}
      & \tokyes & \tokno & \tokyes & \tokno
      & {\scriptsize Unconstrained samples may not produce the fixed-shape
        coefficient array required for decoding.} \\
    \quest~\floatcite{mete2024quest}
      & \tokyes & \tokyes & \tokno & \tokno
      & {\scriptsize A reconstruction loss need not place control-relevant
        information early.} \\
    \acodec~\floatcite{dong2026actioncodecmakesgoodaction}
      & \tokyes & \tokyes & \tokna & \tokyes
      & {\scriptsize Parallel decoding does not require serial ordering,
        but the full latent block is a hard joint target.} \\
    \oat~\floatcite{liu2026orderedactiontokenization}
      & \tokyes & \tokyes & \tokyes & \tokno
      & {\scriptsize Causal register ordering supports token-wise prediction,
        but is not aligned with grouped block prediction.} \\
    \oatpowtwo
      & \tokyes & \tokyes & \tokyes & \tokyes
      & {\scriptsize Block-causal register ordering makes each token group
        jointly predictable.} \\
    \bottomrule
  \end{tabular*}
  \caption{\textbf{Action tokenizer comparison.} Rows compare representative
  action tokenizers as policy interfaces. The first three property columns
  indicate compactness, total decodability, and ordered structure. Block compat.
  records a side property for scalable grouped autoregressive inference, while
  Policy implication summarizes the downstream consequence. A dash marks
  one-shot parallel decoding, for which serial ordered structure is not
  applicable.}
  \label{tab:tokenizer_comparison}
\end{table}

\myparagraph{Block compatibility.}
Block-wise generation is not a core tokenizer desideratum, but it is useful for
scalable autoregressive inference. As formalized in \cref{sec:bar}, a block
autoregressive schedule can ask the policy to emit several new tokens at the
same prefix budget in a single call~\citep{stern2018blockwiseparalleldecodingdeep}. Those tokens must be
jointly predictable from the observation, the realized prefix, and prediction
masks, without relying on realized within-block tokens as serial inputs. A
compact, total, ordered tokenizer can still fail this property if its
token structure was trained only as a strict token-wise chain.
\cref{tab:tokenizer_comparison} summarizes how the tokenizers discussed in the
paper satisfy these interface properties; the final rows show the token-wise and
power-of-two \oatfamily{} variants defined in \cref{sec:oat_instantiations}.

\subsection{Where Existing Tokenizers Fall Short}

Existing action tokenizers satisfy different parts of this interface, but no
standard baseline satisfies the three core desiderata while also providing
block compatibility as shown in \cref{tab:tokenizer_comparison}.
Per-dimension \bin is reliable because every generated bin index maps back to a
scalar action value, and coordinate groups can be decoded into valid actions.
Its limitations are high rate and weak ordering: the token horizon grows with
$H_aD_a$, and the
manual coordinate serialization does not place coarse trajectory information
early in the sequence.

\fast addresses rate and ordering by representing action trajectories through
frequency-domain coefficients and compressing the coefficient stream with
byte-pair encoding (BPE). Low-frequency components appear before high-frequency
components, so early tokens tend to describe coarse motion. During
detokenization, each BPE token expands into a variable-length coefficient
subsequence, whereas the inverse frequency transform expects a fixed
coefficient topology. Detokenizing unconstrained policy samples may therefore
fail or require padding, truncation, rejection, or constrained decoding, each
of which changes the policy interface.

Learned latent tokenizers such as \quest and \acodec decode through bottlenecks
based on vector quantization or finite scalar quantization~\citep{mentzer2024fsq,mete2024quest,dong2026actioncodecmakesgoodaction}.
\quest-style sequential latents are attractive from a rate--distortion
viewpoint, but full-sequence reconstruction alone does not explicitly assign the
first latent a global role over the action chunk. \acodec instead
predicts fixed latent blocks jointly, reducing serial depth but making each
block a harder joint prediction problem.

Thus reconstruction quality alone does not determine whether an action
tokenizer is a good policy interface: the representation must also be compact,
total over policy samples, and ordered for generation and supervision. For
scalable autoregressive inference, compatibility with the intended block
generation pattern is an additional side property. A detailed discussion of these baseline tokenizers is
provided in \cref{appendix:baseline_tokenizers}.

%% file: body/oat.tex
% !TEX root = ../main.tex

\section{\texorpdfstring{\oatfamily: \oatlong}{Ordered Action Tokenization}}
\label{sec:oat_method}

\begin{figure}[t]
    \centering
    \includegraphics[width=\linewidth]{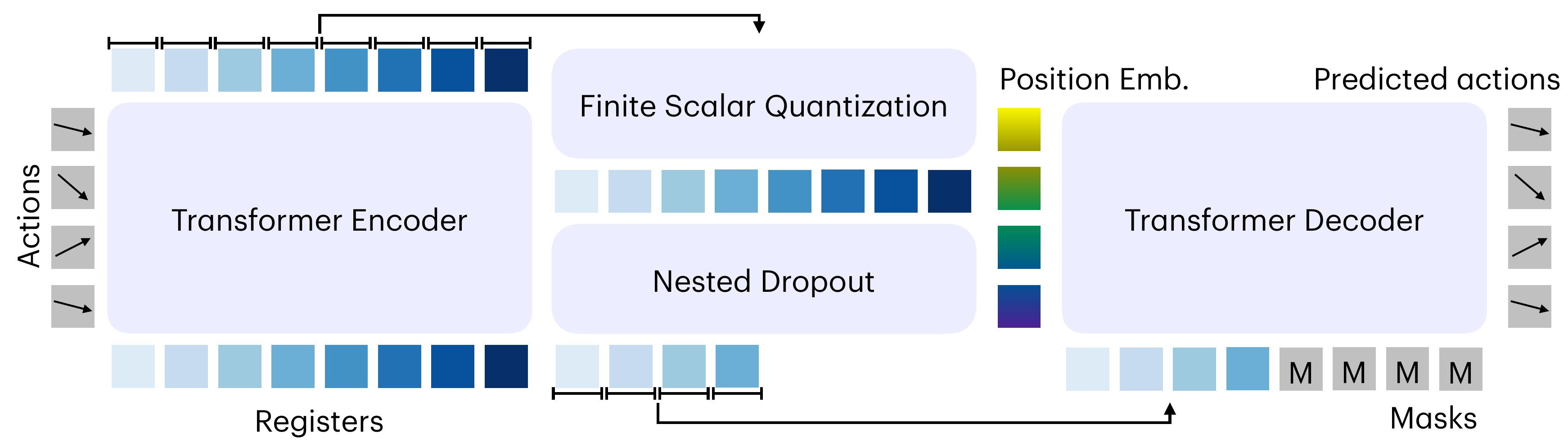}
    \caption{\textbf{\oatfamily tokenizer pipeline.}
    \oatfamily encodes a continuous action chunk into register states,
    discretizes these states with finite scalar quantization, and trains a decoder to
    reconstruct the full action chunk from retained token prefixes. Nested
    dropout samples budgets and replaces suffixes with learned
    masks, forcing early tokens to carry coarse executable control while later
    tokens refine residual detail.}
    \label{fig:oat_method}
\end{figure}

The tokenizer analysis above motivates \oatfamily as a tokenizer for compact,
totally decodable, ordered representations. This section defines the
encoder--quantizer--decoder backbone, the ordered prefix training objective,
and the token-wise and power-of-two variants studied in this paper.

\subsection{Tokenization \texorpdfstring{$\mathcal{T}$}{T} and Detokenization \texorpdfstring{$\mathcal{T}^{-1}$}{invT}}

\oatfamily maps a continuous action chunk to a sequence of discrete action
tokens and decodes token sequences back to continuous control. The tokenizer
$\mathcal{T}$ is instantiated by an encoder $E_\phi$ and a quantization
bottleneck, while the detokenizer $\mathcal{T}^{-1}$ is instantiated by a
decoder $D_\theta$, as summarized in \cref{fig:oat_method}. This autoencoder
architecture ensures total decodability for the policy interface: every
vocabulary index sequence maps to a continuous action chunk.
We next describe the bottleneck design that gives \oatfamily its compact and
ordered token structure.

\myparagraph{Register bottleneck.}
The encoder uses learnable registers $r_{1:H_l}$, inspired by ViT
registers~\citep{darcet2024vitneedsregisters}, to compress a continuous action
chunk $a_{1:H_a}$ into a fixed sequence of register states. Each register
cross-attends~\citep{vaswani2017attnisallyouneed} to all action positions, and
the ordered prefix objective in the next subsection specifies the register
self-attention mask. The encoder returns the register states
$z_{1:H_l}=E_\phi(a_{1:H_a},r_{1:H_l})$.
Each latent $z_i\in\mathbb{R}^{D_l}$, where $D_l$ is the latent register dimension, is
discretized into token $T_i$ using
finite scalar quantization (FSQ)~\citep{mentzer2024fsq}. The resulting token sequence $T_{1:H_l}$ serves as
the token target for policy learning. The FSQ levels determine the vocabulary
size, and $H_l$ determines how many action tokens the policy must predict.

The decoder is a cross-attention Transformer: action position embeddings query
the quantized register states, and the resulting query states are projected to
$H_a$ output action vectors, with no self-attention among the action position
queries.
Next, we discuss training \oatfamily with an ordered prefix objective.
% The ordered prefix objective trains this decoder under masked token inputs.
\subsection{Ordered Prefix Training for Progressive Tokens}

Effective action tokenization requires more than compact reconstruction: the
token sequence should have an order that policies can exploit. Our goal is to
make early tokens capture coarse, globally salient
aspects of an action chunk, while later tokens refine residual details. We use
two complementary mechanisms to induce this ordering and support variable token
budgets.

\myparagraph{Nested Prefix Reconstruction.}
\oatfamily induces order through an increasing set of reconstruction budgets.
Let $\mathcal{K}=\{k_1,\ldots,k_M\}$ denote the trained budget set, with
$0=k_0<k_1<\cdots<k_M=H_l$. During tokenizer training, we sample
$K\sim\mathrm{Uniform}(\mathcal{K})$, retain only the prefix $T_{1:K}$, and replace the
suffix with learned mask tokens $\mathtt{MASK}$ before decoding.
This produces the masked decoder input
\[
    \widetilde{T}^{(K)}_{1:H_l}
    =
    T_{1:K}\oplus
    \langle\mathtt{MASK}\rangle_{K+1:H_l}.
\]
Here $\oplus$ denotes sequence concatenation.
The decoder must reconstruct the full action chunk from this partial code:
\[
    \mathcal{L}_{\mathrm{prefix}}
    =
    \mathbb{E}_{a,K}
    \left[
    \left\|D_\theta(\widetilde{T}^{(K)}_{1:H_l})-a_{1:H_a}\right\|_2^2
    \right].
\]
This is nested dropout over action tokens~\citep{ripple2014learnorderedreprwithnesteddropout,kusupati2022matryoshkareprlearning,cai2025matryoshkamultimodal,bachmann2025flextok}.
Unlike an autoencoder objective trained only on full token sequences, it trains
the decoder to map every sampled prefix to an executable action chunk. These budgets
specify the prefixes that receive direct reconstruction supervision; generated
prefixes at intermediate budgets can also be decoded.
\Cref{algo:oat_tokenizer_training} summarizes this tokenizer training loop.

\begin{figure}[t]
    \centering
    \includegraphics[width=\linewidth]{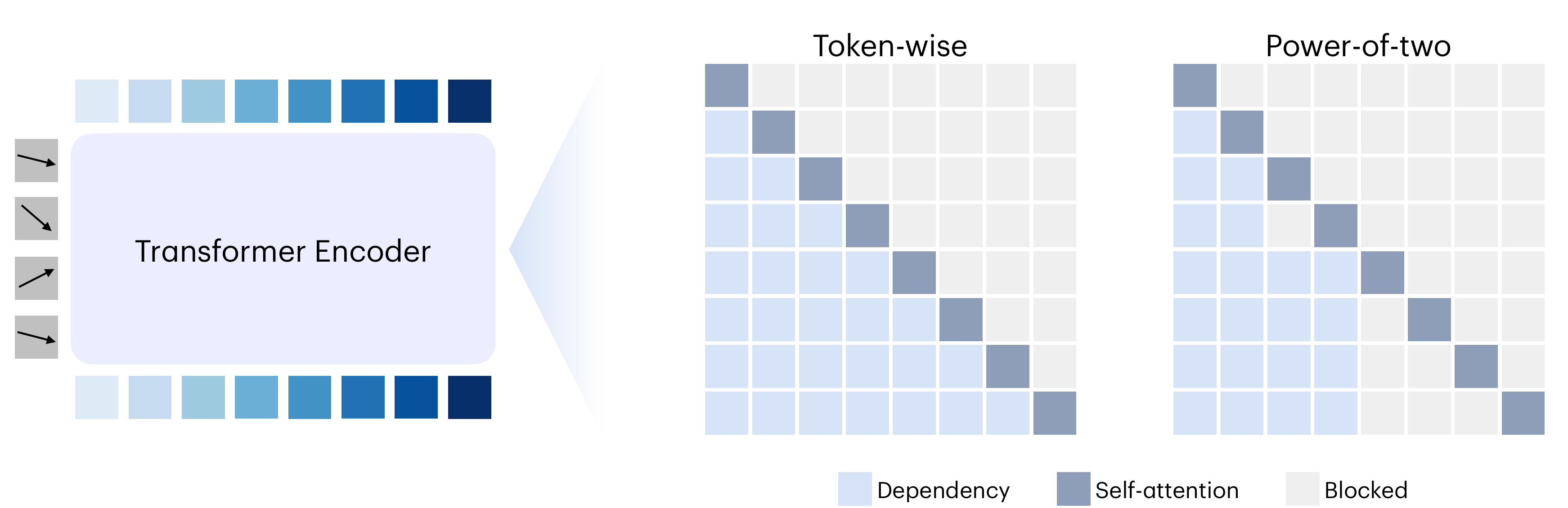}
    \caption{\textbf{\oatfamily encoder attention masks.}
    Each matrix entry at row $i$ and column $j$ indicates whether register $i$
    can attend to register $j$. The token-wise mask gives causal
    register attention. The power-of-two mask preserves attention to
    earlier budget groups and self-attention, while blocking other registers in
    the same new group. The action inputs, registers, quantizer, nested
    dropout, and decoder remain unchanged.}
    \label{fig:oat_attention_masks}
\end{figure}

\myparagraph{Register Flow Constraints.}
The register self-attention mask supplies an architectural ordering constraint
independently of the reconstruction budget set. It determines whether registers
introduced within the same reconstruction interval form a causal chain or have
no direct cross-register dependencies. In
both cases, each register can attend to earlier budget groups and itself. This
separates two aspects of ordering: nested prefix reconstruction determines which
budgets receive direct supervision, while the register mask determines the
dependency structure among token positions. \Cref{fig:oat_attention_masks}
visualizes the two variants studied in the paper, which are detailed in \cref{sec:oat_instantiations}.

\subsection{Progressive Information Allocation}
The two mechanisms above give token ordering a source-coding-inspired
information allocation interpretation. In classical source coding, common source
patterns can be represented with shorter expected
descriptions~\citep{shannon1948maththeoryofcommunication}. Here, the analogue of
code length is the number of retained action tokens. Let $\varepsilon(K)$ denote
the expected reconstruction error when only the first $K$ tokens are retained,
with $\varepsilon(0)$ denoting the all-mask error, and let
$\Delta_i=\varepsilon(i-1)-\varepsilon(i)$ be the marginal gain from retaining
token $i$. The expected nested dropout objective expands as
\[
    \mathbb{E}_{K}[\varepsilon(K)]
    =
    \varepsilon(0)
    -
    \sum_{i=1}^{H_l}
    \Pr(K\geq i)\,\Delta_i .
\]
Thus token $i$ is weighted by its survival probability
$w_i=\Pr(K\geq i)$. The weights are nonincreasing and are equal for positions
introduced between the same pair of successive budgets. Nested prefix
reconstruction therefore prioritizes earlier budget intervals without imposing
an additional order among positions introduced at the same budget. This
pressure favors placing coarse trajectory structure in earlier budgets and
residual refinements in later budgets. The resulting information allocation is
learned rather than assigned to coordinates, timesteps, or frequencies.

The resulting survival weights specify how reconstruction pressure is
distributed over token positions, but do not uniquely determine the register
dependency graph within each budget interval.

\subsection{Token-Wise and Power-of-Two Attention Masks}
\label{sec:oat_instantiations}

We study two representative register attention masks for token-wise and
power-of-two ordering. Both variants use
$\mathcal{K}=\{1,2,4,\ldots,H_l\}$, sample $K$ uniformly from this set, and
otherwise share the same tokenizer architecture and training objective. Other
budget sets and register groupings are possible; \cref{sec:bar} develops the
broader generation schedule space and motivates the power-of-two choice.

\myparagraph{Token-wise ordering.}
\oat uses ordinary causal register attention, imposing a dependency order at
every token position, including positions introduced within the same budget
interval. Intermediate positions can therefore participate in token-wise
generation, although only budgets in $\mathcal{K}$ receive direct reconstruction
supervision.

\myparagraph{Power-of-two ordering.}
\oatpowtwo groups positions introduced between successive reconstruction
budgets and uses the block-causal register mask in
\cref{fig:oat_attention_masks}. For $H_l=16$ in our VLM experiments, the
reconstruction budgets are $1,2,4,8,16$ with $k_0=0$.
Positions introduced at the same budget $k_m$ share the survival weight
$\Pr(K\geq k_m)$, and direct cross-register attention within the group is
blocked. They can therefore be encoded from the action input and earlier groups
without within-group register dependencies, and subsequently treated as one
generation block. \oatpowtwo imposes ordering across budget groups without
introducing an additional order within each group.
The variants therefore differ in the granularity of register dependencies.

Because both variants share the same trained budgets, the information-allocation
view also defines a common diagnostic. For a fixed retained prefix length $K$,
the token budget is a proxy for
rate\footnote{The corresponding code length is proportional to
$K\log_2(|\mathcal{V}|)$, where $\mathcal{V}$ is the token vocabulary.}, and
distortion is the reconstruction error $\varepsilon(K)$ defined above.
Autoencoders trained only on full token sequences optimize only the endpoint
$\varepsilon(H_l)$. \oatfamily instead trains and evaluates this curve at
multiple budgets $K\in\mathcal{K}$. \Cref{fig:oat_prefix_reconstruction} gives a
qualitative example of prefix refinement, and
\cref{fig:rate_distortion_curves} later quantifies whether the learned prefixes
reduce distortion smoothly as the token budget increases. The ordering
ablation in \cref{fig:oat_ordering_ablation_plot} further tests whether this
learned ordering is important for downstream policy performance. We next describe how visuomotor policies use these token structures for
generation and supervision.

%% file: body/oat_policy.tex
% !TEX root = ../main.tex

\begin{figure}[t]
\centering
\begin{minipage}{\linewidth}
\begin{minipage}[t]{0.485\linewidth}
\captionof{algorithm}{\textbf{\oatfamily tokenizer training.} Encode an action
chunk, mask the suffix at a sampled budget, and reconstruct from the prefix.}
\label{algo:oat_tokenizer_training}
\footnotesize
\vspace{-0.25em}
\hrule
\vspace{0.35em}
{\raggedright
\textbf{Input:} dataset $\mathcal{D}$; encoder $E_\phi$; registers $r_{1:H_l}$;
FSQ quantizer; decoder $D_\theta$; mask token $\mathtt{MASK}$; reconstruction budget set
$\mathcal{K}$.\par}
\begin{enumerate}[leftmargin=1.6em,itemsep=0.08em,topsep=0.25em]
    \item \textbf{while} not converged \textbf{do}
    \item \hspace{0.9em}Sample $a_{1:H_a}\sim\mathcal{D}$ and
    $K\sim\mathrm{Uniform}(\mathcal{K})$.
    \item \hspace{0.9em}$z_{1:H_l}\gets E_\phi(a_{1:H_a},r_{1:H_l})$.
    \item \hspace{0.9em}$T_{1:H_l}\gets\mathrm{FSQ}(z_{1:H_l})$.
    \item \hspace{0.9em}$\widetilde{T}\gets
    T_{1:K}\oplus\langle\mathtt{MASK}\rangle_{K+1:H_l}$.
    \item \hspace{0.9em}$\hat{a}_{1:H_a}\gets D_\theta(\widetilde{T})$.
    \item \hspace{0.9em}Update $\{\phi,\theta,r,\mathtt{MASK}\}$ on
    $\lVert \hat{a}_{1:H_a}-a_{1:H_a}\rVert_2^2$.
    \item \textbf{end while}
    \item \textbf{return} $\mathcal{T}$ and $\mathcal{T}^{-1}$.
\end{enumerate}
\end{minipage}\hfill
\begin{minipage}[t]{0.485\linewidth}
\captionof{algorithm}{\textbf{Autoregressive \oatfamily generation.} Generate tokens
up to a target budget, mask the suffix, and return the decoded action chunk.}
\label{algo:oat_policy_inference}
\footnotesize
\vspace{-0.25em}
\hrule
\vspace{0.35em}
{\raggedright
\textbf{Input:} observation history $o_{1:H_o}$; action token policy $\pi$;
$\mathcal{T}^{-1}=\{D_\theta,\mathtt{MASK}\}$; generation endpoint list
$\mathbf{b}=(b_0=0,\ldots,b_m=K)$ with $g_s=b_s-b_{s-1}$.\par}
\begin{enumerate}[leftmargin=1.6em,itemsep=0.08em,topsep=0.25em]
    \item Initialize $\widehat{T}_{1:0}\gets\varnothing$ and $g_0\gets0$.
    \item \textbf{for} $s=1,\ldots,m$ \textbf{do}
    \item \hspace{0.9em}$\widehat{X}^{(s)}\gets
    \widehat{T}_{1:b_{s-1}}\oplus
    \langle\mathtt{MASK}\rangle^{g_s-g_{s-1}}$.
    \item \hspace{0.9em}$p_s(\cdot)\gets
    \pi(\cdot\mid\widehat{X}^{(s)},o_{1:H_o})$.
    \item \hspace{0.9em}Sample $\widehat{G}_s$ from the final $g_s$ logit-read slots.
    \item \hspace{0.9em}$\widehat{T}_{1:b_s}\gets
    \widehat{T}_{1:b_{s-1}}\oplus\widehat{G}_s$.
    \item \textbf{end for}
    \item $\widetilde{T}\gets
    \widehat{T}_{1:K}\oplus\langle\mathtt{MASK}\rangle_{K+1:H_l}$.
    \item \textbf{return} action chunk
    $\hat{a}_{1:H_a}=\mathcal{T}^{-1}(\widetilde{T})$.
\end{enumerate}
\end{minipage}
\vspace{0.15em}
\hrule
\end{minipage}
\end{figure}

\section{\texorpdfstring{\oatfamily}{OAT} for Visuomotor Policies}
\label{sec:oat_policies}

We instantiate \oatfamily in the two policy interfaces introduced in
\cref{sec:prelim}. In autoregressive control, block-wise autoregression (\BAR)
specifies how ordered token positions are grouped into policy calls, while
\oatfamily provides the progressive, prefix-decodable action representation. In
token co-training, the full \oatfamily sequence supervises the VLM during
training, while a flow-matching expert generates continuous action chunks from
detached VLM context. The following subsections develop these interfaces in
turn.

\subsection{Block-wise Autoregressive \texorpdfstring{\oatfamily}{OAT} Generation}
\label{sec:bar}

\begin{figure}[t]
    \centering
    \includegraphics[width=\linewidth]{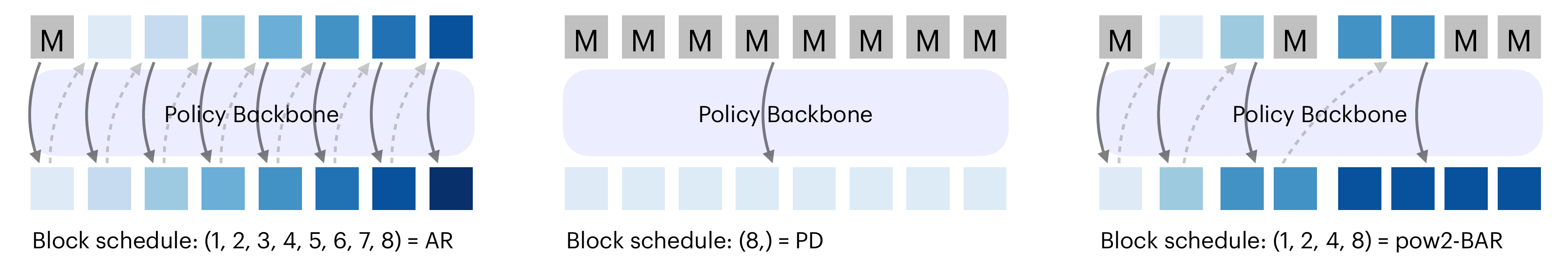}
    \caption{\textbf{\BAR block patterns.}
    \BAR generates action tokens block by block according to an endpoint list.
    For action token horizon $H_l=8$, token-wise autoregression uses endpoints
    $(1,2,3,4,5,6,7,8)$, parallel decoding uses endpoint $(8)$, and the
    power-of-two pattern uses endpoints $(1,2,4,8)$. Gray tokens marked ``M''
    are masks, and blue tokens are newly predicted action tokens.}
    \label{fig:bar_patterns}
\end{figure}

In the autoregressive role, a policy must choose how many action tokens to
predict per policy call. We formulate \BAR as a generation schedule over a
target budget $K\leq H_l$ within the fixed-length token sequence
$T_{1:H_l}$. An endpoint list $\mathbf{b}=(b_0,b_1,\ldots,b_S)$, with
$0=b_0<b_1<\cdots<b_S=K$, partitions the generated prefix into blocks
$G_s=T_{b_{s-1}+1:b_s}$ of size $g_s=b_s-b_{s-1}$. Stage $s$ predicts $G_s$ in
one policy call conditioned on the realized prefix $T_{1:b_{s-1}}$. Thus $S$
sets the sequential policy depth, while $g_s$ sets the number of tokens
predicted in parallel. Token-wise autoregression, fixed-size block
prediction~\citep{liu2025fasterefficientautoregressivevision,dong2026actioncodecmakesgoodaction},
one-shot parallel decoding~\citep{dong2026actioncodecmakesgoodaction}, and
intermediate variable-size patterns are all special cases, as illustrated in
\cref{fig:bar_patterns}. Full generation sets $K=H_l$.

For the power-of-two horizons considered here, $H_l=2^{S-1}$, we use endpoints
$(1,2,4,\ldots,H_l)$, so early stages use smaller blocks and preserve
fine-grained sequential dependencies, while later stages increase parallelism.
Full generation then requires
$S=1+\log_2 H_l$ policy calls, reducing the depth from $O(H_l)$ to
$O(\log H_l)$. \Cref{appendix:bar} shows that this depth is minimal when each
new block is no larger than the realized prefix.

To train a nondecreasing schedule $g_1\leq\cdots\leq g_S$, we set
$T_{1:b_0}=\varnothing$ and $g_0=0$, and use the block-shifted input
\[
    X^{(s)}
    =
    T_{1:b_{s-1}}
    \oplus
    \langle\mathtt{MASK}\rangle^{g_s-g_{s-1}}.
\]
The final $g_s$ positions provide the logit-read slots for predicting the
current block $G_s$. For $s>1$, these slots comprise the previous block
$G_{s-1}$ followed by $g_s-g_{s-1}$ new masks; at $s=1$, they contain $g_1$
masks. Let
$p_{\pi,j}^{(s)}(\cdot\mid X^{(s)},o)$ denote the categorical distribution read
from the $j$-th such slot under policy $\pi$ and observation context $o$. The
training objective is
\[
    \mathcal{L}_{\mathrm{BAR}}
    =
    -\frac{1}{S}
    \sum_{s=1}^{S}
    \frac{1}{g_s}
    \sum_{j=1}^{g_s}
    \log
    p_{\pi,j}^{(s)}\!\left(
      T_{b_{s-1}+j} \mid X^{(s)}, o
    \right).
\]
Thus, we average within each block and weight all generation stages equally.
This objective produces all $g_s$ logits in one policy forward pass without
exposing ground-truth tokens from $G_s$. At inference, the same construction
uses the generated prefix and appends the predicted block $\widehat{G}_s$ after
each call.

Both evaluated variants train the tokenizer decoder at
$\mathcal{K}=\{1,2,4,\ldots,H_l\}$. The \BAR endpoint list instead specifies how
the policy reaches a target budget $K$ and may include intermediate generation
endpoints. \oat uses singleton blocks with $b_s=s$, requiring $K$ policy calls
to reach a length-$K$ prefix. \oatpowtwo uses endpoints
$(1,2,4,8,16)$ for $H_l=16$, requiring five calls for complete generation. At
any generation endpoint, the generated prefix can be suffix-padded with learned
mask tokens and decoded immediately. Budgets in $\mathcal{K}$ receive direct
reconstruction supervision and typically yield higher reconstruction quality;
generation can also continue to a larger budget.
\Cref{algo:oat_policy_inference,appendix:bar} provide the full inference procedure
and \BAR specification.

\subsection{Token Co-Training with \texorpdfstring{\oatfamily}{OAT}}
\label{sec:tc}

In the token co-training role, \oatfamily tokens supervise the VLM rather than
being decoded into actions. Let $c$ denote the image, language, and robot state
context. Given a frozen tokenizer $\mathcal{T}$, each training action chunk
$a_{1:H_a}$ defines a target sequence
$T_{1:H_l}=\mathcal{T}(a_{1:H_a})$. The VLM predicts this sequence under
teacher forcing, while the action expert learns from the same action chunk and
a detached VLM key/value (K/V) cache. Their joint objective
is~\citep{driess2025knowledge,fang2026molmoact2actionreasoningmodels}
\begin{equation}
\label{eq:tc_objective}
    \mathcal{L}_{\mathrm{TC}}
    =
    \mathcal{L}_{\mathrm{tok}}(T_{1:H_l}\mid c)
    +
    \lambda\,
    \mathcal{L}_{\mathrm{flow}}
    \!\left(a_{1:H_a},\widetilde{a}_{1:H_a}^{\,\tau},\tau;
    \operatorname{sg}(\mathrm{KV}_{\mathrm{VLM}}(c))\right),
\end{equation}
where $\mathcal{L}_{\mathrm{tok}}$ is the cross-entropy for predicting each $T_i$
from $c$ and the shifted ground-truth prefix
$\langle\mathtt{MASK}\rangle,T_{1:i-1}$. The leading mask supplies the prediction
slot for $T_1$, and $\lambda$ weights the flow objective relative to the token
objective. The term $\widetilde{a}_{1:H_a}^{\,\tau}$ denotes the noised
action chunk at flow time $\tau$, and $\operatorname{sg}(\cdot)$ denotes
stop-gradient. The layer-wise cache $\mathrm{KV}_{\mathrm{VLM}}(c)$ is computed
from the image, language, and robot state prefill only, excluding the
teacher-forced action token positions.

The two losses follow separate gradient paths. The token loss provides action
supervision to the VLM, whereas the flow loss trains the action expert without
propagating through the detached cache. At inference, the VLM is prefilled once
on $c$, but its action token logits are not sampled. The flow-matching expert
instead generates a continuous action chunk from the cached context for
execution; no action tokens are generated or detokenized.

Because token co-training uses the full \oatfamily sequence in a teacher-forced
token objective rather than for action decoding, the relevant benefit of
\oatfamily is the objective it imposes on the VLM prefill representation.
The one-token prefix
objective trains the first \oatfamily token to support reconstruction of the
complete action chunk. Under token co-training, this is the first target in the
teacher-forced sequence and is therefore predicted directly from $c$, without
preceding action tokens. Its cross-entropy loss therefore imposes a
chunk-summary objective on the prefill representation later consumed by the
action expert. We view this as plan-like supervision: the VLM must infer a
summary of the complete action chunk from the
image, language, and robot state context alone. By comparison, coordinate
binning assigns the first target to a single action scalar, while learned latent
baselines whose decoders are trained only on complete code sequences do not
explicitly train their first target to support full-chunk reconstruction. These
alternatives therefore do not provide analogous chunk-summary supervision for
the VLM's first prediction.
\Cref{fig:vlm_tc_success_grid} evaluates this supervision against alternative
action tokenizers; \cref{appendix:tc} provides further flow-matching and inference
details.

%% file: body/experiments.tex
% !TEX root = ../main.tex

\section{Experiments}
\label{sec:experiments}

Experiments assess \oatfamily at three levels of the action token interface.
We begin at the tokenizer level in \cref{sec:exp_rate_distortion}, testing
whether \oatfamily forms a compact, progressive action representation whose
prefixes remain executable rather than only optimizing full-sequence
reconstruction. We then move to closed-loop policy learning in
\cref{sec:exp_policy}, spanning lightweight policies and VLM-scale policies
under autoregressive (AR) generation and token co-training (TC). Finally,
\cref{sec:exp_ablation} analyzes token
ordering, action and token horizons, codebook capacity, and grouped generation
to clarify the design choices behind \oatfamily.

\begin{figure}[!t]
    \centering
    \begin{minipage}[t]{0.188\linewidth}
      \centering
      \includegraphics[width=\linewidth]{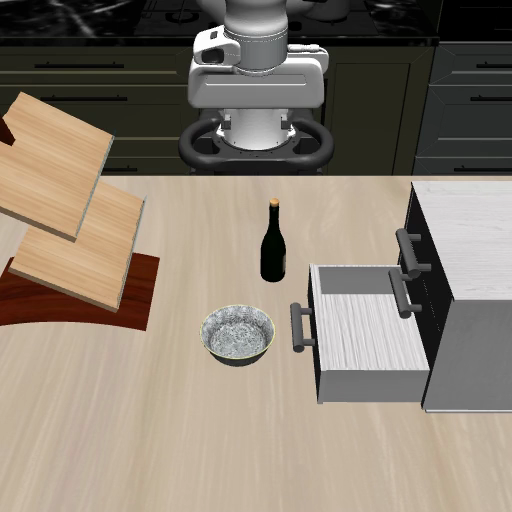}\\[-0.2em]
      \makebox[\linewidth][c]{\footnotesize \libero~\floatcite{liu2023liberobenchmark}}
    \end{minipage}\hfill
    \begin{minipage}[t]{0.188\linewidth}
      \centering
      \includegraphics[width=\linewidth]{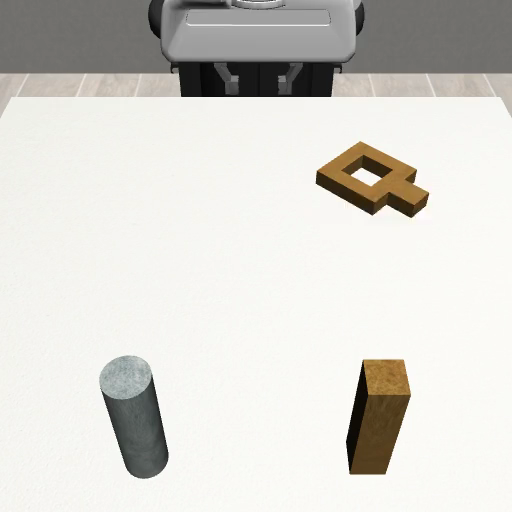}\\[-0.2em]
      \makebox[\linewidth][c]{\footnotesize \robomimic~\floatcite{robomimic2021}}
    \end{minipage}\hfill
    \begin{minipage}[t]{0.188\linewidth}
      \centering
      \includegraphics[width=\linewidth]{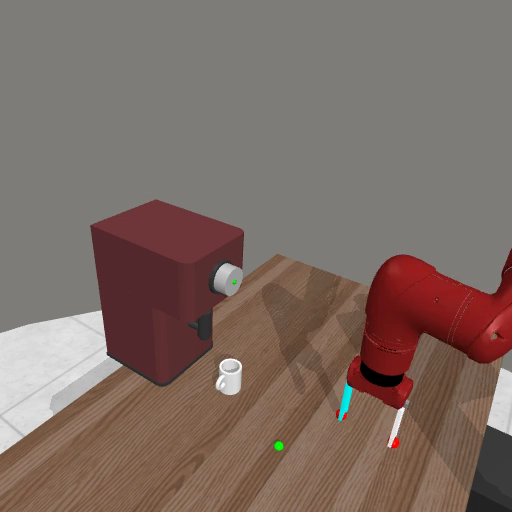}\\[-0.2em]
      \makebox[\linewidth][c]{\footnotesize \metaworld~\floatcite{yu2020metaworldbenchmark}}
    \end{minipage}\hfill
    \begin{minipage}[t]{0.188\linewidth}
      \centering
      \includegraphics[width=\linewidth]{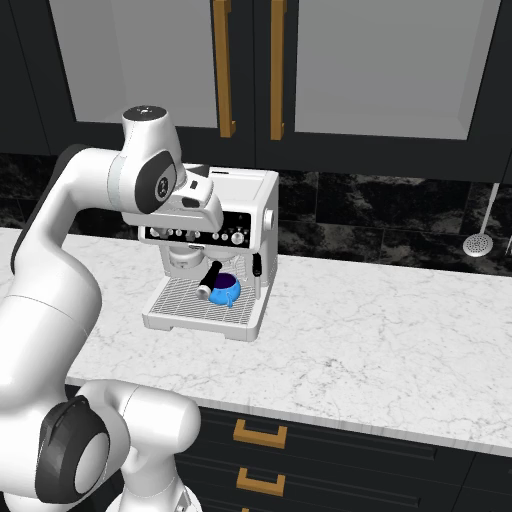}\\[-0.2em]
      \makebox[\linewidth][c]{\footnotesize \robocasa~\floatcite{robocasa2024,robocasa365}}
    \end{minipage}\hfill
    \begin{minipage}[t]{0.188\linewidth}
      \centering
      \includegraphics[width=\linewidth]{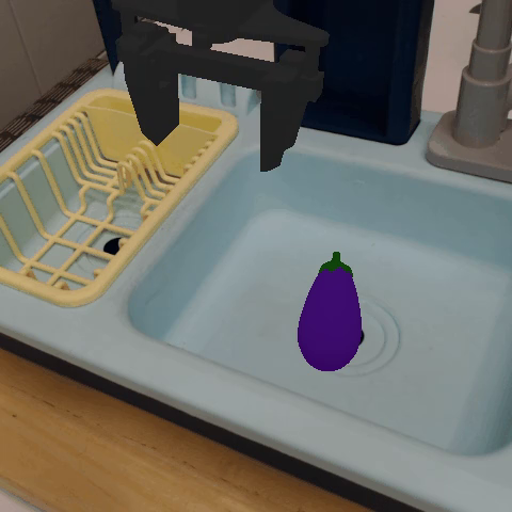}\\[-0.2em]
      \makebox[\linewidth][c]{\footnotesize \simpler~\floatcite{li24simpler}}
    \end{minipage}

    \vspace{0.55em}
    \begin{minipage}[t]{0.4925\linewidth}
      \centering
      \includegraphics[width=\linewidth]{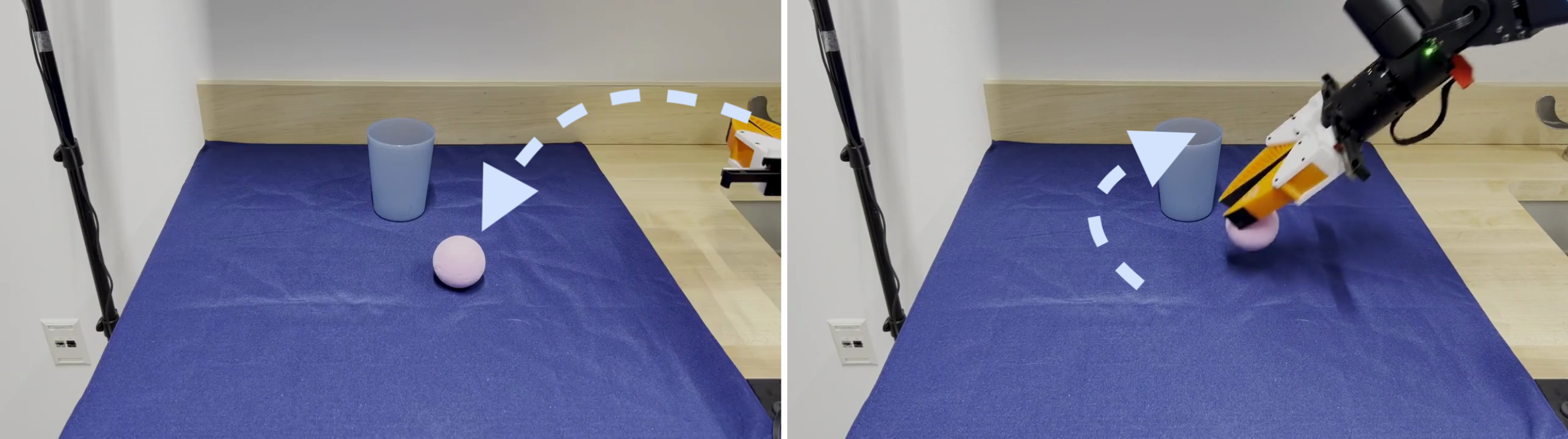}\\[-0.25em]
      \makebox[\linewidth][c]{\footnotesize Pick-and-Place Ball}
    \end{minipage}\hfill
    \begin{minipage}[t]{0.4925\linewidth}
      \centering
      \includegraphics[width=\linewidth]{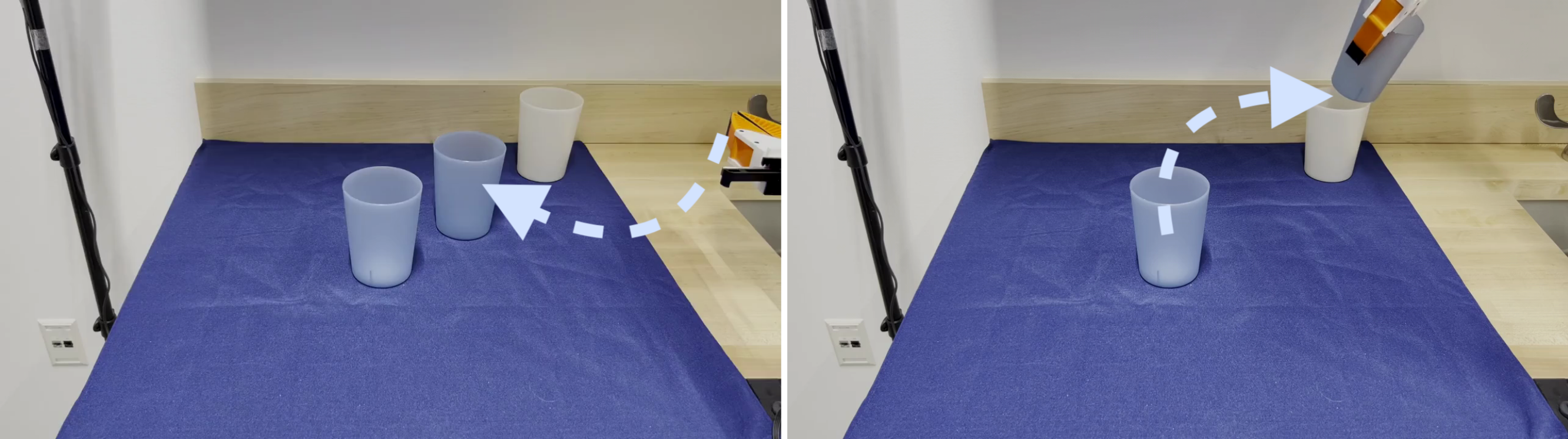}\\[-0.25em]
      \makebox[\linewidth][c]{\footnotesize Stack Cups}
    \end{minipage}
    \caption{\textbf{Evaluation environments.} \textbf{Top row:} simulated
    manipulation benchmarks used for lightweight policies, VLM AR policies,
    and VLM TC policies.
    \textbf{Bottom row:} real-world tabletop tasks evaluated with a fixed-base
    ARX-5 arm and a single Logitech webcam; each filmstrip shows one
    representative rollout.}
    \label{fig:evaluation_envs}
\end{figure}

\subsection{Experimental Setup}

We evaluate \oatfamily at two policy scales. The lightweight regime fixes the
\ctrltransformer backbone and compares action representations across
\libero-Long~\citep{liu2023liberobenchmark},
\robomimic~\citep{robomimic2021}, \metaworld~\citep{yu2020metaworldbenchmark},
\robocasa~\citep{robocasa2024}, and two real-world tasks. The VLM-scale regime
uses \paligemma and \qwenvl backbones under AR generation and TC supervision.
The evaluation covers \libero, \robomimic, and
\simpler~\citep{li24simpler}. We additionally use
\robocasaThreeSixFive~\citep{robocasa365}.

Within each comparison block, methods share the observation interface, data
split, rollout protocol, and policy backbone. The varied factors are the action
representation and, for AR policies, the generation pattern; TC comparisons
instead vary token supervision while holding the flow-matching expert
architecture, objective, and training recipe fixed.
\Cref{fig:evaluation_envs} shows the environments,
\cref{tab:benchmark_setup} summarizes benchmark coverage, and
\cref{tab:policy_backbones,tab:method_setup} specify the policy interfaces and
generation costs. Full rollout, tokenizer, policy, and optimization details are
provided in \cref{appendix:implementation_details}.

\begin{table}[!t]
  \centering
  \footnotesize
  \setlength{\tabcolsep}{2.8pt}
  \renewcommand{\arraystretch}{1.16}
  \begin{tabular*}{\linewidth}{
    @{\hspace{0.55em}\extracolsep{\fill}}
    >{\raggedright\arraybackslash}p{0.102\linewidth}
    >{\raggedright\arraybackslash}p{0.160\linewidth}
    >{\raggedright\arraybackslash}p{0.398\linewidth}
    >{\centering\arraybackslash}p{0.070\linewidth}
    r@{\,$\times$\,}l
    >{\centering\arraybackslash}p{0.045\linewidth}
    >{\centering\arraybackslash}p{0.065\linewidth}
    @{\hspace{0.55em}}
    }
  \toprule
    Setting & Benchmark & Tasks / suites & \# Tasks & \multicolumn{2}{c}{\makecell{$H_a$\\$\times D_a$}} & \makecell{Freq.\\(Hz)} & \makecell{Med.\\len.} \\
    \midrule
    \multirow{5}{*}{Lightweight}
      & \libero
      & Long suite
      & 10 & 32 & 7 & 10 & 259 \\
      & \robomimic
      & Lift; Square; Can
      & \phantom{0}3 & 32 & 7 & 20 & 114 \\
      & \metaworld
      & Box Close; Coffee Pull; Disassemble; Stick Pull
      & \phantom{0}4 & 32 & 4 & 80 & \phantom{0}72 \\
      & \robocasa
      & Close Drawer; Coffee Press Button; Turn Off Microwave; Turn Off Sink Faucet
      & \phantom{0}4 & 32 & 12 & 20 & 184 \\
      \cmidrule(lr){2-8}
      & Real-world
      & Pick-and-Place Ball; Stack Cups
      & \phantom{0}2 & 32 & 7 & 10 & \phantom{0}98 \\
    \midrule
    \multirow{4}{*}{VLM}
      & \libero
      & Long, Goal, Object, and Spatial suites
      & 40 & 32 & 7 & 10 & 140 \\
      & \robomimic
      & Lift; Square; Can; Tool Hang
      & \phantom{0}4 & 32 & 7 & 20 & 130 \\
      & \simpler
      & WidowX+Bridge series
      & \phantom{0}4 & 8 & 7 & \phantom{0}5 & \phantom{0}37 \\
      & \robocasaThreeSixFive
      & Close Toaster Oven Door; Open Drawer; Pick Place Drawer to Counter; Turn On Electric Kettle; Slide Dishwasher Rack
      & \phantom{0}5 & 32 & 12 & 20 & 194 \\
    \bottomrule
  \end{tabular*}
  \caption{\textbf{Benchmark setup.} Rows summarize the simulated and real-world
  settings used for policy evaluation. $H_a\times D_a$ gives the predicted action
  horizon and action dimension; Med. len. is the median episode length in
  environment steps. Task subsets differ by evaluation regime as listed. The
  lightweight regime fixes a \ctrltransformer policy, while the VLM regime uses
  the backbones in \cref{tab:policy_backbones}; within each regime, tokenizer
  comparisons keep the policy backbone fixed.}
  \label{tab:benchmark_setup}
\end{table}

\begin{table}[t]
  \centering
  \begin{minipage}[t]{0.54\linewidth}
    \vspace{0pt}
    \footnotesize
    \setlength{\tabcolsep}{2.5pt}
    \renewcommand{\arraystretch}{1.14}
    \begin{tabular}{@{\hspace{0.55em}}lcccc@{\hspace{0.55em}}}
      \toprule
      Backbone & Scale & Interface & Ctx. attn. & Act. attn. \\
      \midrule
      \ctrltransformer & 5M & Cross-attn. & Cross & Causal \\
      \paligemma~\floatcite{steiner2024paligemma2} & 3B & Prefix VLM & Full & Block-causal \\
      \qwenvl~\floatcite{bai2025qwen3vl} & 2B & Causal VLM & Causal & Block-causal \\
      \bottomrule
    \end{tabular}
  \end{minipage}\hfill
  \begin{minipage}[t]{0.44\linewidth}
    \vspace{0pt}
    \captionsetup{justification=raggedright,singlelinecheck=false}
    \caption{\textbf{Policy backbone interfaces.} Rows define the backbone
    attention interfaces used in later comparisons. Ctx. attn. covers
    observation, language, and state tokens; Act. attn. covers action token
    positions.}
    \label{tab:policy_backbones}
  \end{minipage}
\end{table}

\begin{table}[t]
  \centering
  \footnotesize
  \setlength{\tabcolsep}{2.0pt}
  \renewcommand{\arraystretch}{1.12}
  \begin{tabular*}{\linewidth}{@{\hspace{0.55em}\extracolsep{\fill}}lllcc@{\hspace{0.55em}}}
    \toprule
    Scheme & Token structure & Generation & \# Calls & Max block \\
    \midrule
    \bin & Raw coordinate bins & stepwise $D_a$ blocks & $H_a$ & $D_a$ \\
    \fast & Frequency coefficients with BPE & token-wise BPE sequence & $|T|$ & $1$ \\
    \quest & Learned temporal latents & token-wise latent sequence & $H_l$ & $1$ \\
    \acodec & Joint latent block & one-shot latent block & $1$ & $H_l$ \\
    \addlinespace[0.15em]
    \oat[k] & token-wise ordering & token-wise AR & $k$ & $1$ \\
    \oatpowtwo[k] & power-of-two ordering & power-of-two \BAR & $1+\lceil\log_2 k\rceil$ & $\lceil k/2\rceil$ \\
    \bottomrule
  \end{tabular*}
  \caption{\textbf{Action token budget and generation settings.} Rows compare each
  tokenizer setting, generation pattern, and call cost. \# Calls is the
  number of sequential policy calls for one action chunk, and Max block is
  the largest action token block generated in one policy call. For the
  \oatfamily rows, both variants train at power-of-two reconstruction budgets,
  and $k$ denotes one such budget. \oatpowtwo additionally uses power-of-two
  register groups and \BAR endpoints. For \fast, $|T|$ denotes generated BPE
  length.}
  \label{tab:method_setup}
\end{table}

\subsection{Rate--Distortion of Action Tokens}
\label{sec:exp_rate_distortion}

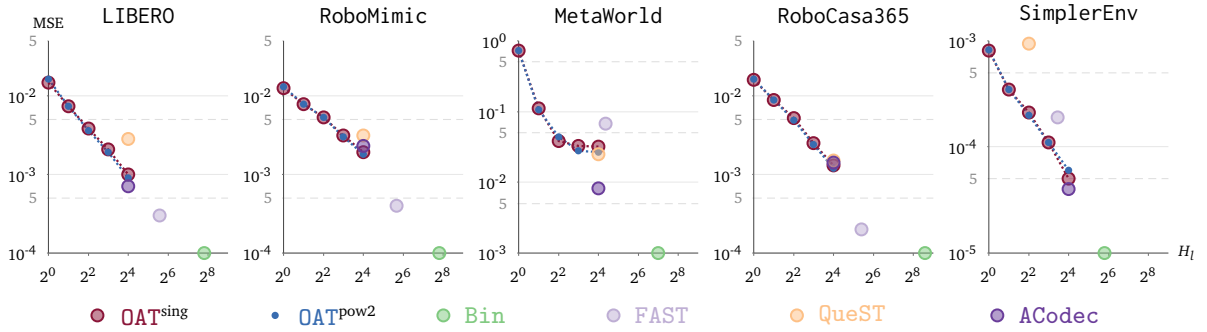
\begin{figure}[t]
  \centering
  \input{figs/tikz/rate_distortion_curves}
  \caption{\textbf{Rate--distortion curves for action tokenizers.} Each panel
  plots raw reconstruction MSE against token budget on log--log axes; lower and
  leftward positions indicate better rate--distortion tradeoffs. The \oatfamily
  traces show mask-padded partial
  decodings under token-wise and power-of-two variants for
  $k\in\{1,2,4,8,16\}$, while \bin, \fast, \quest, and \acodec appear as
  fixed-budget operating points at their respective full token lengths, with \fast using its
  average BPE length. Near-zero \bin errors are drawn on the x-axis for visual
  consistency. Detailed token counts and MSE values are listed in
  \cref{appendix:tab:num_tokens}.}
  \label{fig:rate_distortion_curves}
\end{figure}

\Cref{fig:rate_distortion_curves} measures reconstruction error from truncated
token prefixes. For token budget $k$, we retain $T_{1:k}$, fill the ungenerated
suffix with mask tokens, decode the partial sequence, and report its mean
squared error against the original action chunk. This diagnostic quantifies how
much action information each prefix budget preserves before the tokens are used
as policy targets.

The baselines expose distinct rate--distortion operating points. \bin is nearly
lossless but requires $H_aD_a$ tokens. \fast shortens the sequence through
frequency-domain coding and BPE, while \quest and \acodec provide compact
learned-latent operating points at their full token horizons. In contrast,
\oatfamily traces a family of operating points from one-token sketches to
full-length reconstructions, allowing the same tokenizer to trade token budget
for action fidelity.

Both \oatfamily variants reduce reconstruction error smoothly as the token
budget increases, and the power-of-two mask closely tracks the token-wise mask
across budgets. Thus grouped register dependencies preserve progressive
rate--distortion behavior without materially degrading reconstruction.
Exact token counts and MSEs, reported in units of $10^{-3}$, are listed in
\cref{appendix:rate_distortion}.

Rate--distortion measures information retention, but it does not establish
whether the resulting tokens support effective policy learning.

\subsection{Policy Evaluation}
\label{sec:exp_policy}

We next ask whether tokenizer design translates into closed-loop policy
performance. We evaluate this question across policy scales and action-token
roles.

\subsubsection{Lightweight Policies}

\begin{table}[t]
  \centering
  \footnotesize
  \setlength{\tabcolsep}{9.0pt}
  \renewcommand{\arraystretch}{1.08}
  \begin{tabular*}{\linewidth}{@{\hspace{0.55em}\extracolsep{\fill}}lccccccc@{\hspace{0.55em}}}
    \toprule
    \multirow{2}{*}{Scheme}
    & \multicolumn{4}{c}{Simulation}
    & \multicolumn{2}{c}{Real-world}
    & \multirow{2}{*}{Avg. Rank} \\
    \cmidrule(lr){2-5}\cmidrule(lr){6-7}
    & \libero-Long & \robomimic & \metaworld & \robocasa
    & P\&P Ball & Stack Cups & \\
    \midrule
    \bin & 14.4 & 39.5 & 14.5 & 27.7 & \phantom{0}4/20 & \phantom{0}8/20 & 5.8 \\
    \fast & 23.0 & 24.0 & \phantom{0}7.1 & 13.2 & \phantom{0}8/20 & \phantom{0}6/20 & 6.2 \\
    \quest & 48.2 & 66.9 & 17.9 & 52.3 & 11/20 & \phantom{0}8/20 & 2.8 \\
    \midrule
    \oat[1] & 11.7 & 50.8 & 11.3 & 47.7 & \phantom{0}7/20 & \phantom{0}3/20 & 6.0 \\
    \oat[2] & 39.8 & 52.5 & 16.4 & 50.3 & 11/20 & \phantom{0}9/20 & 3.8 \\
    \oat[4] & 46.4 & 65.3 & 19.5 & 51.7 & 13/20 & 12/20 & 2.5 \\
    \oat[8] & 56.3 & 73.1 & 24.4 & 54.6 & 16/20 & 16/20 & 1.0 \\
    \bottomrule
  \end{tabular*}
  \caption{\textbf{Lightweight policy simulation and real-world results.}
  Simulation entries are mean success rates in percent with a fixed
  \ctrltransformer policy; real-world entries are successful trials out of 20
  independent rollouts. For this lightweight comparison, \fast uses strict decoding, so invalid
  \fast token sequences are rejected as described in
  \cref{sec:action_token_prelim}. Avg. Rank is balanced over the four
  simulation benchmarks and two real-world tasks; lower is better.}
  \label{tab:oat_controlled_sim_real}
\end{table}

\Cref{tab:oat_controlled_sim_real} compares tokenizers in closed-loop control
using the same small \ctrltransformer policy, isolating the action
representation and generation pattern. We evaluate on \libero-Long,
\robomimic, \metaworld, and \robocasa, followed by the Pick-and-Place Ball and
Stack Cups real-robot tasks in \cref{fig:evaluation_envs}.

\oatfamily performance improves with token budget: the shortest budgets are
executable but coarse, whereas \oat[8] gives the highest point estimate on every
simulation benchmark and real-world task and the best average rank.
Both \quest and \oatfamily use compact learned latents, but only \oatfamily
trains ordered prefixes to place control-relevant action information early; the
matched ordering ablation in \cref{fig:oat_ordering_ablation_plot} tests this
factor directly. These results provide closed-loop evidence that \oatfamily
prefixes improve policy learning and motivate evaluation at VLM scale.

At VLM scale, AR policies generate and detokenize action tokens, whereas TC
policies use token losses to supervise the VLM while a flow-matching expert
executes actions.

\subsubsection{Autoregressive VLM Policies}

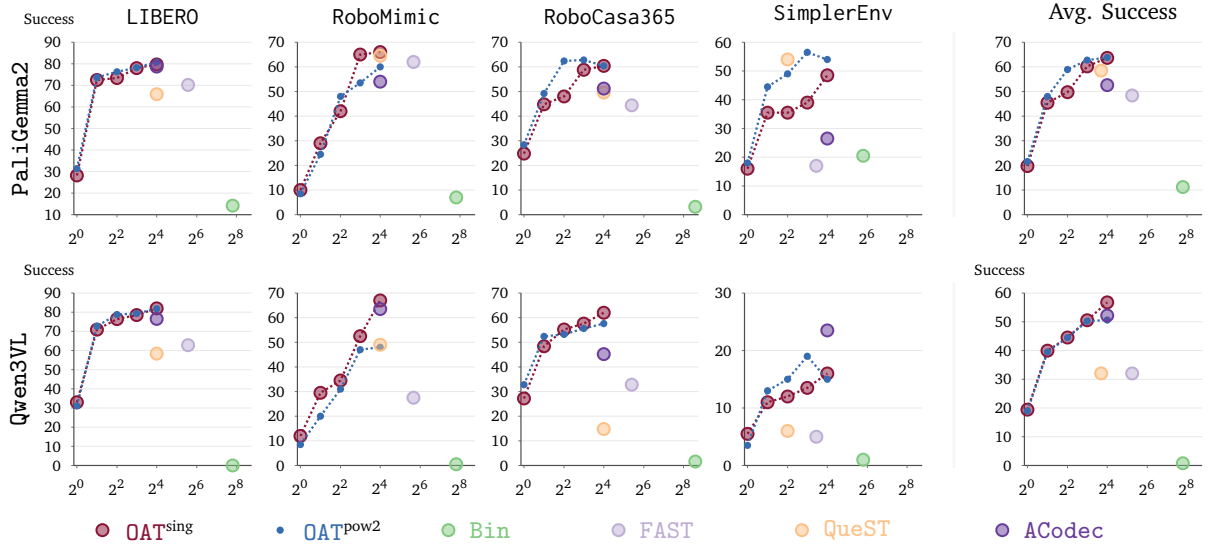
\begin{figure}[p]
  \centering
  \input{figs/tikz/vlm_bar_success_grid}
  \caption{\textbf{Closed-loop autoregressive VLM policy success rates.} Each panel reports
  mean task success over 50 rollouts per task for one VLM backbone and benchmark
  group. The x-axis is token budget on a $2^t$ scale. Dotted curves trace
  \oatfamily token budgets; fixed-budget baselines are plotted at their
  tokenizer lengths. Call counts are listed in \cref{tab:method_setup}.
  Rightmost panels report
  benchmark-balanced average success for each backbone. Numeric values are
  provided in \cref{appendix:tab:simulation_benchmarking}.}
  \label{fig:vlm_bar_success_grid}
\end{figure}

\Cref{fig:vlm_bar_success_grid} shows that \oatfamily remains effective for VLM
AR policies. Across both backbones, success generally improves with token
budget; the best token-wise results occur at \oat[16], reaching $63.7$ with \paligemma and
$56.8$ with \qwenvl.

\BAR exposes the depth--accuracy tradeoff for the same ordered token family. At
$k=16$, the power-of-two variant matches token-wise generation with \paligemma
($63.8$ vs. $63.7$) using 5 rather than 16 policy calls; with \qwenvl,
token-wise generation performs better ($56.8$ vs. $50.6$). Full values appear in
\cref{appendix:vlm_policy_results}.

Baseline comparisons confirm that tokenizer design remains consequential at
VLM scale. \bin reconstructs almost exactly in \cref{appendix:tab:num_tokens},
yet its long action suffix gives poor average success with both backbones.
\acodec instead predicts all action tokens in one policy call, but varies across
benchmarks and backbones. Useful action tokens must therefore be compact and
decodable while remaining learnable by the policy interface.

\subsubsection{Token Co-Training VLM Policies}

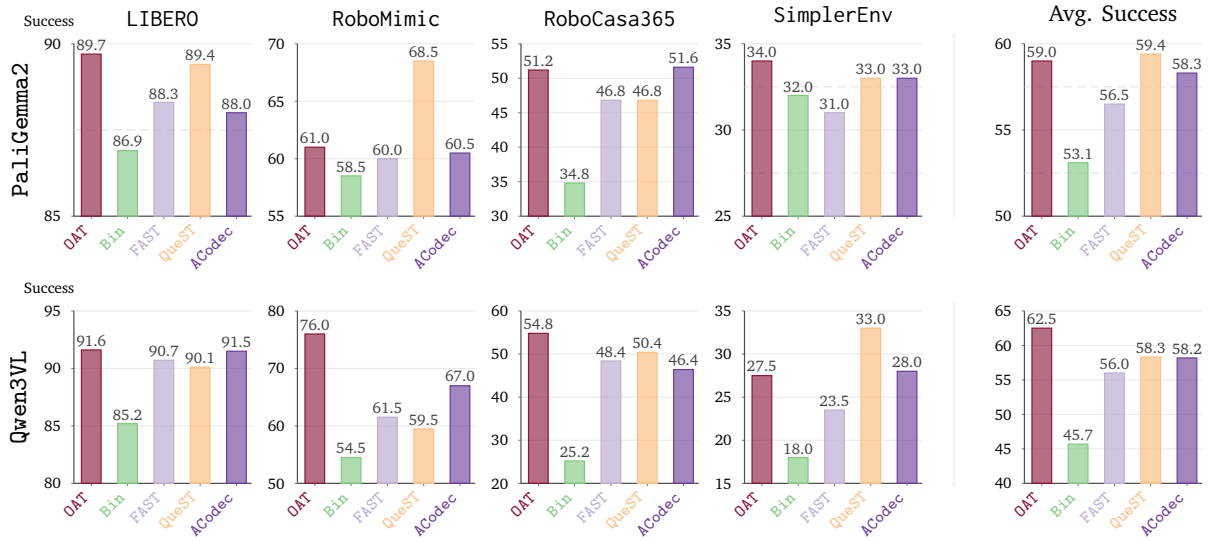
\begin{figure}[p]
  \centering
  \input{figs/tikz/vlm_ki_success_grid}
  \caption{\textbf{Closed-loop token co-training VLM policy success rates.} Rows are VLM
  backbones trained with action token supervision and a stop-gradient boundary
  that blocks flow-matching expert losses from updating the VLM. Columns report
  mean task success over 50 evaluation episodes per task on
  \libero, \robomimic, \robocasaThreeSixFive, and \simpler, plus the
  average with equal weight across the four benchmark panels. Within each row, bars
  compare the action token supervision used during training: \oatfamily, \bin, \fast,
  \quest, and \acodec. At inference, these tokens are not decoded; the
  flow-matching expert produces actions from cached, detached VLM K/V context.
  Numeric labels give success in percent. Each panel uses its own vertical axis
  scale, shown by tick labels, so compare bar heights within each panel.}
  \label{fig:vlm_tc_success_grid}
\end{figure}

\Cref{fig:vlm_tc_success_grid} shows that tokenizer choice remains consequential
under TC supervision. With \paligemma, \oatfamily reaches an average success
rate of $59.0$, comparable to \quest ($59.4$) and \acodec ($58.3$), while
outperforming \bin and \fast. It leads on \libero and \simpler, whereas \quest is
stronger on \robomimic ($68.5$ vs. $61.0$). With \qwenvl, \oatfamily gives the highest average success at $62.5$
and leads on \robomimic and \robocasaThreeSixFive; \quest remains stronger on
\simpler ($33.0$ vs. $27.5$). Thus tokenizer design matters even when tokens
supervise the VLM rather than define the executed action, although the best
tokenizer can depend on the backbone and benchmark. This supports
\cref{sec:tc}: the plan-like first-token target directly supervises the VLM
prefill representation consumed by the expert.

\subsection{Ablation and Analysis}
\label{sec:exp_ablation}

Ablations examine the main design choices behind \oatfamily: token ordering,
action and token horizons, codebook capacity, and grouped generation. Unless
otherwise specified, these studies use the lightweight
\ctrltransformer setting, so each comparison keeps the policy backbone fixed
and changes only the action representation.

\subsubsection{Does Token Ordering Improve Policy Performance?}
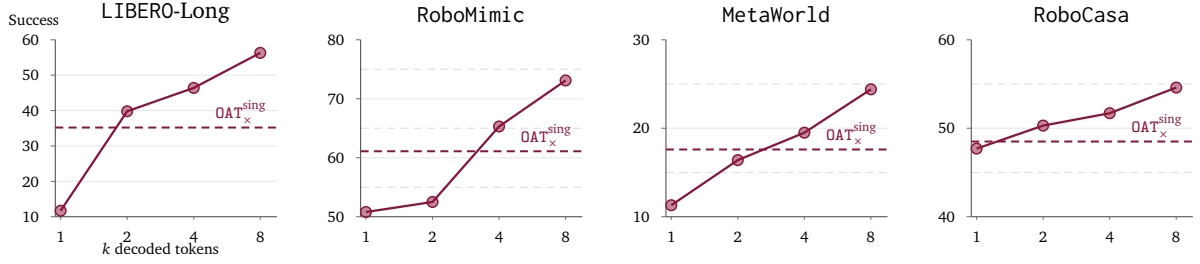
\begin{figure}[t]
  \centering
  \input{figs/tikz/ordering_ablation_plot}
  \caption{\textbf{Token ordering ablation.} Each panel reports one benchmark in
  the lightweight setting, with mean success over 50 rollouts per task when
  decoding the first $k$ \oat tokens. Solid curves show ordered \oat budgets;
  dashed lines show \oat[\times], which removes nested dropout during tokenizer
  training while keeping the tokenizer architecture and policy interface fixed.}
  \label{fig:oat_ordering_ablation_plot}
\end{figure}

\Cref{fig:oat_ordering_ablation_plot} evaluates the effect of token ordering.
\oat[\times] removes nested dropout, so short prefixes are no longer trained to
reconstruct the action chunk. This separates the ordering objective from
learnable registers and compact latent capacity.

Removing ordering consistently degrades lightweight policy success.
\oat[\times] still uses the full latent horizon, so it is often stronger than
the shortest \oat budgets, but it remains well below \oat[8] and is often closer
to \oat[2] or \oat[4]. This supports the claim that compact latent tokens alone
are insufficient for strong policies. Ordering is an important factor:
next-token prediction benefits when high-impact action structure appears early
and residual detail later.

\subsubsection{How Do Action and Token Horizons Trade Off?}

\Cref{fig:oat_ha_hl_ablation} studies two central factors in chunk-level action
tokenization: the predicted action horizon $H_a$ and the latent token horizon
$H_l$. Larger $H_a$ provides more future context to the policy, but also
requires the tokenizer to compress a longer continuous trajectory. Larger $H_l$
provides more register slots, but increases the action token suffix that the
policy must model. We train models for $H_a\in\{8,16,32,64\}$ and
$H_l\in\{1,2,4,8\}$ on \libero-Long and evaluate two execution protocols:
execution after half the action horizon, which matches the protocol used
elsewhere, and fixed 8-action execution, which holds execution frequency
constant.

\begin{figure}[t]
  \centering
  \input{figs/tikz/oat_ha_hl_heatmaps}
  \begin{minipage}[t]{0.40\linewidth}
    \vspace{0pt}
    \vspace{1.8em}
    \centering
    \footnotesize
    \setlength{\tabcolsep}{2.2pt}
    \renewcommand{\arraystretch}{1.08}
    \begin{tabular*}{\linewidth}[t]{@{\hspace{0.55em}\extracolsep{\fill}}lcc@{\hspace{0.55em}}}
      \toprule
      FSQ levels & $|\mathcal{V}|$ & \libero \\
      \midrule
      $[8,6,5]$ & \phantom{0}240 & 29.2 \\
      $[8,8,8]$ & \phantom{0}512 & 53.5 \\
      $[8,5,5,5]$ & 1000 & 56.3 \\
      $[8,8,6,5]$ & 1920 & 54.6 \\
      $[7,5,5,5,5]$ & 4375 & 46.9 \\
      \bottomrule
    \end{tabular*}
  \end{minipage}

  \vspace{4pt}
  \begin{minipage}[t]{0.56\linewidth}
    \vspace{5pt}
    \captionsetup{justification=raggedright,singlelinecheck=false}
    \caption{\textbf{Action and token horizons.} \oat[H_l] success on
    \libero-Long with a fixed \ctrltransformer policy as action horizon $H_a$
    and latent token horizon $H_l$ vary. \textbf{(a)} queries again after executing
    $\tfrac{1}{2}H_a$ actions; \textbf{(b)} always executes 8 actions, isolating
    execution frequency. The panels expose compression and replanning
    tradeoffs. Values are mean success rates over 50 rollouts per task; darker
    cells are better.}
    \label{fig:oat_ha_hl_ablation}
  \end{minipage}\hfill
  \begin{minipage}[t]{0.40\linewidth}
    \vspace{5pt}
    \captionsetup{justification=raggedright,singlelinecheck=false}
    \captionof{table}{\textbf{Codebook capacity scaling.} \oat success on \libero-Long
    with fixed \ctrltransformer ($H_a=32$, $H_l=8$) as FSQ levels
    vary~\floatcite{mentzer2024fsq}. Moderate vocabularies work best: too few
    codes can restrict the representation, while too many can make token targets
    harder to learn.}
    \label{tab:oat_codebook_ablation}
  \end{minipage}
\end{figure}

Executing half the action horizon exposes the compression side of the tradeoff.
For a fixed $H_l$, success generally drops as $H_a$ grows because the same
number of tokens must represent a longer future chunk. Increasing $H_l$
mitigates this drop, indicating that additional register slots are needed to
preserve fine temporal structure. This supports using a larger latent horizon
such as $H_l=8$, while the action horizon must also account for execution
frequency.

The fixed 8-action protocol separates prediction horizon from execution
frequency. For a fixed $H_l$, a longer $H_a$ can initially help because the
policy predicts further into the future while executing the same number of
actions per query. However, very long horizons again become difficult to
compress, especially for $H_l\in\{1,2\}$. In this setting, $H_a=32$ with
$H_l=8$ is a strong compromise between lookahead and compression. The two
heatmaps therefore point to the same design rule: action chunking and token
capacity should be chosen jointly.

\subsubsection{How Does Codebook Capacity Affect Policy Success?}
\Cref{tab:oat_codebook_ablation} varies the FSQ levels while keeping the rest
of the tokenizer fixed, isolating discrete vocabulary capacity from the number
of latent tokens. The trend is non-monotonic. Success improves as the vocabulary
increases from 240 to roughly 1000--2000 codes, consistent with the intuition
that a small codebook is restrictive. Beyond that range, performance drops even
though the tokenizer has more discrete capacity. One plausible explanation is
that larger vocabularies spread supervision over more discrete targets, so the
policy observes fewer examples per code. This reinforces the broader point from
\cref{sec:action_token_prelim}: effective action tokens must reconstruct
actions and remain learnable policy targets.

\subsubsection{Does Grouped Generation Require Matched \texorpdfstring{\oatfamily}{OAT} Ordering?}

The final ablation tests whether \oatfamily can be regrouped at inference
without matching its register dependency structure to the generation blocks.
Both tokenizers use the same reconstruction budgets. We apply power-of-two
grouped generation post hoc to \oat[16] by overriding only its generation
endpoint list while retaining its causal register attention. We compare this
condition with \oatpowtwo[16], whose block-causal register groups match the
generation blocks.

\begin{table}[t]
  \centering
  \begin{minipage}[t]{0.655\linewidth}
    \vspace{0pt}
    \centering
    \footnotesize
    \setlength{\tabcolsep}{1.5pt}
    \renewcommand{\arraystretch}{1.10}
    \begin{tabular*}{\linewidth}{@{\hspace{0.55em}\extracolsep{\fill}}lllcc@{\hspace{0.55em}}}
      \toprule
      Scheme & Ordering & Generation & Calls & \libero \\
      \midrule
      \oat[16] & token-wise & token-wise AR & 16 & 79.7 \\
      \oat[16] + post-hoc \BAR & token-wise & power-of-two \BAR & \phantom{0}5 & 66.3 \\
      \oatpowtwo[16] & power-of-two & power-of-two \BAR & \phantom{0}5 & 80.8 \\
      \bottomrule
    \end{tabular*}
  \end{minipage}\hfill
  \begin{minipage}[t]{0.325\linewidth}
    \captionsetup{justification=raggedright,singlelinecheck=false}
    \caption{\textbf{Grouped \oatfamily generation on \libero.}
    Mean \paligemma success (\%). Post-hoc \BAR changes only endpoints;
    \oatpowtwo[16] matches register ordering to blocks.}
    \label{tab:oat_posthoc_bar}
  \end{minipage}
\end{table}

\Cref{tab:oat_posthoc_bar} separates faster generation from tokenizer
compatibility. For \oat[16], switching only the generation pattern reduces the
number of calls from 16 to 5, but success drops from $79.7$ to $66.3$.
\oatpowtwo[16] reaches $80.8$ with the same call count. Grouped generation
therefore works best when the tokenizer's register dependency structure matches
the generation blocks.

%% file: figs/tikz/rate_distortion_curves.tex
% !TEX root = ../../main.tex

\begingroup
\definecolor{rdOAT}{HTML}{8A1538}
\definecolor{rdOATPowTwo}{HTML}{386CB0}
\definecolor{rdBin}{HTML}{7FC97F}
\definecolor{rdFAST}{HTML}{BEAED4}
\definecolor{rdQueST}{HTML}{FDC086}
\definecolor{rdACodec}{HTML}{6A3D9A}
\newcommand{\rdlegendmethod}[2]{%
  {\begingroup
  \renewcommand{\methodfont}[1]{\textcolor{#1}{\texttt{##1}}}#2%
  \endgroup}}
\footnotesize
\tikzset{
  rdplot/.style={
    x=0.50cm,
    y=0.92cm,
    baseline=(current bounding box.south),
    oatline/.style={rdOAT,line width=0.85pt,densely dotted,opacity=1},
    oatpowtwoline/.style={rdOATPowTwo,line width=0.85pt,densely dotted,opacity=1},
    basebin/.style={fill=rdBin,fill opacity=0.5,draw=rdBin,draw opacity=1,line width=0.75pt},
    basefast/.style={fill=rdFAST,fill opacity=0.5,draw=rdFAST,draw opacity=1,line width=0.75pt},
    basequest/.style={fill=rdQueST,fill opacity=0.5,draw=rdQueST,draw opacity=1,line width=0.75pt},
    baseacodec/.style={fill=rdACodec,fill opacity=0.5,draw=rdACodec,draw opacity=1,line width=0.75pt},
    oatmark/.style={fill=rdOAT,fill opacity=0.5,draw=rdOAT,draw opacity=1,line width=0.75pt},
    oatpowtwomark/.style={fill=rdOATPowTwo,fill opacity=1,draw=rdOATPowTwo,draw opacity=1,line width=0.75pt},
  },
}
\newcommand{\rdaxescommon}[1]{%
  \foreach \y in {0.79,1.92} {
    \draw[gray!25,densely dashed] (0,\y) -- (4.75,\y);
    \draw[black!35] (0,\y) -- (-0.025,\y);
    \node[anchor=east,font=\tiny,text=black!45] at (-0.08,\y) {$5$};
  }
  \draw[black!35] (0,3.05) -- (-0.025,3.05);
  \node[anchor=east,font=\tiny,text=black!45] at (-0.08,3.05) {$5$};
  \foreach \y in {0,1.13,2.26} {
    \draw[gray!18] (0,\y) -- (4.75,\y);
  }
  \draw[black!70] (0,0) -- (4.75,0);
  \draw[black!70] (0,0) -- (0,3.05);
  \node[anchor=south,font=\footnotesize] at (2.38,3.18) {#1};
  \foreach \x/\lab in {0/{2^0},1.06/{2^2},2.11/{2^4},3.17/{2^6},4.22/{2^8}} {
    \draw[black!55] (\x,0) -- (\x,-0.04);
    \node[anchor=north,font=\tiny] at (\x,-0.09) {$\lab$};
  }
  \foreach \y/\lab in {0/4,1.13/3,2.26/2} {
    \draw[black!55] (0,\y) -- (-0.04,\y);
    \node[anchor=east,font=\tiny] at (-0.08,\y) {$10^{\text{-}\lab}$};
  }}
\newcommand{\rdaxesmeta}[1]{%
  \foreach \y in {0.71,1.73,2.75} {
    \draw[gray!25,densely dashed] (0,\y) -- (4.75,\y);
    \draw[black!35] (0,\y) -- (-0.025,\y);
    \node[anchor=east,font=\tiny,text=black!45] at (-0.08,\y) {$5$};
  }
  \foreach \y in {1.02,2.03} {
    \draw[gray!18] (0,\y) -- (4.75,\y);
  }
  \draw[black!70] (0,0) -- (4.75,0);
  \draw[black!70] (0,0) -- (0,3.05);
  \node[anchor=south,font=\footnotesize] at (2.38,3.18) {#1};
  \foreach \x/\lab in {0/{2^0},1.06/{2^2},2.11/{2^4},3.17/{2^6},4.22/{2^8}} {
    \draw[black!55] (\x,0) -- (\x,-0.04);
    \node[anchor=north,font=\tiny] at (\x,-0.09) {$\lab$};
  }
  \foreach \y/\lab in {0/3,1.02/2,2.03/1} {
    \draw[black!55] (0,\y) -- (-0.04,\y);
    \node[anchor=east,font=\tiny] at (-0.08,\y) {$10^{\text{-}\lab}$};
  }
  \draw[black!55] (0,3.05) -- (-0.04,3.05);
  \node[anchor=east,font=\tiny] at (-0.08,3.05) {$10^0$};}
\newcommand{\rdaxessimpler}[1]{%
  \foreach \y in {1.07,2.59} {
    \draw[gray!25,densely dashed] (0,\y) -- (4.75,\y);
    \draw[black!35] (0,\y) -- (-0.025,\y);
    \node[anchor=east,font=\tiny,text=black!45] at (-0.08,\y) {$5$};
  }
  \foreach \y in {1.53} {
    \draw[gray!18] (0,\y) -- (4.75,\y);
  }
  \draw[black!70] (0,0) -- (4.75,0);
  \draw[black!70] (0,0) -- (0,3.05);
  \node[anchor=south,font=\footnotesize] at (2.38,3.18) {#1};
  \foreach \x/\lab in {0/{2^0},1.06/{2^2},2.11/{2^4},3.17/{2^6},4.22/{2^8}} {
    \draw[black!55] (\x,0) -- (\x,-0.04);
    \node[anchor=north,font=\tiny] at (\x,-0.09) {$\lab$};
  }
  \foreach \y/\lab in {0/5,1.53/4,3.05/3} {
    \draw[black!55] (0,\y) -- (-0.04,\y);
    \node[anchor=east,font=\tiny] at (-0.08,\y) {$10^{\text{-}\lab}$};
  }}
% Generated by scripts/rate_distortion_coords.py; paste into the rate--distortion figure if values change.
  \newcommand{\rdliberodata}{%
    \draw[oatline] (0.00,2.45) -- (0.53,2.11) -- (1.06,1.79) -- (1.58,1.49) -- (2.11,1.13);
    \draw[oatpowtwoline] (0.00,2.50) -- (0.53,2.11) -- (1.06,1.76) -- (1.58,1.45) -- (2.11,1.08);
    \foreach \p in {(0.00,2.45),(0.53,2.11),(1.06,1.79),(1.58,1.49),(2.11,1.13)} \filldraw[oatmark] \p circle (2.25pt);
    \foreach \p in {(0.00,2.50),(0.53,2.11),(1.06,1.76),(1.58,1.45),(2.11,1.08)} \filldraw[oatpowtwomark] \p circle (0.98pt);
    \filldraw[basebin] (4.12,0.00) circle (2.25pt);
    \filldraw[basefast] (2.94,0.54) circle (2.25pt);
    \filldraw[basequest] (2.11,1.64) circle (2.25pt);
    \filldraw[baseacodec] (2.11,0.96) circle (2.25pt);}
  \newcommand{\rdrobomimicdata}{%
    \draw[oatline] (0.00,2.37) -- (0.53,2.14) -- (1.06,1.95) -- (1.58,1.69) -- (2.11,1.45);
    \draw[oatpowtwoline] (0.00,2.39) -- (0.53,2.14) -- (1.06,1.95) -- (1.58,1.67) -- (2.11,1.42);
    \foreach \p in {(0.00,2.37),(0.53,2.14),(1.06,1.95),(1.58,1.69),(2.11,1.45)} \filldraw[oatmark] \p circle (2.25pt);
    \foreach \p in {(0.00,2.39),(0.53,2.14),(1.06,1.95),(1.58,1.67),(2.11,1.42)} \filldraw[oatpowtwomark] \p circle (0.98pt);
    \filldraw[basebin] (4.12,0.00) circle (2.25pt);
    \filldraw[basefast] (2.99,0.68) circle (2.25pt);
    \filldraw[basequest] (2.11,1.69) circle (2.25pt);
    \filldraw[baseacodec] (2.11,1.54) circle (2.25pt);}
  \newcommand{\rdmetadata}{%
    \draw[oatline] (0.00,2.91) -- (0.53,2.08) -- (1.06,1.61) -- (1.58,1.54) -- (2.11,1.53);
    \draw[oatpowtwoline] (0.00,2.91) -- (0.53,2.06) -- (1.06,1.67) -- (1.58,1.47) -- (2.11,1.45);
    \foreach \p in {(0.00,2.91),(0.53,2.08),(1.06,1.61),(1.58,1.54),(2.11,1.53)} \filldraw[oatmark] \p circle (2.25pt);
    \foreach \p in {(0.00,2.91),(0.53,2.06),(1.06,1.67),(1.58,1.47),(2.11,1.45)} \filldraw[oatpowtwomark] \p circle (0.98pt);
    \filldraw[basebin] (3.69,0.00) circle (2.25pt);
    \filldraw[basefast] (2.30,1.86) circle (2.25pt);
    \filldraw[basequest] (2.11,1.42) circle (2.25pt);
    \filldraw[baseacodec] (2.11,0.93) circle (2.25pt);}
  \newcommand{\rdrobocasadata}{%
    \draw[oatline] (0.00,2.49) -- (0.53,2.20) -- (1.06,1.94) -- (1.58,1.58) -- (2.11,1.26);
    \draw[oatpowtwoline] (0.00,2.50) -- (0.53,2.20) -- (1.06,1.91) -- (1.58,1.56) -- (2.11,1.22);
    \foreach \p in {(0.00,2.49),(0.53,2.20),(1.06,1.94),(1.58,1.58),(2.11,1.26)} \filldraw[oatmark] \p circle (2.25pt);
    \foreach \p in {(0.00,2.50),(0.53,2.20),(1.06,1.91),(1.58,1.56),(2.11,1.22)} \filldraw[oatpowtwomark] \p circle (0.98pt);
    \filldraw[basebin] (4.53,0.00) circle (2.25pt);
    \filldraw[basefast] (2.85,0.34) circle (2.25pt);
    \filldraw[basequest] (2.11,1.33) circle (2.25pt);
    \filldraw[baseacodec] (2.11,1.30) circle (2.25pt);}
  \newcommand{\rdsimplerdata}{%
    \draw[oatline] (0.00,2.91) -- (0.53,2.35) -- (1.06,2.02) -- (1.58,1.59) -- (2.11,1.07);
    \draw[oatpowtwoline] (0.00,2.92) -- (0.53,2.35) -- (1.06,1.98) -- (1.58,1.59) -- (2.11,1.19);
    \foreach \p in {(0.00,2.91),(0.53,2.35),(1.06,2.02),(1.58,1.59),(2.11,1.07)} \filldraw[oatmark] \p circle (2.25pt);
    \foreach \p in {(0.00,2.92),(0.53,2.35),(1.06,1.98),(1.58,1.59),(2.11,1.19)} \filldraw[oatpowtwomark] \p circle (0.98pt);
    \filldraw[basebin] (3.06,0.00) circle (2.25pt);
    \filldraw[basefast] (1.82,1.95) circle (2.25pt);
    \filldraw[basequest] (1.06,3.01) circle (2.25pt);
    \filldraw[baseacodec] (2.11,0.92) circle (2.25pt);}
\noindent
\begin{tikzpicture}[rdplot]
  \rdaxescommon{\libero}
  \node[anchor=south,font=\tiny] at (0,3.12) {MSE};
  \rdliberodata
\end{tikzpicture}\hfill
\begin{tikzpicture}[rdplot]
  \rdaxescommon{\robomimic}
  \rdrobomimicdata
\end{tikzpicture}\hfill
\begin{tikzpicture}[rdplot]
  \rdaxesmeta{\metaworld}
  \rdmetadata
\end{tikzpicture}\hfill
\begin{tikzpicture}[rdplot]
  \rdaxescommon{\robocasaThreeSixFive}
  \rdrobocasadata
\end{tikzpicture}\hfill
\begin{tikzpicture}[rdplot]
  \rdaxessimpler{\simpler}
  \rdsimplerdata
  \node[anchor=west,font=\tiny] at (4.80,0) {$H_l$};
\end{tikzpicture}
\par\vspace{0.6ex}
\noindent
\begin{tikzpicture}[rdplot]
  \filldraw[oatmark] (0.80,0) circle (2.25pt);
  \node[anchor=west,font=\footnotesize] at (1.20,0) {\rdlegendmethod{rdOAT}{\oat}};
  \filldraw[oatpowtwomark] (5.50,0) circle (0.98pt);
  \node[anchor=west,font=\footnotesize] at (5.90,0) {\rdlegendmethod{rdOATPowTwo}{\oatpowtwo}};
  \filldraw[basebin] (9.95,0) circle (2.25pt);
  \node[anchor=west,font=\footnotesize] at (10.30,0) {\textcolor{rdBin}{\bin}};
  \filldraw[basefast] (14.45,0) circle (2.25pt);
  \node[anchor=west,font=\footnotesize] at (14.80,0) {\textcolor{rdFAST}{\fast}};
  \filldraw[basequest] (19.30,0) circle (2.25pt);
  \node[anchor=west,font=\footnotesize] at (19.65,0) {\textcolor{rdQueST}{\quest}};
  \filldraw[baseacodec] (24.60,0) circle (2.25pt);
  \node[anchor=west,font=\footnotesize] at (24.95,0) {\textcolor{rdACodec}{\acodec}};
\end{tikzpicture}
\endgroup

%% file: figs/tikz/vlm_bar_success_grid.tex
% !TEX root = ../../main.tex

\begingroup
\definecolor{rdOAT}{HTML}{8A1538}
\definecolor{rdOATPowTwo}{HTML}{386CB0}
\definecolor{rdBin}{HTML}{7FC97F}
\definecolor{rdFAST}{HTML}{BEAED4}
\definecolor{rdQueST}{HTML}{FDC086}
\definecolor{rdACodec}{HTML}{6A3D9A}
\definecolor{vlmSep}{HTML}{F5F5F5}
\newcommand{\rdlegendmethod}[2]{%
  {\begingroup
  \renewcommand{\methodfont}[1]{\textcolor{#1}{\texttt{##1}}}#2%
  \endgroup}}
\footnotesize
\tikzset{
  rdplot/.style={
    x=0.50cm,
    y=0.92cm,
    baseline=(current bounding box.south),
    oatline/.style={rdOAT,line width=0.85pt,densely dotted,opacity=1},
    oatpowtwoline/.style={rdOATPowTwo,line width=0.85pt,densely dotted,opacity=1},
      basebin/.style={fill=rdBin,fill opacity=0.5,draw=rdBin,draw opacity=1,line width=0.75pt},
      basefast/.style={fill=rdFAST,fill opacity=0.5,draw=rdFAST,draw opacity=1,line width=0.75pt},
      basequest/.style={fill=rdQueST,fill opacity=0.5,draw=rdQueST,draw opacity=1,line width=0.75pt},
      baseacodec/.style={fill=rdACodec,fill opacity=0.5,draw=rdACodec,draw opacity=1,line width=0.75pt},
      oatmark/.style={fill=rdOAT,fill opacity=0.5,draw=rdOAT,draw opacity=1,line width=0.75pt},
      oatpowtwomark/.style={fill=rdOATPowTwo,fill opacity=1,draw=rdOATPowTwo,draw opacity=1,line width=0.75pt},
  },
  vlmplot/.style={
    x=0.50cm,
    y=0.0285cm,
    baseline=(current bounding box.south),
  },
  vlmsummaryplot/.style={
    x=0.50cm,
    y=0.0285cm,
    baseline=(current bounding box.south),
  },
  vlmrowtagplot/.style={
    x=1pt,
    y=0.0285cm,
    baseline=(current bounding box.south),
  },
  vlmsepplot/.style={
    x=1pt,
    y=0.0285cm,
    baseline=(current bounding box.south),
  },
  oatline/.style={rdOAT,line width=0.85pt,densely dotted,opacity=1},
  oatpowtwoline/.style={rdOATPowTwo,line width=0.85pt,densely dotted,opacity=1},
  basebin/.style={fill=rdBin,fill opacity=0.5,draw=rdBin,draw opacity=1,line width=0.75pt},
  basefast/.style={fill=rdFAST,fill opacity=0.5,draw=rdFAST,draw opacity=1,line width=0.75pt},
  basequest/.style={fill=rdQueST,fill opacity=0.5,draw=rdQueST,draw opacity=1,line width=0.75pt},
  baseacodec/.style={fill=rdACodec,fill opacity=0.5,draw=rdACodec,draw opacity=1,line width=0.75pt},
  oatmark/.style={fill=rdOAT,fill opacity=0.5,draw=rdOAT,draw opacity=1,line width=0.75pt},
  oatpowtwomark/.style={fill=rdOATPowTwo,fill opacity=1,draw=rdOATPowTwo,draw opacity=1,line width=0.75pt},
  vlmseparator/.style={vlmSep,line width=1.0pt},
  vlmaxis/.style={black!70,line width=0.45pt},
  vlmtick/.style={black!55,line width=0.35pt},
  vlmgrid/.style={gray!18,line width=0.35pt},
}
\newcommand{\vlmbbox}{%
  \path[use as bounding box] (-0.55,-13.4) rectangle (4.80,91.5);}
\newcommand{\vlmxaxis}{%
  \foreach \x/\lab in {0/{2^0},1.06/{2^2},2.11/{2^4},3.17/{2^6},4.22/{2^8}} {
    \draw[vlmtick] (\x,0) -- (\x,-1.25);
    \node[anchor=north,font=\tiny] at (\x,-2.80) {$\lab$};
  }}
\newcommand{\vlmframe}[2]{%
  \vlmbbox
  \foreach \y/\lab in {#2} {
    \draw[vlmgrid] (-0.08,\y) -- (4.62,\y);
    \draw[vlmtick] (-0.08,\y) -- (-0.15,\y);
    \node[anchor=east,font=\tiny] at (-0.20,\y) {$\lab$};
  }
  \draw[vlmaxis] (-0.08,0) -- (4.62,0);
  \draw[vlmaxis] (-0.08,0) -- (-0.08,80);
  \node[anchor=south,font=\footnotesize] at (2.27,84.5) {#1};
  \vlmxaxis}
\newcommand{\vlmframenoheading}[1]{%
  \vlmbbox
  \foreach \y/\lab in {#1} {
    \draw[vlmgrid] (-0.08,\y) -- (4.62,\y);
    \draw[vlmtick] (-0.08,\y) -- (-0.15,\y);
    \node[anchor=east,font=\tiny] at (-0.20,\y) {$\lab$};
  }
  \draw[vlmaxis] (-0.08,0) -- (4.62,0);
  \draw[vlmaxis] (-0.08,0) -- (-0.08,80);
  \vlmxaxis}
\newcommand{\vlmrowtag}[1]{%
  \begin{tikzpicture}[vlmrowtagplot]
    \path[use as bounding box] (0,-13.4) rectangle (6.0,91.5);
    \node[anchor=center,rotate=90,font=\footnotesize\bfseries] at (3.6,40) {#1};
  \end{tikzpicture}}
\newcommand{\vlmylabel}{%
  \node[anchor=south east,font=\tiny] at (0.15,84.5) {Success};}
\newcommand{\vlmblank}{%
  \fill[black!3] (-0.08,0) rectangle (4.62,80);
  \node[font=\scriptsize,text=black!45] at (2.27,40) {not populated};}
\newcommand{\vlmsummarysep}{%
  \begin{tikzpicture}[vlmsepplot]
    \path[use as bounding box] (-0.5,-13.4) rectangle (0.5,91.5);
    \draw[vlmseparator] (0,-1.0) -- (0,84.5);
  \end{tikzpicture}}
\newcommand{\vlmrowgap}{\hspace{8.0pt}}
\newcommand{\vlmfirstgap}{\hspace{11.2pt}}
\newcommand{\vlmsepcompensate}{\hspace{-10.4pt}}
\newcommand{\vlmrowwidth}{\dimexpr\linewidth-2pt\relax}

% Coordinates below are generated by scripts/vlm_success_coords.py.
\noindent
\begingroup
\setlength{\tabcolsep}{0pt}
\begin{tabular*}{\vlmrowwidth}{@{}c@{\vlmfirstgap}c@{\vlmrowgap}c@{\vlmrowgap}c@{\vlmrowgap}c@{\extracolsep{\fill}\vlmsepcompensate}c@{\extracolsep{\fill}}c@{}}
\vlmrowtag{\paligemma} &
  \begin{tikzpicture}[vlmplot]
    \vlmframe{\libero}{0/10,10/20,20/30,30/40,40/50,50/60,60/70,70/80,80/90}
    \vlmylabel
    % PaliGemma2 / LIBERO
    \draw[oatline] (0.00,18.20) -- (0.53,62.50) -- (1.06,63.40) -- (1.58,68.00) -- (2.11,69.70);
    \foreach \p in {(0.00,18.20),(0.53,62.50),(1.06,63.40),(1.58,68.00),(2.11,69.70)} \filldraw[oatmark] \p circle (2.25pt);
    \draw[oatpowtwoline] (0.00,21.40) -- (0.53,63.70) -- (1.06,66.30) -- (1.58,68.20) -- (2.11,70.80);
    \foreach \p in {(0.00,21.40),(0.53,63.70),(1.06,66.30),(1.58,68.20),(2.11,70.80)} \filldraw[oatpowtwomark] \p circle (0.98pt);
    \filldraw[basebin] (4.12,4.20) circle (2.25pt);
    \filldraw[basefast] (2.94,60.20) circle (2.25pt);
    \filldraw[basequest] (2.11,55.90) circle (2.25pt);
    \filldraw[baseacodec] (2.11,68.70) circle (2.25pt);
  \end{tikzpicture} &
  \begin{tikzpicture}[vlmplot]
    \vlmframe{\robomimic}{0/0,11.43/10,22.86/20,34.29/30,45.71/40,57.14/50,68.57/60,80/70}
    % PaliGemma2 / RoboMimic
    \draw[oatline] (0.00,11.43) -- (0.53,33.14) -- (1.06,48.00) -- (1.58,74.29) -- (2.11,75.43);
    \foreach \p in {(0.00,11.43),(0.53,33.14),(1.06,48.00),(1.58,74.29),(2.11,75.43)} \filldraw[oatmark] \p circle (2.25pt);
    \draw[oatpowtwoline] (0.00,9.71) -- (0.53,28.00) -- (1.06,54.86) -- (1.58,61.14) -- (2.11,68.57);
    \foreach \p in {(0.00,9.71),(0.53,28.00),(1.06,54.86),(1.58,61.14),(2.11,68.57)} \filldraw[oatpowtwomark] \p circle (0.98pt);
    \filldraw[basebin] (4.12,8.00) circle (2.25pt);
    \filldraw[basefast] (2.99,70.86) circle (2.25pt);
    \filldraw[basequest] (2.11,73.71) circle (2.25pt);
    \filldraw[baseacodec] (2.11,61.71) circle (2.25pt);
  \end{tikzpicture} &
  \begin{tikzpicture}[vlmplot]
    \vlmframe{\robocasaThreeSixFive}{0/0,11.43/10,22.86/20,34.29/30,45.71/40,57.14/50,68.57/60,80/70}
    % PaliGemma2 / RoboCasa365
    \draw[oatline] (0.00,28.34) -- (0.53,51.20) -- (1.06,54.86) -- (1.58,67.20) -- (2.11,69.03);
    \foreach \p in {(0.00,28.34),(0.53,51.20),(1.06,54.86),(1.58,67.20),(2.11,69.03)} \filldraw[oatmark] \p circle (2.25pt);
    \draw[oatpowtwoline] (0.00,32.46) -- (0.53,56.23) -- (1.06,71.31) -- (1.58,71.77) -- (2.11,69.03);
    \foreach \p in {(0.00,32.46),(0.53,56.23),(1.06,71.31),(1.58,71.77),(2.11,69.03)} \filldraw[oatpowtwomark] \p circle (0.98pt);
    \filldraw[basebin] (4.53,3.66) circle (2.25pt);
    \filldraw[basefast] (2.85,50.74) circle (2.25pt);
    \filldraw[basequest] (2.11,56.69) circle (2.25pt);
    \filldraw[baseacodec] (2.11,58.51) circle (2.25pt);
  \end{tikzpicture} &
  \begin{tikzpicture}[vlmplot]
    \vlmframe{\simpler}{0/0,13.33/10,26.67/20,40/30,53.33/40,66.67/50,80/60}
    % PaliGemma2 / SimplerEnv
    \draw[oatline] (0.00,21.33) -- (0.53,47.33) -- (1.06,47.33) -- (1.58,52.00) -- (2.11,64.67);
    \foreach \p in {(0.00,21.33),(0.53,47.33),(1.06,47.33),(1.58,52.00),(2.11,64.67)} \filldraw[oatmark] \p circle (2.25pt);
    \draw[oatpowtwoline] (0.00,24.00) -- (0.53,59.33) -- (1.06,65.33) -- (1.58,75.33) -- (2.11,72.00);
    \foreach \p in {(0.00,24.00),(0.53,59.33),(1.06,65.33),(1.58,75.33),(2.11,72.00)} \filldraw[oatpowtwomark] \p circle (0.98pt);
    \filldraw[basebin] (3.06,27.33) circle (2.25pt);
    \filldraw[basefast] (1.82,22.67) circle (2.25pt);
    \filldraw[basequest] (1.06,72.00) circle (2.25pt);
    \filldraw[baseacodec] (2.11,35.33) circle (2.25pt);
  \end{tikzpicture} &
  \vlmsummarysep &
  \begin{tikzpicture}[vlmsummaryplot]
    \vlmframe{Avg. Success}{0/0,11.43/10,22.86/20,34.29/30,45.71/40,57.14/50,68.57/60,80/70}
    % PaliGemma2 / Avg. Success
    \draw[oatline] (0.00,22.57) -- (0.53,51.94) -- (1.06,56.83) -- (1.58,68.80) -- (2.11,72.74);
    \foreach \p in {(0.00,22.57),(0.53,51.94),(1.06,56.83),(1.58,68.80),(2.11,72.74)} \filldraw[oatmark] \p circle (2.25pt);
    \draw[oatpowtwoline] (0.00,24.66) -- (0.53,54.83) -- (1.06,67.34) -- (1.58,71.71) -- (2.11,72.91);
    \foreach \p in {(0.00,24.66),(0.53,54.83),(1.06,67.34),(1.58,71.71),(2.11,72.91)} \filldraw[oatpowtwomark] \p circle (0.98pt);
    \filldraw[basebin] (4.11,12.83) circle (2.25pt);
    \filldraw[basefast] (2.77,55.31) circle (2.25pt);
    \filldraw[basequest] (1.95,66.86) circle (2.25pt);
    \filldraw[baseacodec] (2.11,60.11) circle (2.25pt);
  \end{tikzpicture}
\end{tabular*}
\endgroup

\par\vspace{1.2ex}
\noindent
\begingroup
\setlength{\tabcolsep}{0pt}
\begin{tabular*}{\vlmrowwidth}{@{}c@{\vlmfirstgap}c@{\vlmrowgap}c@{\vlmrowgap}c@{\vlmrowgap}c@{\extracolsep{\fill}\vlmsepcompensate}c@{\extracolsep{\fill}}c@{}}
\vlmrowtag{\qwenvl} &
  \begin{tikzpicture}[vlmplot]
    \vlmframenoheading{0/0,8.89/10,17.78/20,26.67/30,35.56/40,44.44/50,53.33/60,62.22/70,71.11/80,80/90}
    \vlmylabel
    % Qwen3VL / LIBERO
    \draw[oatline] (0.00,29.33) -- (0.53,63.02) -- (1.06,67.91) -- (1.58,69.78) -- (2.11,72.89);
    \foreach \p in {(0.00,29.33),(0.53,63.02),(1.06,67.91),(1.58,69.78),(2.11,72.89)} \filldraw[oatmark] \p circle (2.25pt);
    \draw[oatpowtwoline] (0.00,27.64) -- (0.53,64.62) -- (1.06,70.04) -- (1.58,70.49) -- (2.11,72.71);
    \foreach \p in {(0.00,27.64),(0.53,64.62),(1.06,70.04),(1.58,70.49),(2.11,72.71)} \filldraw[oatpowtwomark] \p circle (0.98pt);
    \filldraw[basebin] (4.12,0.00) circle (2.25pt);
    \filldraw[basefast] (2.94,55.82) circle (2.25pt);
    \filldraw[basequest] (2.11,51.91) circle (2.25pt);
    \filldraw[baseacodec] (2.11,68.00) circle (2.25pt);
  \end{tikzpicture} &
  \begin{tikzpicture}[vlmplot]
    \vlmframenoheading{0/0,11.43/10,22.86/20,34.29/30,45.71/40,57.14/50,68.57/60,80/70}
    % Qwen3VL / RoboMimic
    \draw[oatline] (0.00,13.71) -- (0.53,33.71) -- (1.06,39.43) -- (1.58,60.00) -- (2.11,76.57);
    \foreach \p in {(0.00,13.71),(0.53,33.71),(1.06,39.43),(1.58,60.00),(2.11,76.57)} \filldraw[oatmark] \p circle (2.25pt);
    \draw[oatpowtwoline] (0.00,9.71) -- (0.53,22.86) -- (1.06,35.43) -- (1.58,53.71) -- (2.11,54.86);
    \foreach \p in {(0.00,9.71),(0.53,22.86),(1.06,35.43),(1.58,53.71),(2.11,54.86)} \filldraw[oatpowtwomark] \p circle (0.98pt);
    \filldraw[basebin] (4.12,0.57) circle (2.25pt);
    \filldraw[basefast] (2.99,31.43) circle (2.25pt);
    \filldraw[basequest] (2.11,56.00) circle (2.25pt);
    \filldraw[baseacodec] (2.11,72.57) circle (2.25pt);
  \end{tikzpicture} &
  \begin{tikzpicture}[vlmplot]
    \vlmframenoheading{0/0,11.43/10,22.86/20,34.29/30,45.71/40,57.14/50,68.57/60,80/70}
    % Qwen3VL / RoboCasa365
    \draw[oatline] (0.00,31.09) -- (0.53,55.31) -- (1.06,63.09) -- (1.58,65.83) -- (2.11,70.86);
    \foreach \p in {(0.00,31.09),(0.53,55.31),(1.06,63.09),(1.58,65.83),(2.11,70.86)} \filldraw[oatmark] \p circle (2.25pt);
    \draw[oatpowtwoline] (0.00,37.49) -- (0.53,59.89) -- (1.06,60.80) -- (1.58,63.54) -- (2.11,65.83);
    \foreach \p in {(0.00,37.49),(0.53,59.89),(1.06,60.80),(1.58,63.54),(2.11,65.83)} \filldraw[oatpowtwomark] \p circle (0.98pt);
    \filldraw[basebin] (4.53,1.83) circle (2.25pt);
    \filldraw[basefast] (2.85,37.49) circle (2.25pt);
    \filldraw[basequest] (2.11,16.91) circle (2.25pt);
    \filldraw[baseacodec] (2.11,51.66) circle (2.25pt);
  \end{tikzpicture} &
  \begin{tikzpicture}[vlmplot]
    \vlmframenoheading{0/0,26.67/10,53.33/20,80/30}
    % Qwen3VL / SimplerEnv
    \draw[oatline] (0.00,14.67) -- (0.53,29.33) -- (1.06,32.00) -- (1.58,36.00) -- (2.11,42.67);
    \foreach \p in {(0.00,14.67),(0.53,29.33),(1.06,32.00),(1.58,36.00),(2.11,42.67)} \filldraw[oatmark] \p circle (2.25pt);
    \draw[oatpowtwoline] (0.00,9.33) -- (0.53,34.67) -- (1.06,40.00) -- (1.58,50.67) -- (2.11,40.00);
    \foreach \p in {(0.00,9.33),(0.53,34.67),(1.06,40.00),(1.58,50.67),(2.11,40.00)} \filldraw[oatpowtwomark] \p circle (0.98pt);
    \filldraw[basebin] (3.06,2.67) circle (2.25pt);
    \filldraw[basefast] (1.82,13.33) circle (2.25pt);
    \filldraw[basequest] (1.06,16.00) circle (2.25pt);
    \filldraw[baseacodec] (2.11,62.67) circle (2.25pt);
  \end{tikzpicture} &
  \vlmsummarysep &
  \begin{tikzpicture}[vlmsummaryplot]
    \vlmframenoheading{0/0,13.33/10,26.67/20,40/30,53.33/40,66.67/50,80/60}
    \vlmylabel
    % Qwen3VL / Avg. Success
    \draw[oatline] (0.00,25.90) -- (0.53,53.27) -- (1.06,59.37) -- (1.58,67.37) -- (2.11,75.67);
    \foreach \p in {(0.00,25.90),(0.53,53.27),(1.06,59.37),(1.58,67.37),(2.11,75.67)} \filldraw[oatmark] \p circle (2.25pt);
    \draw[oatpowtwoline] (0.00,25.30) -- (0.53,52.70) -- (1.06,59.33) -- (1.58,66.97) -- (2.11,67.47);
    \foreach \p in {(0.00,25.30),(0.53,52.70),(1.06,59.33),(1.58,66.97),(2.11,67.47)} \filldraw[oatpowtwomark] \p circle (0.98pt);
    \filldraw[basebin] (4.11,1.03) circle (2.25pt);
    \filldraw[basefast] (2.77,42.70) circle (2.25pt);
    \filldraw[basequest] (1.95,42.73) circle (2.25pt);
    \filldraw[baseacodec] (2.11,69.57) circle (2.25pt);
  \end{tikzpicture}
\end{tabular*}
\endgroup

\par\vspace{0.6ex}
\noindent
\begin{tikzpicture}[rdplot]
  \filldraw[oatmark] (0.80,0) circle (2.25pt);
  \node[anchor=west,font=\footnotesize] at (1.20,0) {\rdlegendmethod{rdOAT}{\oat}};
  \filldraw[oatpowtwomark] (5.50,0) circle (0.98pt);
  \node[anchor=west,font=\footnotesize] at (5.90,0) {\rdlegendmethod{rdOATPowTwo}{\oatpowtwo}};
  \filldraw[basebin] (9.95,0) circle (2.25pt);
  \node[anchor=west,font=\footnotesize] at (10.30,0) {\textcolor{rdBin}{\bin}};
  \filldraw[basefast] (14.45,0) circle (2.25pt);
  \node[anchor=west,font=\footnotesize] at (14.80,0) {\textcolor{rdFAST}{\fast}};
  \filldraw[basequest] (19.30,0) circle (2.25pt);
  \node[anchor=west,font=\footnotesize] at (19.65,0) {\textcolor{rdQueST}{\quest}};
  \filldraw[baseacodec] (24.60,0) circle (2.25pt);
  \node[anchor=west,font=\footnotesize] at (24.95,0) {\textcolor{rdACodec}{\acodec}};
\end{tikzpicture}
\endgroup

%% file: figs/tikz/vlm_ki_success_grid.tex
% !TEX root = ../../main.tex

\begingroup
\definecolor{rdOAT}{HTML}{8A1538}
\definecolor{rdBin}{HTML}{7FC97F}
\definecolor{rdFAST}{HTML}{BEAED4}
\definecolor{rdQueST}{HTML}{FDC086}
\definecolor{rdACodec}{HTML}{6A3D9A}
\definecolor{vlmSep}{HTML}{F5F5F5}
\footnotesize
\tikzset{
  vlmplot/.style={
    x=0.50cm,
    y=0.0285cm,
    baseline=(current bounding box.south),
  },
  vlmsummaryplot/.style={
    x=0.50cm,
    y=0.0285cm,
    baseline=(current bounding box.south),
  },
  vlmrowtagplot/.style={
    x=1pt,
    y=0.0285cm,
    baseline=(current bounding box.south),
  },
  vlmsepplot/.style={
    x=1pt,
    y=0.0285cm,
    baseline=(current bounding box.south),
  },
  vlmseparator/.style={vlmSep,line width=1.0pt},
  vlmaxis/.style={black!70,line width=0.45pt},
  vlmtick/.style={black!55,line width=0.35pt},
  vlmgrid/.style={gray!18,line width=0.35pt},
  vlmhalfgrid/.style={gray!25,densely dashed,line width=0.35pt},
  vlmbarvalue/.style={font=\tiny,text=black!75,inner sep=0.2pt},
  kioatbar/.style={fill=rdOAT,fill opacity=0.62,draw=rdOAT,draw opacity=1,line width=0.45pt},
  kibinbar/.style={fill=rdBin,fill opacity=0.62,draw=rdBin,draw opacity=1,line width=0.45pt},
  kifastbar/.style={fill=rdFAST,fill opacity=0.70,draw=rdFAST,draw opacity=1,line width=0.45pt},
  kiquestbar/.style={fill=rdQueST,fill opacity=0.70,draw=rdQueST,draw opacity=1,line width=0.45pt},
  kiacodecbar/.style={fill=rdACodec,fill opacity=0.62,draw=rdACodec,draw opacity=1,line width=0.45pt},
}
\newcommand{\vlmkibbox}{%
  \path[use as bounding box] (-0.55,-21.0) rectangle (4.80,91.5);}
\newcommand{\vlmkixaxis}{%
  \foreach \x/\lab in {0.38/{\oatfamily},1.34/{\textcolor{rdBin}{\bin}},2.30/{\textcolor{rdFAST}{\fast}},3.26/{\textcolor{rdQueST}{\quest}},4.22/{\textcolor{rdACodec}{\acodec}}} {
    \draw[vlmtick] (\x,0) -- (\x,-1.25);
    \node[anchor=east,rotate=45,font=\tiny] at (\x,-2.60) {\lab};
  }}
\newcommand{\vlmkiframe}[2]{%
  \vlmkibbox
  \foreach \y/\lab in {#2} {
    \draw[vlmgrid] (-0.08,\y) -- (4.62,\y);
    \draw[vlmtick] (-0.08,\y) -- (-0.15,\y);
    \node[anchor=east,font=\tiny] at (-0.20,\y) {$\lab$};
  }
  \draw[vlmaxis] (-0.08,0) -- (4.62,0);
  \draw[vlmaxis] (-0.08,0) -- (-0.08,80);
  \node[anchor=south,font=\footnotesize] at (2.27,84.5) {#1};
  \vlmkixaxis}
\newcommand{\vlmkiframenoheading}[1]{%
  \vlmkibbox
  \foreach \y/\lab in {#1} {
    \draw[vlmgrid] (-0.08,\y) -- (4.62,\y);
    \draw[vlmtick] (-0.08,\y) -- (-0.15,\y);
    \node[anchor=east,font=\tiny] at (-0.20,\y) {$\lab$};
  }
  \draw[vlmaxis] (-0.08,0) -- (4.62,0);
  \draw[vlmaxis] (-0.08,0) -- (-0.08,80);
  \vlmkixaxis}
\newcommand{\vlmkirowtag}[1]{%
  \begin{tikzpicture}[vlmrowtagplot]
    \path[use as bounding box] (0,-21.0) rectangle (6.0,91.5);
    \node[anchor=center,rotate=90,font=\footnotesize\bfseries] at (3.6,40) {#1};
  \end{tikzpicture}}
\newcommand{\vlmkiylabel}{%
  \node[anchor=south east,font=\tiny] at (0.15,84.5) {Success};}
\newcommand{\vlmkisummarysep}{%
  \begin{tikzpicture}[vlmsepplot]
    \path[use as bounding box] (-0.5,-21.0) rectangle (0.5,91.5);
    \draw[vlmseparator] (0,-1.0) -- (0,84.5);
  \end{tikzpicture}}
\newcommand{\vlmkirowgap}{\hspace{8.0pt}}
\newcommand{\vlmkifirstgap}{\hspace{11.2pt}}
\newcommand{\vlmkisepcompensate}{\hspace{-10.4pt}}
\newcommand{\vlmkirowwidth}{\dimexpr\linewidth-2pt\relax}

\noindent
\begingroup
\setlength{\tabcolsep}{0pt}
\begin{tabular*}{\vlmkirowwidth}{@{}c@{\vlmkifirstgap}c@{\vlmkirowgap}c@{\vlmkirowgap}c@{\vlmkirowgap}c@{\extracolsep{\fill}\vlmkisepcompensate}c@{\extracolsep{\fill}}c@{}}
\vlmkirowtag{\paligemma} &
  \begin{tikzpicture}[vlmplot]
    \vlmkiframe{\libero}{0/85,80/90}
    \vlmkiylabel
    % LIBERO TC success rates: OAT, Bin, FAST, QueST, ACodec.
    \draw[vlmhalfgrid] (-0.08,40) -- (4.62,40);
    \filldraw[kioatbar] (0.12,0) rectangle (0.64,75.2);
    \node[vlmbarvalue,anchor=south] at (0.38,76.4) {89.7};
    \filldraw[kibinbar] (1.08,0) rectangle (1.60,30.4);
    \node[vlmbarvalue,anchor=south] at (1.34,31.6) {86.9};
    \filldraw[kifastbar] (2.04,0) rectangle (2.56,52.8);
    \node[vlmbarvalue,anchor=south] at (2.30,54.0) {88.3};
    \filldraw[kiquestbar] (3.00,0) rectangle (3.52,70.4);
    \node[vlmbarvalue,anchor=south] at (3.26,71.6) {89.4};
    \filldraw[kiacodecbar] (3.96,0) rectangle (4.48,48.0);
    \node[vlmbarvalue,anchor=south] at (4.22,49.2) {88.0};
  \end{tikzpicture} &
  \begin{tikzpicture}[vlmplot]
    \vlmkiframe{\robomimic}{0/55,26.67/60,53.33/65,80/70}
    % RoboMimic TC success rates: OAT, Bin, FAST, QueST, ACodec.
    \filldraw[kioatbar] (0.12,0) rectangle (0.64,32.0);
    \node[vlmbarvalue,anchor=south] at (0.38,33.2) {61.0};
    \filldraw[kibinbar] (1.08,0) rectangle (1.60,18.7);
    \node[vlmbarvalue,anchor=south] at (1.34,19.9) {58.5};
    \filldraw[kifastbar] (2.04,0) rectangle (2.56,26.7);
    \node[vlmbarvalue,anchor=south] at (2.30,27.9) {60.0};
    \filldraw[kiquestbar] (3.00,0) rectangle (3.52,72.0);
    \node[vlmbarvalue,anchor=south] at (3.26,73.2) {68.5};
    \filldraw[kiacodecbar] (3.96,0) rectangle (4.48,29.3);
    \node[vlmbarvalue,anchor=south] at (4.22,30.5) {60.5};
  \end{tikzpicture} &
  \begin{tikzpicture}[vlmplot]
    \vlmkiframe{\robocasaThreeSixFive}{0/30,16/35,32/40,48/45,64/50,80/55}
    % RoboCasa365 TC success rates: OAT, Bin, FAST, QueST, ACodec.
    \filldraw[kioatbar] (0.12,0) rectangle (0.64,67.8);
    \node[vlmbarvalue,anchor=south] at (0.38,69.0) {51.2};
    \filldraw[kibinbar] (1.08,0) rectangle (1.60,15.4);
    \node[vlmbarvalue,anchor=south] at (1.34,16.6) {34.8};
    \filldraw[kifastbar] (2.04,0) rectangle (2.56,53.8);
    \node[vlmbarvalue,anchor=south] at (2.30,55.0) {46.8};
    \filldraw[kiquestbar] (3.00,0) rectangle (3.52,53.8);
    \node[vlmbarvalue,anchor=south] at (3.26,55.0) {46.8};
    \filldraw[kiacodecbar] (3.96,0) rectangle (4.48,69.1);
    \node[vlmbarvalue,anchor=south] at (4.22,70.3) {51.6};
  \end{tikzpicture} &
  \begin{tikzpicture}[vlmplot]
    \vlmkiframe{\simpler}{0/25,40/30,80/35}
    % SimplerEnv / Bridge TC success rates: OAT, Bin, FAST, QueST, ACodec.
    \draw[vlmhalfgrid] (-0.08,20) -- (4.62,20);
    \draw[vlmhalfgrid] (-0.08,60) -- (4.62,60);
    \filldraw[kioatbar] (0.12,0) rectangle (0.64,72.0);
    \node[vlmbarvalue,anchor=south] at (0.38,73.2) {34.0};
    \filldraw[kibinbar] (1.08,0) rectangle (1.60,56.0);
    \node[vlmbarvalue,anchor=south] at (1.34,57.2) {32.0};
    \filldraw[kifastbar] (2.04,0) rectangle (2.56,48.0);
    \node[vlmbarvalue,anchor=south] at (2.30,49.2) {31.0};
    \filldraw[kiquestbar] (3.00,0) rectangle (3.52,64.0);
    \node[vlmbarvalue,anchor=south] at (3.26,65.2) {33.0};
    \filldraw[kiacodecbar] (3.96,0) rectangle (4.48,64.0);
    \node[vlmbarvalue,anchor=south] at (4.22,65.2) {33.0};
  \end{tikzpicture} &
  \vlmkisummarysep &
  \begin{tikzpicture}[vlmsummaryplot]
    \vlmkiframe{Avg. Success}{0/50,40/55,80/60}
    % Average TC success rates: OAT, Bin, FAST, QueST, ACodec.
    \draw[vlmhalfgrid] (-0.08,20) -- (4.62,20);
    \draw[vlmhalfgrid] (-0.08,60) -- (4.62,60);
    \filldraw[kioatbar] (0.12,0) rectangle (0.64,72.0);
    \node[vlmbarvalue,anchor=south] at (0.38,73.2) {59.0};
    \filldraw[kibinbar] (1.08,0) rectangle (1.60,24.8);
    \node[vlmbarvalue,anchor=south] at (1.34,26.0) {53.1};
    \filldraw[kifastbar] (2.04,0) rectangle (2.56,52.0);
    \node[vlmbarvalue,anchor=south] at (2.30,53.2) {56.5};
    \filldraw[kiquestbar] (3.00,0) rectangle (3.52,75.2);
    \node[vlmbarvalue,anchor=south] at (3.26,76.4) {59.4};
    \filldraw[kiacodecbar] (3.96,0) rectangle (4.48,66.4);
    \node[vlmbarvalue,anchor=south] at (4.22,67.6) {58.3};
  \end{tikzpicture}
\end{tabular*}
\endgroup

\par\vspace{1.2ex}
\noindent
\begingroup
\setlength{\tabcolsep}{0pt}
\begin{tabular*}{\vlmkirowwidth}{@{}c@{\vlmkifirstgap}c@{\vlmkirowgap}c@{\vlmkirowgap}c@{\vlmkirowgap}c@{\extracolsep{\fill}\vlmkisepcompensate}c@{\extracolsep{\fill}}c@{}}
\vlmkirowtag{\qwenvl} &
  \begin{tikzpicture}[vlmplot]
    \vlmkiframenoheading{0/80,26.67/85,53.33/90,80/95}
    \vlmkiylabel
    % LIBERO TC success rates: OAT, Bin, FAST, QueST, ACodec.
    \filldraw[kioatbar] (0.12,0) rectangle (0.64,61.9);
    \node[vlmbarvalue,anchor=south] at (0.38,63.1) {91.6};
    \filldraw[kibinbar] (1.08,0) rectangle (1.60,27.7);
    \node[vlmbarvalue,anchor=south] at (1.34,28.9) {85.2};
    \filldraw[kifastbar] (2.04,0) rectangle (2.56,57.1);
    \node[vlmbarvalue,anchor=south] at (2.30,58.3) {90.7};
    \filldraw[kiquestbar] (3.00,0) rectangle (3.52,53.9);
    \node[vlmbarvalue,anchor=south] at (3.26,55.1) {90.1};
    \filldraw[kiacodecbar] (3.96,0) rectangle (4.48,61.3);
    \node[vlmbarvalue,anchor=south] at (4.22,62.5) {91.5};
  \end{tikzpicture} &
  \begin{tikzpicture}[vlmplot]
    \vlmkiframenoheading{0/50,26.67/60,53.33/70,80/80}
    % RoboMimic TC success rates: OAT, Bin, FAST, QueST, ACodec.
    \filldraw[kioatbar] (0.12,0) rectangle (0.64,69.3);
    \node[vlmbarvalue,anchor=south] at (0.38,70.5) {76.0};
    \filldraw[kibinbar] (1.08,0) rectangle (1.60,12.0);
    \node[vlmbarvalue,anchor=south] at (1.34,13.2) {54.5};
    \filldraw[kifastbar] (2.04,0) rectangle (2.56,30.7);
    \node[vlmbarvalue,anchor=south] at (2.30,31.9) {61.5};
    \filldraw[kiquestbar] (3.00,0) rectangle (3.52,25.3);
    \node[vlmbarvalue,anchor=south] at (3.26,26.5) {59.5};
    \filldraw[kiacodecbar] (3.96,0) rectangle (4.48,45.3);
    \node[vlmbarvalue,anchor=south] at (4.22,46.5) {67.0};
  \end{tikzpicture} &
  \begin{tikzpicture}[vlmplot]
    \vlmkiframenoheading{0/20,20/30,40/40,60/50,80/60}
    % RoboCasa365 TC success rates: OAT, Bin, FAST, QueST, ACodec.
    \filldraw[kioatbar] (0.12,0) rectangle (0.64,69.6);
    \node[vlmbarvalue,anchor=south] at (0.38,70.8) {54.8};
    \filldraw[kibinbar] (1.08,0) rectangle (1.60,10.4);
    \node[vlmbarvalue,anchor=south] at (1.34,11.6) {25.2};
    \filldraw[kifastbar] (2.04,0) rectangle (2.56,56.8);
    \node[vlmbarvalue,anchor=south] at (2.30,58.0) {48.4};
    \filldraw[kiquestbar] (3.00,0) rectangle (3.52,60.8);
    \node[vlmbarvalue,anchor=south] at (3.26,62.0) {50.4};
    \filldraw[kiacodecbar] (3.96,0) rectangle (4.48,52.8);
    \node[vlmbarvalue,anchor=south] at (4.22,54.0) {46.4};
  \end{tikzpicture} &
  \begin{tikzpicture}[vlmplot]
    \vlmkiframenoheading{0/15,20/20,40/25,60/30,80/35}
    % SimplerEnv / Bridge TC success rates: OAT, Bin, FAST, QueST, ACodec.
    \filldraw[kioatbar] (0.12,0) rectangle (0.64,50.0);
    \node[vlmbarvalue,anchor=south] at (0.38,51.2) {27.5};
    \filldraw[kibinbar] (1.08,0) rectangle (1.60,12.0);
    \node[vlmbarvalue,anchor=south] at (1.34,13.2) {18.0};
    \filldraw[kifastbar] (2.04,0) rectangle (2.56,34.0);
    \node[vlmbarvalue,anchor=south] at (2.30,35.2) {23.5};
    \filldraw[kiquestbar] (3.00,0) rectangle (3.52,72.0);
    \node[vlmbarvalue,anchor=south] at (3.26,73.2) {33.0};
    \filldraw[kiacodecbar] (3.96,0) rectangle (4.48,52.0);
    \node[vlmbarvalue,anchor=south] at (4.22,53.2) {28.0};
  \end{tikzpicture} &
  \vlmkisummarysep &
  \begin{tikzpicture}[vlmsummaryplot]
    \vlmkiframenoheading{0/40,16/45,32/50,48/55,64/60,80/65}
    % Average TC success rates: OAT, Bin, FAST, QueST, ACodec.
    \filldraw[kioatbar] (0.12,0) rectangle (0.64,72.0);
    \node[vlmbarvalue,anchor=south] at (0.38,73.2) {62.5};
    \filldraw[kibinbar] (1.08,0) rectangle (1.60,18.2);
    \node[vlmbarvalue,anchor=south] at (1.34,19.4) {45.7};
    \filldraw[kifastbar] (2.04,0) rectangle (2.56,51.2);
    \node[vlmbarvalue,anchor=south] at (2.30,52.4) {56.0};
    \filldraw[kiquestbar] (3.00,0) rectangle (3.52,58.6);
    \node[vlmbarvalue,anchor=south] at (3.26,59.8) {58.3};
    \filldraw[kiacodecbar] (3.96,0) rectangle (4.48,58.2);
    \node[vlmbarvalue,anchor=south] at (4.22,59.4) {58.2};
  \end{tikzpicture}
\end{tabular*}
\endgroup
\endgroup

%% file: figs/tikz/ordering_ablation_plot.tex
% !TEX root = ../../main.tex

\begingroup
\footnotesize
\definecolor{ordOAT}{HTML}{8A1538}
\tikzset{
  orderingplot/.style={x=0.88cm,y=0.02925cm,baseline=(current bounding box.south)},
  ordline/.style={line width=0.85pt},
  ordpoint/.style={line width=0.55pt},
  ordxline/.style={densely dashed,line width=0.75pt},
}
\newcommand{\ordbbox}{%
  \path[use as bounding box] (-0.85,-11.2) rectangle (3.38,88.0);
}
\newcommand{\ordxaxis}{%
  \foreach \x/\lab in {0/1,1/2,2/4,3/8} {
    \draw[black!55] (\x,0) -- (\x,-1.4);
    \node[anchor=north,font=\tiny] at (\x,-3.0) {$\lab$};
  }}
\newcommand{\ordframe}[1]{%
  \ordbbox
  \draw[black!70] (-0.08,0) -- (3.25,0);
  \draw[black!70] (-0.08,0) -- (-0.08,80);
  \node[anchor=south,font=\footnotesize] at (1.58,84) {#1};
  \ordxaxis}
\newcommand{\ordaxeslibero}{%
  \foreach \y in {16,32,48,64} {
    \draw[gray!18] (-0.08,\y) -- (3.25,\y);
  }
  \ordframe{\libero-Long}
  \node[anchor=south east,font=\tiny] at (0.10,83) {Success};
  \foreach \y/\lab in {0/10,16/20,32/30,48/40,64/50,80/60} {
    \draw[black!55] (-0.08,\y) -- (-0.15,\y);
    \node[anchor=east,font=\tiny] at (-0.20,\y) {$\lab$};
  }}
\newcommand{\ordaxesrobomimic}{%
  \foreach \y in {13.33,40,66.67} {
    \draw[gray!25,densely dashed] (-0.08,\y) -- (3.25,\y);
  }
  \foreach \y in {26.67,53.33} {
    \draw[gray!18] (-0.08,\y) -- (3.25,\y);
  }
  \ordframe{\robomimic}
  \foreach \y/\lab in {0/50,26.67/60,53.33/70,80/80} {
    \draw[black!55] (-0.08,\y) -- (-0.15,\y);
    \node[anchor=east,font=\tiny] at (-0.20,\y) {$\lab$};
  }
}
\newcommand{\ordaxesmetaworld}{%
  \foreach \y in {20,60} {
    \draw[gray!25,densely dashed] (-0.08,\y) -- (3.25,\y);
  }
  \draw[gray!18] (-0.08,40) -- (3.25,40);
  \ordframe{\metaworld}
  \foreach \y/\lab in {0/10,40/20,80/30} {
    \draw[black!55] (-0.08,\y) -- (-0.15,\y);
    \node[anchor=east,font=\tiny] at (-0.20,\y) {$\lab$};
  }}
\newcommand{\ordaxesrobocasa}{%
  \foreach \y in {20,60} {
    \draw[gray!25,densely dashed] (-0.08,\y) -- (3.25,\y);
  }
  \draw[gray!18] (-0.08,40) -- (3.25,40);
  \ordframe{\robocasa}
  \foreach \y/\lab in {0/40,40/50,80/60} {
    \draw[black!55] (-0.08,\y) -- (-0.15,\y);
    \node[anchor=east,font=\tiny] at (-0.20,\y) {$\lab$};
  }}
\newcommand{\ordxref}[2]{%
  \draw[ordxline,#1] (-0.08,#2) -- (3.25,#2);
  \node[anchor=east,font=\tiny,text=#1] at (3.22,{#2+7.0}) {\oat[\times]};
}
\newcommand{\ordmark}[2]{%
  \foreach \p in {#2}
    \filldraw[ordpoint,fill=#1,fill opacity=0.5,draw=#1] \p circle (2.0pt);
}
% Coordinates below are generated by scripts/ordering_ablation_coords.py.
\noindent
\begin{tikzpicture}[orderingplot]
  \ordaxeslibero
  \draw[ordline,ordOAT] (0,2.72) -- (1,47.68) -- (2,58.24) -- (3,74.08);
  \ordmark{ordOAT}{(0,2.72),(1,47.68),(2,58.24),(3,74.08)}
  \ordxref{ordOAT}{40.32}
  \node[anchor=north,font=\tiny] at (1.5,-8.2) {$k$ decoded tokens};
\end{tikzpicture}\hfill
\begin{tikzpicture}[orderingplot]
  \ordaxesrobomimic
  \draw[ordline,ordOAT] (0,2.13) -- (1,6.67) -- (2,40.80) -- (3,61.60);
  \ordmark{ordOAT}{(0,2.13),(1,6.67),(2,40.80),(3,61.60)}
  \ordxref{ordOAT}{29.60}
\end{tikzpicture}\hfill
\begin{tikzpicture}[orderingplot]
  \ordaxesmetaworld
  \draw[ordline,ordOAT] (0,5.20) -- (1,25.60) -- (2,38.00) -- (3,57.60);
  \ordmark{ordOAT}{(0,5.20),(1,25.60),(2,38.00),(3,57.60)}
  \ordxref{ordOAT}{30.40}
\end{tikzpicture}\hfill
\begin{tikzpicture}[orderingplot]
  \ordaxesrobocasa
  \draw[ordline,ordOAT] (0,30.80) -- (1,41.20) -- (2,46.80) -- (3,58.40);
  \ordmark{ordOAT}{(0,30.80),(1,41.20),(2,46.80),(3,58.40)}
  \ordxref{ordOAT}{34.00}
\end{tikzpicture}
\endgroup

%% file: figs/tikz/oat_ha_hl_heatmaps.tex
% !TEX root = ../../main.tex

\begingroup
\newcommand{\oatheatcell}[5]{%
  \filldraw[fill=blue!#1, draw=white, line width=0.45pt]
    (#2,#3) rectangle ++(0.74,0.74);
  \node[text=#5, font=\footnotesize] at ({#2+0.37},{#3+0.37}) {#4};
}
\newcommand{\oatheatcorner}{%
  \draw[gray!70, line width=0.65pt] (-0.43,3.32) -- (-0.02,2.98);
  \node[font=\footnotesize] at (0.00,3.28) {$H_a$};
  \node[font=\footnotesize] at (-0.43,2.99) {$H_l$};
}

\begin{minipage}[t]{0.245\linewidth}
  \vspace{0pt}
  \centering
  \subfloat[Half-horizon execution.\label{fig:oat_ha_hl_ablation_half_ha}]{
    \begin{tikzpicture}[x=0.98cm,y=0.98cm]
      \oatheatcorner
      \foreach \x/\lab in {0.37/8,1.11/16,1.85/32,2.59/64}
        \node[font=\footnotesize] at (\x,3.18) {\lab};
      \foreach \y/\lab in {2.59/1,1.85/2,1.11/4,0.37/8}
        \node[font=\footnotesize] at (-0.18,\y) {\lab};
      \oatheatcell{48}{0}{2.22}{38.3}{white}
      \oatheatcell{33}{0.74}{2.22}{26.0}{black}
      \oatheatcell{10}{1.48}{2.22}{6.4}{black}
      \oatheatcell{5}{2.22}{2.22}{2.8}{black}
      \oatheatcell{62}{0}{1.48}{50.2}{white}
      \oatheatcell{56}{0.74}{1.48}{45.9}{white}
      \oatheatcell{34}{1.48}{1.48}{26.6}{black}
      \oatheatcell{18}{2.22}{1.48}{12.9}{black}
      \oatheatcell{82}{0}{0.74}{68.1}{white}
      \oatheatcell{66}{0.74}{0.74}{54.8}{white}
      \oatheatcell{47}{1.48}{0.74}{37.7}{white}
      \oatheatcell{18}{2.22}{0.74}{13.1}{black}
      \oatheatcell{76}{0}{0}{63.0}{white}
      \oatheatcell{69}{0.74}{0}{57.1}{white}
      \oatheatcell{68}{1.48}{0}{56.3}{white}
      \oatheatcell{25}{2.22}{0}{19.4}{black}
    \end{tikzpicture}
  }
\end{minipage}\hfill
\begin{minipage}[t]{0.245\linewidth}
  \vspace{0pt}
  \centering
  \subfloat[Fixed 8-action execution.\label{fig:oat_ha_hl_ablation_fixed_8}]{
    \begin{tikzpicture}[x=0.98cm,y=0.98cm]
      \oatheatcorner
      \foreach \x/\lab in {0.37/8,1.11/16,1.85/32,2.59/64}
        \node[font=\footnotesize] at (\x,3.18) {\lab};
      \foreach \y/\lab in {2.59/1,1.85/2,1.11/4,0.37/8}
        \node[font=\footnotesize] at (-0.18,\y) {\lab};
      \oatheatcell{37}{0}{2.22}{28.7}{black}
      \oatheatcell{33}{0.74}{2.22}{26.0}{black}
      \oatheatcell{11}{1.48}{2.22}{7.6}{black}
      \oatheatcell{3}{2.22}{2.22}{0.8}{black}
      \oatheatcell{46}{0}{1.48}{37.1}{white}
      \oatheatcell{56}{0.74}{1.48}{45.9}{white}
      \oatheatcell{36}{1.48}{1.48}{28.4}{black}
      \oatheatcell{23}{2.22}{1.48}{17.5}{black}
      \oatheatcell{78}{0}{0.74}{64.6}{white}
      \oatheatcell{66}{0.74}{0.74}{54.8}{white}
      \oatheatcell{53}{1.48}{0.74}{42.9}{white}
      \oatheatcell{24}{2.22}{0.74}{18.3}{black}
      \oatheatcell{66}{0}{0}{54.5}{white}
      \oatheatcell{69}{0.74}{0}{57.1}{white}
      \oatheatcell{75}{1.48}{0}{62.5}{white}
      \oatheatcell{35}{2.22}{0}{27.3}{black}
    \end{tikzpicture}
  }
\end{minipage}\hfill
\endgroup

%% file: body/related_work.tex
% !TEX root = ../main.tex

\section{Related Work}
\sloppy

We organize the most relevant work around visuomotor policy interfaces, action
token representations, and block autoregressive generation.

\subsection{Visuomotor Policy Interfaces}

Action chunking predicts temporally coherent control segments and amortizes
inference across multiple environment
steps~\citep{zhao2023actpolicy,zhang2025actionchunkingexploratorydata}.
Diffusion and flow-matching policies provide strong continuous action decoders
for high-frequency control~\citep{chi2024diffusionpolicy,black2024pi0vla,black2025realtimeexecutionactionchunking},
while multitask systems extend this interface across task families, sensing
modalities, and symbolic-continuous planning
settings~\citep{liu2026factorizeddiffusionpolicy,chen2025multi,hoeg2025hybriddiffusionpolicy,liu2025hybridvla}.
These methods establish the action chunk as an effective control unit, but
usually leave it in continuous space. We study the complementary discrete
interface, where the chunk is represented as a token sequence.

Large robot policies increasingly condition on language and often reuse VLM
backbones for embodied
control~\citep{rt22023arxiv,driess2023palmeembodiedmultimodallanguage,octo2023,kim24openvla,li2024visionlanguagefoundationmodelseffective,li2024cogactfoundationalvisionlanguageactionmodel,qu2025spatialvlaexploringspatialrepresentations,wen2025tinyvlafastdataefficientvisionlanguageaction,intelligence2025pi05vla,nvidia2025gr00tn1openfoundation},
supported by large-scale datasets and generalist policy
efforts~\citep{oneill2023openxembodiment,walke2024bridgedatav2datasetrobot,khazatsky2025droidlargescaleinthewildrobot,bousmalis2023robocatselfimprovinggeneralistagent}.
For policies that generate action tokens, tokenization determines output
length, action validity, and next-token prediction difficulty, making it a
central design axis~\citep{zhong2025surveyvisionlanguageactionmodelsaction,vuong2025actiontokenizermatters,wang2025vqvlaimprovingvisionlanguageactionmodels}.
Discrete diffusion decoders retain a token interface while replacing fixed
left-to-right generation with iterative parallel
refinement~\citep{liang2025discretediffusionvlabringing}.

Token co-training systems, sometimes referred to as knowledge insulation in
other literature, instead use discrete action losses to update the VLM while a
diffusion or flow-matching expert consumes detached VLM context and executes
continuous actions~\citep{intelligence2025pi05vla,driess2025knowledge,intelligence2026pi07steerablegeneralistrobotic,fang2026molmoact2actionreasoningmodels}.
Data-mixture studies likewise examine how vision-language and cross-embodiment
data preserve VLM knowledge during robot
training~\citep{lin2026systematicstudydatamodalities}. Together, direct token
generation and token co-training motivate action representations that are both
predictable and useful as supervision.

\subsection{Action Tokenization and Ordered Representations}

Existing action tokenizers trade off compression, guaranteed decodability, and
ease of policy prediction. Per-dimension \bin is simple and totally
decodable but produces sequences whose length grows with action dimension and
chunk horizon~\citep{rt12022arxiv,rt22023arxiv,kim24openvla}. \fast compresses
chunks with frequency-domain structure and BPE, introducing a
low-to-high-frequency order together with variable-length decoding
issues~\citep{pertsch2025fast}. Learned tokenizers use neural encoders and
discrete bottlenecks for skill or latent action
abstractions~\citep{mete2024quest,lee2024behaviorgenwithlatentactions,ye2025latentactionpretrainingvideos}.
These systems commonly use discrete quantization methods, including vector
quantization and finite scalar
quantization~\citep{oord2017vq,mentzer2024fsq}; recent VLA implementations
include VQ-VLA and
\acodec~\citep{wang2025vqvlaimprovingvisionlanguageactionmodels,dong2026actioncodecmakesgoodaction}.
Token surrogates also appear in in-context imitation learning and robotic
sequence modeling~\citep{dipalo2024keypointactiontokensenable,fu2024incontextimitationlearningnexttoken,bonatti2022pactperceptionactioncausaltransformer}.
These methods establish action tokens as policy targets, but do not resolve
which token properties make policy learning reliable.

Ordered representations provide an inductive bias for consuming or generating
partial codes. Nested dropout and Matryoshka-style objectives retain information
at multiple budgets~\citep{ripple2014learnorderedreprwithnesteddropout,kusupati2022matryoshkareprlearning,cai2025matryoshkamultimodal,bachmann2025flextok}.
Image and sequence models further show that representation design and
autoregressive factorization change the modeling problem faced by a downstream
generator~\citep{kolesnikov2022uvim,wen2025principalcomponentsenablenew}.
This supports a broader view of latent-space generation in which
representations should be selected for downstream predictability as well as
reconstruction~\citep{tschannen2018recentadvvaerepr,dieleman2025latents,zhao2025npq,kim2025trainworstplanbest}.
Action tokens add a control-specific requirement: retained prefixes should
decode to executable action chunks rather than only embeddings.

\subsection{Block Autoregressive Generation}

Partially parallel generation reduces autoregressive latency by predicting,
verifying, or refining multiple tokens per stage. Block-wise parallel decoding,
speculative decoding, masked prediction, and non-autoregressive refinement
instantiate this idea in language and sequence
models~\citep{stern2018blockwiseparalleldecodingdeep,leviathan2023fastinferencetransformers,ghazvininejad2019maskpredict,lee2018deterministicnonautoregressiveneuralsequence}.
Recent VLA tokenizers likewise target latency and prediction difficulty through
fixed-size block prediction in action
generation~\citep{liu2025fasterefficientautoregressivevision,dong2026actioncodecmakesgoodaction},
while block diffusion interpolates between autoregressive and parallel
generation through block-level
denoising~\citep{arriola2025blockdiffusioninterpolatingautoregressive}.
\BAR provides a unified view of these action token generation schedules,
with fixed-size block prediction as one point in the broader family.
\fussy

%% file: body/conclusion.tex
% !TEX root = ../main.tex

\section{Conclusion and Limitations}

This paper studied action tokenization as an interface for visuomotor policy
learning. We formalized three desiderata: high compression, total decodability,
and ordered token structure. We then introduced \oatfamily, a learned tokenizer
whose mask-padded prefixes decode to executable action chunks. Early tokens
capture coarse control, while later tokens refine residual detail, enabling
flexible prefix-based generation in autoregressive policies.

Tokenizer diagnostics, lightweight control experiments, VLM-scale evaluations,
and targeted ablations consistently show that \oatfamily provides a strong
action token interface. For autoregressive policies, \oatfamily improves
closed-loop control. \BAR further formalizes token-wise, block-wise, one-shot,
and power-of-two schedules, making the tradeoff between policy call depth and
block prediction difficulty explicit. Under token co-training, tokenizer choice
remains consequential even when action tokens are not decoded at inference. In
this setting, the cross-entropy loss on the first \oatfamily target provides a
plan-like, action-chunk-level objective for the VLM prefill representation
consumed by the action expert.

The study is intentionally scoped to isolate tokenizer effects. We evaluate a
fixed set of manipulation benchmarks, policy backbones, action horizons, and
token budgets; broader embodiments, longer-horizon tasks, and larger real-world
mixtures remain important tests of the same design principles. We also keep the
flow-matching expert fixed across token co-training comparisons, leaving the
joint design of expert architecture and token supervision for future work.

Adaptive computation is a natural next
step~\citep{graves2017adaptivecomputationtimerecurrent,banino2021pondernetlearningponder,dehghani2019universaltransformers,elbayad2020depthadaptivetransformer}.
Current policies choose token or
block budgets before inference, but \oatfamily prefixes can be decoded at any
budget and \BAR makes the cost of additional generated blocks explicit.
Future policies could decide online whether another token or block warrants an
additional policy call, using token entropy, reconstruction uncertainty, or
downstream value estimates. This would reserve deeper autoregressive refinement
for precise or high-risk moments while limiting computation for simple control
decisions.

%% file: body/appendix.tex
% !TEX root = ../main.tex

\tableofcontents
\clearpage

\section{Block-Wise Autoregressive Training and Inference}
\label{appendix:bar}

This appendix completes the specification of block-wise autoregression (\BAR)
introduced in \cref{sec:bar} and its integration with \oatfamily. It details
block patterns, block-shifted teacher forcing, inference, and tokenizer
compatibility. Within autoregressive (AR) generation, \BAR covers token-wise AR,
one-shot parallel
decoding~\citep{dong2026actioncodecmakesgoodaction}, prior fixed-size block
prediction~\citep{liu2025fasterefficientautoregressivevision,dong2026actioncodecmakesgoodaction},
and intermediate schedules, thereby exposing the tradeoff between sequential
policy depth and within-block prediction difficulty. Related depth-reduction
methods include block-wise parallel decoding, masked prediction, and
non-autoregressive
generation~\citep{stern2018blockwiseparalleldecodingdeep,ghazvininejad2019maskpredict,lee2018deterministicnonautoregressiveneuralsequence}.

\subsection{Block Patterns}

For a target budget $K\leq H_l$ within a fixed-length action token sequence
$T_{1:H_l}$, let $\mathbf{b}=(b_0,b_1,\ldots,b_S)$ denote the endpoint list,
with $0=b_0<b_1<\cdots<b_S=K$. At stage $s$, \BAR generates the block
$G_s=T_{b_{s-1}+1:b_s}$, whose size is
\[
    g_s=|G_s|=b_s-b_{s-1}.
\]
After this stage, the realized prefix is $T_{1:b_s}$. For indexing, we use the
empty-prefix convention $G_0=T_{1:0}=\varnothing$ and set $g_0=0$.
The endpoint list therefore determines both the number of sequential policy
calls $S$ and the number of action tokens predicted in parallel at each stage;
full generation sets $K=H_l$.

As illustrated in \cref{fig:bar_patterns}, the \BAR family ranges from
token-wise autoregression to one-shot parallel decoding, with fixed-size and
variable-size blocks between those endpoints. This formulation is consistent with
recent evidence that token ordering and decoding order can strongly affect
partially parallel generative models~\citep{kim2025trainworstplanbest}.

For the shifted-slot implementation below, we restrict \BAR schedules to
nondecreasing generated block sizes,
\[
    g_1\leq g_2\leq\cdots\leq g_S.
\]
Early stages use smaller blocks, preserving finer-grained serial conditioning;
later stages reduce policy calls by predicting larger blocks.
In particular, we use the positive endpoints
\[
    (b_1,b_2,\ldots,b_S)=(1,2,4,8,16,\ldots,2^{S-1}).
\]
These endpoints induce the block sizes
\[
    (g_1,g_2,\ldots,g_S)=(1,1,2,4,8,\ldots,2^{S-2}).
\]
For full generation with $K=H_l=2^{S-1}$, this schedule reduces the sequential
policy depth from $O(H_l)$ to $O(\log H_l)$.

The power-of-two schedule uses the fewest stages under a balanced growth
constraint.
Starting from $b_1=1$, require each later generated block to be no larger than
the realized prefix. For $s\geq2$,
\[
\begin{aligned}
    g_s=b_s-b_{s-1} &\leq b_{s-1}, \\
    b_s=b_{s-1}+g_s &\leq 2b_{s-1}.
\end{aligned}
\]
It follows that $S$ stages reach at most $b_S\leq 2^{S-1}$ tokens. Reaching a
length-$K$ prefix therefore requires at least
$1+\lceil\log_2 K\rceil$ stages. When $K$ is a power of two, doubling the
prefix length at each stage meets this bound exactly.

\subsection{Block-Shifted Teacher Forcing}

\BAR trains each generation stage with the same shifted context structure used
at inference. For stage $s$, define the block-shifted input sequence
\[
    X^{(s)}
    =
    T_{1:b_{s-1}}
    \oplus
    \langle\mathtt{MASK}\rangle^{g_s-g_{s-1}}.
\]
The final $g_s$ slots of $X^{(s)}$ are the logit-read slots. When $s>1$, these
slots contain the previous block context followed by padding masks:
\[
    G_{s-1}\oplus\langle\mathtt{MASK}\rangle^{g_s-g_{s-1}}.
\]
Nondecreasing block sizes ensure that this shifted window can carry the previous
block while adding only new mask slots as the block width grows. The policy
reads logits from these slots and computes cross-entropy against the current
block $G_s$. Let $\pi$ denote the policy, $o$ its
observation context, and
$p_{\pi,j}^{(s)}(\cdot\mid X^{(s)},o)$ denote the categorical distribution read
from the $j$-th final logit-read slot. The loss is
\[
    \mathcal{L}_{\mathrm{BAR}}
    =
    -\frac{1}{S}
    \sum_{s=1}^{S}
    \frac{1}{g_s}
    \sum_{j=1}^{g_s}
    \log
    p_{\pi,j}^{(s)}\!\left(
      T_{b_{s-1}+j} \mid X^{(s)}, o
    \right).
\]

We therefore average within each block and weight all generation stages
equally. This block-shifted objective produces all $g_s$ logits in one policy
forward pass while blocking access to ground-truth tokens from the current
block. Each logit-read slot sees only realized prefix tokens or masks and is
supervised against its target token in $G_s$; a block-causal attention mask still
allows mask-slot states to interact.

\subsection{Block-Wise Inference}

Inference mirrors the same block structure and shifted slots. Stage $1$ feeds
$g_1$ masks and predicts $\widehat{G}_1$. For $s>1$, the policy uses
\[
    \widehat{X}^{(s)}
    =
    \widehat{T}_{1:b_{s-1}}
    \oplus
    \langle\mathtt{MASK}\rangle^{g_s-g_{s-1}},
\]
whose final $g_s$ logit-read slots are
\[
    \widehat{G}_{s-1}\oplus\langle\mathtt{MASK}\rangle^{g_s-g_{s-1}}.
\]
The
previous block symbols in these slots are shifted context, not newly appended
tokens. The policy reads the $g_s$ predictions from the final slots and appends
only the predicted block $\widehat{G}_s$ to the prefix. After the final stage,
the generated action token prefix is
\[
    \widehat{T}_{1:K}
    =
    \widehat{G}_1\oplus\widehat{G}_2\oplus\cdots\oplus\widehat{G}_S.
\]
When $K=H_l$, this is the complete sequence and can be detokenized by any
tokenizer that supports full-sequence decoding. A tokenizer that supports
partial decoding can instead decode $\widehat{T}_{1:K}$ at any target budget;
similarly, after stage $s$, the prefix $\widehat{T}_{1:b_s}$ can be decoded as a
lower-budget action chunk. Prefixes at trained reconstruction budgets receive
direct decoder supervision and typically reconstruct more accurately than
intermediate prefixes.

\subsection{Reconstruction Budgets and Generation Endpoints}

Tokenizer reconstruction budgets and \BAR generation endpoints serve distinct
roles. Both evaluated variants train the decoder at reconstruction budgets
$\mathcal{K}=\{1,2,4,8,\ldots,H_l\}$. The \BAR endpoint list $\mathbf{b}$
specifies the blocks used to reach a target budget $K$ and may contain
intermediate endpoints that are not decoder reconstruction points.

For \oat, ordinary causal register attention orders individual token positions,
and \BAR uses singleton blocks with endpoints $(1,2,\ldots,K)$. Reaching a
length-$K$ prefix therefore requires $K$ sequential policy calls. For
\oatpowtwo, the block-causal register groups and \BAR endpoints follow the
power-of-two reconstruction budgets. With $H_l=16$, endpoints
$(1,2,4,8,16)$ induce block sizes $(1,1,2,4,8)$ and require five calls to
generate the complete sequence. For either variant, a generated prefix
$\widehat{T}_{1:K}$ can be suffix-padded to length $H_l$ with learned mask tokens
and passed to the decoder. Inference can then stop and execute the decoded action
chunk or continue toward a larger budget. Reconstruction budgets in
$\mathcal{K}$ are directly optimized and generally provide stronger
reconstruction quality, as summarized in \cref{algo:oat_policy_inference}.

\clearpage

\section{Token Co-Training for Vision-Language-Action Control}
\label{appendix:tc}

\begin{figure}[H]
  \centering
  \includegraphics[width=\linewidth]{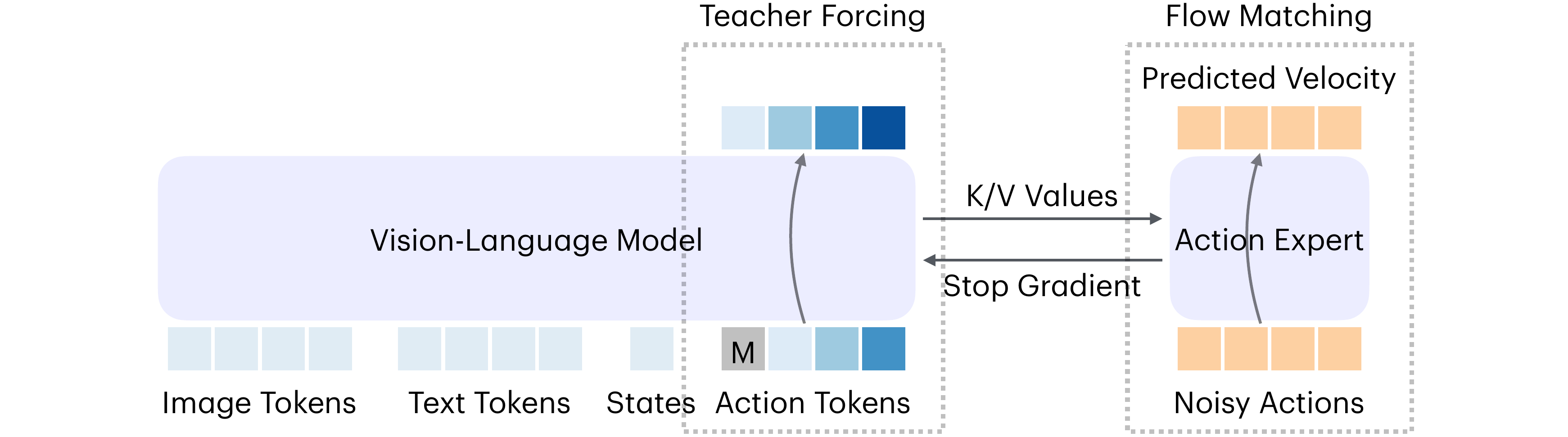}
  \caption{\textbf{TC computation graph.} The VLM predicts
  targets produced by a frozen \oatfamily tokenizer under teacher forcing, and
  the resulting token loss trains the VLM. Gray ``M'' denotes the leading mask
  used to predict the first target. In parallel, the flow-matching expert
  receives noisy actions and a detached K/V cache from the image, language, and
  state prefill; the cache excludes the teacher-forced action token positions.
  At inference, action token logits are not sampled, and the expert generates a
  continuous action chunk from the cached context.}
  \label{appendix:fig:tc_training_interface}
\end{figure}

This appendix expands the token co-training (TC) interface in \cref{sec:tc} for
vision-language-action (VLA) control. TC combines token supervision for the
vision-language model (VLM) with a continuous flow-matching expert. Continuous
action generative models are effective for
control~\citep{chi2024diffusionpolicy,black2024pi0vla}, but allowing the
corresponding continuous action loss to update a pretrained VLM can degrade its
knowledge~\citep{driess2025knowledge}. The stop-gradient boundary in
\cref{eq:tc_objective} isolates the VLM from the flow loss while preserving its
context as input to the expert~\citep{driess2025knowledge}.
\Cref{appendix:fig:tc_training_interface} summarizes the resulting computation
graph.

\paragraph{Teacher-forced token supervision.}
Let $c$ denote the image, language, and robot state context. For a continuous
action chunk $a_{1:H_a}$, the frozen tokenizer supplies targets
$T_{1:H_l}=\mathcal{T}(a_{1:H_a})$. The token loss in
\cref{eq:tc_objective} is the cross-entropy for predicting each $T_i$ from $c$
and the shifted action token prefix
$\langle\mathtt{MASK}\rangle,T_{1:i-1}$. The leading mask supplies the
prediction slot for $T_1$, and this branch updates the VLM through action token
prediction.

\paragraph{Detached context and flow matching.}
The expert receives the layer-wise key/value (K/V) cache
$\mathrm{KV}_{\mathrm{VLM}}(c)$ from the image, language, and state prefill. This
cache excludes K/V entries from the teacher-forced action token positions.
Image, language, and state therefore condition the expert through this detached
cache rather than through separate raw inputs. We min--max normalize each action
dimension to $[-1,1]$.
Given a normalized action chunk $a$, standard Gaussian noise
$\epsilon\sim\mathcal{N}(0,I)$, and flow time
$\tau=0.001+0.999u$ with $u\sim\operatorname{Beta}(1.5,1.0)$, we use
\[
\begin{aligned}
    a_\tau &= \tau\epsilon+(1-\tau)a, \\
    v &= \epsilon-a.
\end{aligned}
\]
Thus the noised action $\widetilde{a}_{1:H_a}^{\,\tau}$ in
\cref{eq:tc_objective} is $a_\tau$. The expert takes the detached K/V cache,
$a_\tau$, and a flow-time embedding as inputs. Its flow loss is the mean squared
error (MSE)
between the predicted velocity and $v$. We use all VLM layers by default, set
both loss weights to $1.0$ (i.e., $\lambda=1$ in \cref{eq:tc_objective}), and
apply stop-gradient only to the VLM context passed to the expert.

\paragraph{Inference.}
For each action chunk, the VLM is prefilled once on $c$ and its logits are
discarded. Starting from Gaussian noise at $\tau=1$, the flow-matching expert
uses the cached VLM context to integrate the learned velocity backward to
$\tau=0$ with 10 uniform Euler steps. Executed actions therefore come from the
expert rather than detokenization.

In principle, any tokenizer that maps continuous action chunks to discrete
symbols can provide TC targets. Prior systems often use Frequency-space Action
Sequence Tokenization
(\fast)~\citep{driess2025knowledge,intelligence2025pi05vla,
intelligence2026pi07steerablegeneralistrobotic,pertsch2025fast}.
Within each backbone, our comparison holds the VLM, action expert, cache
interface, flow objective, loss weights, and inference solver fixed; only the
tokenizer and resulting supervision targets change.
\clearpage

\section{Experimental Protocol and Implementation}
\label{appendix:implementation_details}

This appendix reports the rollout protocol, tokenizer configurations, policy
interfaces, and optimization recipes used in \cref{sec:experiments}. Full
launch-level configuration is provided in the released code.

\subsection{Evaluation Protocol}

Unless otherwise specified, simulated tasks use 50 evaluation episodes per
task and report mean success rate; real-world tasks use 20 independent rollouts
and report completed trials. Within each environment, all methods use the same
benchmark split, rollout horizon, and success criterion. Policies predict
contiguous action chunks with benchmark-specific action dimension $D_a$. All
benchmarks use action horizon $H_a=32$, except \simpler, which uses $H_a=8$.
At test time, we execute only the first half of each predicted chunk before
querying the policy again.
Because success rates are finite rollout estimates, we interpret very small
gaps as practical ties and emphasize consistent trends and benchmark-level
differences rather than rank changes from small margins.

\subsection{Tokenizer Configurations}

The comparison covers the tokenizer families analyzed in
\cref{sec:action_token_prelim}. \bin discretizes each action dimension into
$N=256$ uniform bins, producing $H_aD_a$ scalar tokens following the action
discretization used by \citet{rt12022arxiv}, \citet{rt22023arxiv}, and
\citet{kim24openvla}. \fast~\citep{pertsch2025fast} uses its universal
tokenizer with vocabulary size 2048. \quest~\citep{mete2024quest} compresses
action chunks with a temporal convolution with a downsampling factor of 2 before learned
tokenization. \acodec~\citep{dong2026actioncodecmakesgoodaction} uses learnable
registers that cross-attend to the action chunk without attending to one
another, forming a one-shot parallel decoding endpoint. \oatfamily summarizes
action chunks with learnable registers and ordered partial codes; we evaluate
the token-wise and power-of-two attention masks in
\cref{sec:oat_instantiations}.

For \fast, an arbitrary byte-pair encoding (BPE) token sequence sampled by an
autoregressive (AR) policy may decode to a coefficient stream whose length
differs from the fixed topology required by the inverse transform. In simulated
rollouts, we use nonstrict decoding unless otherwise specified: the coefficient
stream is padded or truncated to the expected length before the inverse discrete
cosine transform (DCT). This keeps every sampled sequence executable but may
shift frequency-coefficient positions. For physical rollouts, we instead use
strict decoding: an invalid sequence is rejected and the policy is queried
again.

For fair comparison among learned chunk tokenizers, \quest, \acodec, and the
\oatfamily variants use the same encoder and decoder capacity: 6 Transformer
layers, 8 attention heads, and model dimension 256. \quest and \oatfamily use
finite scalar quantization (FSQ) with levels $[8,8,6,5]$, corresponding to
$8\times8\times6\times5=1920\approx2048$ discrete
codes~\citep{mentzer2024fsq}. For \acodec, we
follow the official implementation and use vector quantization with latent
dimension 512 and codebook size 2048.

\subsection{Policy Implementations}

The lightweight AR policy uses the \ctrltransformer backbone: image and robot
state observations enter through cross-attention, and the model predicts action
token logits directly. Vision-language model (VLM) policies instead place
observation, language, and state tokens in the language model context and
attach an action token suffix. To make action codes valid VLM outputs, we
extend each VLM tokenizer with the special token set
\[
  \begin{aligned}
  \mathcal{S}_{\mathrm{act}}
  &=
  \{\texttt{<|action\_}i\texttt{|>}: i\in\{0,\ldots,|\mathcal{V}|-1\}\}\\
  &\quad\cup
  \{\texttt{<|action\_mask|>}\}.
  \end{aligned}
\]
Each \texttt{<|action\_}i\texttt{|>} maps to one discrete action code, and
\texttt{<|action\_mask|>} supplies the mask symbol for block-wise
autoregression (\BAR). The VLM retains its backbone-specific context attention:
\paligemma uses full attention over context tokens, whereas \qwenvl uses causal
attention. The action suffix follows the selected block-causal schedule, which
reduces to standard causal attention for singleton blocks.

AR and token co-training (TC) policies differ in how this suffix is used. In
VLM AR policies, the action suffix uses the block-shifted masks from
\cref{sec:bar} and is detokenized into continuous actions. In VLM TC policies,
action token cross-entropy updates the VLM, but token logits are not sampled at
inference. A flow-matching expert generates actions from noisy action inputs
conditioned on detached layer-wise VLM key/value (K/V) context; the flow loss
updates only the expert. We use velocity flow matching with 10 Euler steps.

\subsection{Optimization Recipes}

\cref{appendix:tab:training_recipes} lists the optimization recipe used by each
model family. Learned latent tokenizers use one recipe; all discrete
AR policies, including \BAR patterns, use a single policy optimization recipe
independent of generation pattern; and TC policies use a separate recipe for
their joint token and flow objective. This presentation keeps the appendix
focused on comparability:
within a family, differences in closed-loop success should be read against a
fixed optimizer, learning rate schedule, weight decay, gradient clipping, and
batch size. In \cref{appendix:tab:training_recipes}, Opt., LR, LR sched., Min LR
ratio, WD, and Clip denote the optimizer, learning rate, learning rate schedule,
minimum learning rate ratio, weight decay, and gradient clipping norm,
respectively.

\begin{center}
  \centering
  \scriptsize
  \setlength{\tabcolsep}{4.0pt}
  \renewcommand{\arraystretch}{1.12}
  \begin{tabular*}{\linewidth}{@{\hspace{0.55em}\extracolsep{\fill}}>{\raggedright\arraybackslash}p{0.18\linewidth}ccccccc@{\hspace{0.55em}}}
    \toprule
    Component & Opt. & LR & Batch & LR sched. & Min LR ratio & WD & Clip \\
    \midrule
    Latent tokenizers
      & AdamW & 5e-5 & 512 & constant & -- & 0 & 1.0 \\
    AR policies
      & AdamW & 1e-4 & 16 & cosine & 0.1 & 1e-6 & 1.0 \\
    TC policies
      & AdamW & 5e-5 & 32 & constant & -- & 1e-6 & 1.0 \\
    \bottomrule
  \end{tabular*}
  \captionof{table}{\textbf{Training recipes.} Optimization settings for learned latent
  tokenizers, AR policies, and TC policies. The AR row covers token-wise,
  fixed-size, one-shot, and variable block patterns. Opt. denotes the optimizer;
  LR and LR sched. denote the learning rate and its schedule; Min LR ratio is the
  minimum learning rate divided by the initial rate; WD denotes weight decay;
  and Clip denotes the gradient clipping norm.}
  \label{appendix:tab:training_recipes}
\end{center}
\clearpage

\section{Action Tokenizer Baselines and Their Tradeoffs}
\label{appendix:baseline_tokenizers}

\subsection{Per-Dimension Binning Is Total but Long}

Per-dimension \bin is a standard baseline for autoregressive (AR) robot
policies and provides total decoding over its discrete output
space~\citep{rt12022arxiv,rt22023arxiv,kim24openvla}. Each scalar action
coordinate is normalized to a fixed range, commonly $[-1,1]$. The range is then
divided into $N$ uniform bins, and each coordinate is mapped to the
corresponding bin index. For an action
chunk of shape $H_a\times D_a$, \bin produces the serialization
\[
    \mathcal{T}(a_{1:H_a})
    =
    [T_{1,1},\ldots,T_{1,D_a},T_{2,1},\ldots,T_{H_a,D_a}].
\]
Each coordinate token satisfies $T_{i,j}\in[N]$. If every scalar is emitted as
its own token, the resulting token horizon is $H_l=H_aD_a$.

\bin is reliable because its detokenizer is simple and total. Every bin index
maps back to a scalar action value, so every sampled token sequence of the
expected length maps to an action chunk. It also has near-perfect reconstruction
as the number of bins grows, up to quantization error. Because the $D_a$
coordinates at one time step can be predicted jointly, \bin is compatible with
simple block-wise generation.

The drawbacks are its high token rate and manually imposed ordering. Common
action chunks can require hundreds of tokens, increasing training cost and
leaving at least $H_a$ sequential
prediction steps under the natural block pattern. The token order is
also a manual serialization over dimensions and time: a prefix may
contain a few coordinates of the first action step while saying nothing about
the rest of the trajectory. Thus \bin is reliable as a decoder and supports
coordinate blocks, but remains poor as a compact, predictable action
representation: it satisfies \propertyII and simple block-wise generation, but
fails to satisfy \propertyI and \propertyIII. The next natural attempt is to compress action
chunks while preserving a structured order.

\subsection{Frequency-Domain Tokens Are Ordered but Not Total}

Frequency-domain tokenizers address the rate and ordering weaknesses of
per-dimension binning by representing action trajectories through spectral
coefficients~\citep{cooley1965fourierseries}. Frequency-space Action Sequence
Tokenization (\fast) is one representative example: it applies the discrete
cosine transform (DCT), quantizes the resulting spectral coefficients, and then
applies byte-pair encoding
(BPE)~\citep{ahmed1974dct,gage1994bpe,sennrich2016neural,pertsch2025fast}.
The resulting order is compatible with AR policies because
low-frequency components appear before high-frequency components. Early tokens tend to
describe coarse motion structure, while later tokens encode higher-frequency
detail. This gives \fast a form of AR-friendly ordering and high information
density.

The limitation is structural decodability. A robot action chunk requires a fixed
coefficient topology before it can be reshaped and transformed back into
$H_a\times D_a$ control. After deterministic pruning, quantization, or
flattening, the detokenizer expects a coefficient stream of fixed length
$N_{\mathrm{coef}}$,
and the positions in this stream are not interchangeable: each index
corresponds to a particular frequency basis, temporal component, and action
dimension.

BPE breaks this fixed-topology assumption because tokens expand to
variable-length coefficient sequences. Let $e(T_i)$ be the coefficient
subsequence obtained by expanding token $T_i$. For a generated sequence
$T_{1:L}$, the recovered coefficient stream has length
\begin{align*}
    \left|e(T_1)\oplus e(T_2)\oplus\cdots\oplus e(T_L)\right|
    =
    \sum_{i=1}^{L}|e(T_i)|.
\end{align*}
Valid decoding requires this length to equal $N_{\mathrm{coef}}$. Equivalently, the natural
domain of the \fast detokenizer is
\begin{align*}
    \mathcal{D}_{\mathrm{FAST}}
    =
    \{T_{1:L}:\sum_{i=1}^{L}|e(T_i)|=N_{\mathrm{coef}}\},
\end{align*}
which is a strict subset of the token sequences an AR policy can
emit. A next-token policy is not inherently constrained to stay inside
$\mathcal{D}_{\mathrm{FAST}}$, so a generated sequence can expand to too few or
too many coefficients. In that case, the inverse reshape and inverse frequency
transform are mathematically undefined.

This creates a mismatch with fixed-position policy interfaces. Padding or
truncation can keep the length valid, but later symbols occupy shifted
coefficient slots. Rejection or constrained decoding instead changes what the
policy may sample. Thus \fast is compact and ordered, but not total over
unconstrained policy outputs. Its variable-length expansion also destabilizes
fixed token budgets and block-wise autoregression (\BAR) schedules, motivating
learned latents with a fixed token horizon and a decoder defined over the full
latent space.

\subsection{Learned Latents Are Compact but Not Necessarily Predictable}

Learned latent tokenizers address compression by learning a neural bottleneck
for action chunks. Methods such as Quantized Skill Transformer (\quest) and
\acodec map an action chunk into a latent sequence of shape
$H_l\times D_l$, quantize the latents with vector quantization or finite scalar
quantization (FSQ), and decode them back into
continuous actions~\citep{lee2024behaviorgenwithlatentactions,oord2017vq,mentzer2024fsq,mete2024quest,wang2025vqvlaimprovingvisionlanguageactionmodels,dong2026actioncodecmakesgoodaction}.
The latent horizon $H_l$ and latent dimension $D_l$ are hyperparameters, often
chosen to be much smaller than the raw action dimension $H_aD_a$. For example,
an action chunk with horizon $H_a=32$ can be represented by a latent sequence
with $H_l=8$ tokens.

These tokenizers are attractive from a rate--distortion viewpoint. They can be
much more compact than \bin, and their learned decoders are total over the
discrete latent space: any valid code index at each latent position can be
embedded and decoded into a continuous action chunk. This satisfies
\propertyI and \propertyII under the policy's supported token vocabulary.
However, compactness and total decodability do not by themselves make the tokens
learnable policy targets~\citep{ramanujan2026when}. For sequential latent tokenizers such as \quest, a latent
sequence optimized mainly for endpoint reconstruction need not put important
motion information early. The policy may then have to predict latent indices in
an order that was not designed for next-token learning.

\acodec provides the prior fixed-size block prediction baseline in our
comparison. Its latent groups are designed for joint prediction and joint
decoding, so the policy emits a fixed block of action tokens per generation
step~\citep{dong2026actioncodecmakesgoodaction}.
This reduces serial policy depth without exposing an AR-friendly prefix order.
In our taxonomy, the tradeoff is block modeling difficulty: the policy must
infer the tokens in each fixed block together. The remaining gap is an ordered
latent code that stays compact and total while giving next-token prediction a
more learnable conditional structure.
\clearpage

\section{Rate--Distortion Across Action Token Budgets}
\label{appendix:rate_distortion}

\cref{appendix:tab:num_tokens} gives the numeric values behind the
rate--distortion curves in \cref{fig:rate_distortion_curves}. Full-budget
baselines appear as one operating point per benchmark and include per-dimension
\bin, Frequency-space Action Sequence Tokenization (\fast), Quantized Skill
Transformer (\quest), and ActionCodec (\acodec). \oatfamily reports
partial decodings at $k\in\{1,2,4,8,16\}$ under token-wise and power-of-two
variants. For \fast, the token count is the average byte-pair encoding (BPE)
length. The table reports the number of tokens (\#Tok) and mean squared error
(MSE), reported in units of $10^{-3}$,
to keep the table compact; the main-paper figure plots the corresponding raw
MSE values on logarithmic axes. The table highlights the diagnostic role of
rate--distortion: the power-of-two variant closely tracks the token-wise variant
across token budgets, so the closed-loop differences should be read as
differences in policy prediction difficulty and generation pattern, not as
reconstruction failures.

\begin{center}
  \centering
  \footnotesize
  \setlength{\tabcolsep}{7.2pt}
  \renewcommand{\arraystretch}{1.08}
  \begin{tabular*}{\linewidth}{@{\hspace{0.55em}\extracolsep{\fill}}l|S[table-format=3.1]S[table-format=2.1]|S[table-format=3.1]S[table-format=2.1]|S[table-format=3.1]S[table-format=3.1]|S[table-format=3.1]S[table-format=2.1]|S[table-format=3.1]S[table-format=1.2]@{\hspace{0.55em}}}
    \toprule
    \multirow{2}{*}{Scheme}
    & \multicolumn{2}{c|}{\libero}
    & \multicolumn{2}{c|}{\robomimic}
    & \multicolumn{2}{c|}{\metaworld}
    & \multicolumn{2}{c|}{\robocasaThreeSixFive}
    & \multicolumn{2}{c}{\simpler} \\
    & {\#Tok} & {\makecell{MSE\\($\times 10^{-3}$)}}
    & {\#Tok} & {\makecell{MSE\\($\times 10^{-3}$)}}
    & {\#Tok} & {\makecell{MSE\\($\times 10^{-3}$)}}
    & {\#Tok} & {\makecell{MSE\\($\times 10^{-3}$)}}
    & {\#Tok} & {\makecell{MSE\\($\times 10^{-3}$)}} \\
    \midrule
    \bin      & 224 & 0.0 & 224 & 0.0 & 128 & 1.2 & 384 & 0.0 & 56 & 0.00 \\
    \fast     & 47.5 & 0.3 & 50.9 & 0.4 & 20.5 & 67.9 & 42.2 & 0.2 & 11.0 & 0.19 \\
    \quest    & 16 & 2.8 & 16 & 3.1 & 16 & 24.9 & 16 & 1.5 & 4 & 0.94 \\
    \acodec   & 16 & 0.7 & 16 & 2.3 & 16 & 8.2 & 16 & 1.4 & 16 & 0.04 \\
    \midrule
    \oat[1]   & 1 & 14.7 & 1 & 12.4 & 1 & 731.4 & 1 & 16.0 & 1 & 0.81 \\
    \oat[2]   & 2 & 7.3 & 2 & 7.8 & 2 & 110.0 & 2 & 8.9 & 2 & 0.35 \\
    \oat[4]   & 4 & 3.8 & 4 & 5.3 & 4 & 38.1 & 4 & 5.2 & 4 & 0.21 \\
    \oat[8]   & 8 & 2.1 & 8 & 3.1 & 8 & 32.7 & 8 & 2.5 & 8 & 0.11 \\
    \oat[16]  & 16 & 1.0 & 16 & 1.9 & 16 & 32.1 & 16 & 1.3 & 16 & 0.05 \\
    \midrule
    \oatpowtwo[1]  & 1 & 16.2 & 1 & 12.9 & 1 & 734.6 & 1 & 16.2 & 1 & 0.82 \\
    \oatpowtwo[2]  & 2 & 7.4 & 2 & 7.8 & 2 & 107.1 & 2 & 8.9 & 2 & 0.35 \\
    \oatpowtwo[4]  & 4 & 3.6 & 4 & 5.3 & 4 & 44.4 & 4 & 4.9 & 4 & 0.20 \\
    \oatpowtwo[8]  & 8 & 1.9 & 8 & 3.0 & 8 & 27.7 & 8 & 2.4 & 8 & 0.11 \\
    \oatpowtwo[16] & 16 & 0.9 & 16 & 1.8 & 16 & 26.4 & 16 & 1.2 & 16 & 0.06 \\
    \bottomrule
  \end{tabular*}
  \captionof{table}{\textbf{Rate--distortion of action tokenizers.} \#Tok is the number
  of discrete action tokens used for one action chunk; for \fast, it is the
  average BPE length. MSE is measured after detokenization in raw action space
  and reported in units of $10^{-3}$; lower is better. Baseline tokenizers are evaluated at
  their full generated length, while \oat[k] and \oatpowtwo[k] decode only the
  first $k$ ordered tokens to show token-wise and power-of-two partial
  reconstruction.}
  \label{appendix:tab:num_tokens}
\end{center}
\clearpage

\section{Autoregressive Action Generation with Vision-Language Models}
\label{appendix:vlm_policy_results}

\cref{appendix:tab:simulation_benchmarking} gives the tabular values behind the
vision-language model (VLM) policy success plots with autoregressive (AR) action
token generation in
\cref{fig:vlm_bar_success_grid}. The table compares
\oatfamily with per-dimension \bin, Frequency-space Action Sequence
Tokenization (\fast), Quantized Skill Transformer (\quest), and ActionCodec
(\acodec). For \fast, symbolic token and policy call counts depend on the
generated byte-pair encoding (BPE) length. In the table, \# Tokens and \# Calls
denote the token count and sequential policy call count, while Avg. denotes
average. The table separates the two VLM backbones so that tokenizer effects are
compared within a fixed policy model, then reports success across the four
benchmark groups and two benchmark-balanced summaries. The \libero column
averages the Long, Goal, Object, and Spatial suites, matching the aggregation
used in the main-paper figure.

\begin{table}[!ht]
  \centering
  \footnotesize
  \setlength{\tabcolsep}{5.2pt}
  \renewcommand{\arraystretch}{1.0}
  \begin{tabular*}{\linewidth}{@{\hspace{0.55em}\extracolsep{\fill}}l|cc|S[table-format=2.1]S[table-format=2.1]S[table-format=2.1]S[table-format=2.1]|S[table-format=2.1]S[table-format=2.1]@{\hspace{0.55em}}}
    \toprule
    \multicolumn{9}{c}{\paligemma} \\
    \midrule
    Scheme
    & \# Tokens
    & \# Calls
    & {\libero}
    & {\robomimic}
    & {\makecell{\robocasaThreeSixFive}}
    & {\simpler}
    & {Avg. Success}
    & {Avg. Rank} \\
    \midrule
    \bin & $H_aD_a$ & $H_a$ & 14.2 & 7.0 & 3.2 & 20.5 & 11.2 & 13.3 \\
    \fast & $|T|$ & $|T|$ & 70.2 & 62.0 & 44.4 & 17.0 & 48.4 & 9.5 \\
    \quest & $H_l$ & $H_l$ & 65.9 & 64.5 & 49.6 & 54.0 & 58.5 & 5.9 \\
    \acodec & $H_l$ & 1 & 78.7 & 54.0 & 51.2 & 26.5 & 52.6 & 6.3 \\
    \midrule
    \oat[1] & 1 & 1 & 28.2 & 10.0 & 24.8 & 16.0 & 19.8 & 13.0 \\
    \oat[2] & 2 & 2 & 72.5 & 29.0 & 44.8 & 35.5 & 45.5 & 9.4 \\
    \oat[4] & 4 & 4 & 73.4 & 42.0 & 48.0 & 35.5 & 49.7 & 8.6 \\
    \oat[8] & 8 & 8 & 78.0 & 65.0 & 58.8 & 39.0 & 60.2 & 4.8 \\
    \oat[16] & 16 & 16 & 79.7 & 66.0 & 60.4 & 48.5 & 63.7 & 2.9 \\
    \midrule
    \oatpowtwo[1] & 1 & 1 & 31.4 & 8.5 & 28.4 & 18.0 & 21.6 & 12.3 \\
    \oatpowtwo[2] & 2 & 2 & 73.7 & 24.5 & 49.2 & 44.5 & 48.0 & 8.0 \\
    \oatpowtwo[4] & 4 & 3 & 76.3 & 48.0 & 62.4 & 49.0 & 58.9 & 5.0 \\
    \oatpowtwo[8] & 8 & 4 & 78.2 & 53.5 & 62.8 & 56.5 & 62.8 & 3.3 \\
    \oatpowtwo[16] & 16 & 5 & 80.8 & 60.0 & 60.4 & 54.0 & 63.8 & 3.0 \\
    \bottomrule
  \end{tabular*}

  \vspace{0.9em}

  \begin{tabular*}{\linewidth}{@{\hspace{0.55em}\extracolsep{\fill}}l|cc|S[table-format=2.1]S[table-format=2.1]S[table-format=2.1]S[table-format=2.1]|S[table-format=2.1]S[table-format=2.1]@{\hspace{0.55em}}}
    \toprule
    \multicolumn{9}{c}{\qwenvl} \\
    \midrule
    Scheme
    & \# Tokens
    & \# Calls
    & {\libero}
    & {\robomimic}
    & {\makecell{\robocasaThreeSixFive}}
    & {\simpler}
    & {Avg. Success}
    & {Avg. Rank} \\
    \midrule
    \bin & $H_aD_a$ & $H_a$ & 0.0 & 0.5 & 1.6 & 1.0 & 0.8 & 14.0 \\
    \fast & $|T|$ & $|T|$ & 62.8 & 27.5 & 32.8 & 5.0 & 32.0 & 10.6 \\
    \quest & $H_l$ & $H_l$ & 58.4 & 49.0 & 14.8 & 6.0 & 32.1 & 9.5 \\
    \acodec & $H_l$ & 1 & 76.5 & 63.5 & 45.2 & 23.5 & 52.2 & 4.5 \\
    \midrule
    \oat[1] & 1 & 1 & 33.0 & 12.0 & 27.2 & 5.5 & 19.4 & 11.8 \\
    \oat[2] & 2 & 2 & 70.9 & 29.5 & 48.4 & 11.0 & 40.0 & 8.8 \\
    \oat[4] & 4 & 4 & 76.4 & 34.5 & 55.2 & 12.0 & 44.5 & 6.8 \\
    \oat[8] & 8 & 8 & 78.5 & 52.5 & 57.6 & 13.5 & 50.5 & 4.1 \\
    \oat[16] & 16 & 16 & 82.0 & 67.0 & 62.0 & 16.0 & 56.8 & 1.5 \\
    \midrule
    \oatpowtwo[1] & 1 & 1 & 31.1 & 8.5 & 32.8 & 3.5 & 19.0 & 12.4 \\
    \oatpowtwo[2] & 2 & 2 & 72.7 & 20.0 & 52.4 & 13.0 & 39.5 & 8.3 \\
    \oatpowtwo[4] & 4 & 3 & 78.8 & 31.0 & 53.2 & 15.0 & 44.5 & 5.6 \\
    \oatpowtwo[8] & 8 & 4 & 79.3 & 47.0 & 55.6 & 19.0 & 50.2 & 3.8 \\
    \oatpowtwo[16] & 16 & 5 & 81.8 & 48.0 & 57.6 & 15.0 & 50.6 & 3.5 \\
    \bottomrule
  \end{tabular*}
  \caption{\textbf{Closed-loop VLM AR policy success rates.} Separate table
  blocks report the two VLM backbones. Within each block, rows compare action
  token schemes and, for \oatfamily, token budgets; \# Tokens and \# Calls give
  the token count and sequential policy call count. Entries are mean task
  success rates over 50 rollouts per task. \textit{Avg. Success} gives equal weight to \libero, \robomimic,
  \robocasaThreeSixFive, and \simpler; \textit{Avg. Rank} applies the same
  aggregation across benchmarks to per-benchmark ranks, where lower is better.
  For baseline tokenizers, \# Tokens and \# Calls are symbolic because token
  counts can depend on benchmark action dimensions or generated BPE length; for
  \oat and \oatpowtwo, they report the evaluated token budget and the number of
  policy calls required by the generation pattern.}
  \label{appendix:tab:simulation_benchmarking}
\end{table}